\documentclass[runningheads]{llncs}
\usepackage{graphicx}
\usepackage{tikz}
\usepackage{comment}
\usepackage{amsmath,amssymb} % define this before the line numbering.
\usepackage{color}

\usepackage{graphicx}
\usepackage{amsmath}
\usepackage{booktabs}
\usepackage{float}
\usepackage{graphicx}
\usepackage{epsfig}
\usepackage[ruled, boxed,commentsnumbered]{algorithm2e}
\usepackage{listings}
\usepackage{subfigure}

\newcommand{\eqnref}[1]{Eqn. (\ref{#1})}
\newcommand{\algref}[1]{Alg. (\ref{#1})}
\newcommand{\secref}[1]{Sec. (\ref{#1})}
\newcommand{\figref}[1]{Fig. (\ref{#1})}
\newcommand{\tabref}[1]{Tab. (\ref{#1})}

\newcommand{\nascent}{{\sf NAScenT}\xspace}

\newcommand{\vvec}[1]{\mathbf{#1}}   % vector
\newcommand{\model}{\mathcal{M}}% the full hybrid model (!!! change)
\newcommand{\mlp}{\Phi}         % mlp network
\newcommand{\level}{l}          % level index

\newcommand{\x}{\vvec{x}}       % 3D point
\newcommand{\z}{z}              % depth (projected)
\newcommand{\dir}{\vvec{d}}     % direction
\newcommand{\ray}{r}            % generic ray
\newcommand{\rgb}{\vvec{c}}     % color vector
\newcommand{\density}{\sigma}   % voxel opacity / density
\newcommand{\sweight}{\delta}   % weight for each sample
\newcommand{\trans}{T}          % accumulated transmission to a point

\newcommand{\img}{I}            % image

\begin{document}
% \linenumbers
\pagestyle{headings}
\mainmatter
\def\ECCVSubNumber{7628}  % Insert your submission number here

\title{Neural Adaptive SCEne Tracing (\nascent)} % Replace with your title

%\begin{comment}
\titlerunning{ECCV-22 submission ID \ECCVSubNumber} 
\authorrunning{ECCV-22 submission ID \ECCVSubNumber} 
\author{Anonymous ECCV submission}
\institute{Paper ID \ECCVSubNumber}
%\end{comment}
%******************

%\begin{comment}
\titlerunning{\nascent}
%\author{Rui Li\inst{1}%\orcidID{0000-1111-2222-3333} 
%	\and Darius R\"uckert\inst{1,2} %\orcidID{1111-2222-3333-4444} 
%	\and Yuanhao Wang\inst{1} %\orcidID{2222--3333-4444-5555}}
%	\and Ramzi Idoughi\inst{1} %\orcidID{2222--3333-4444-5555}}
%	\and Wolfgang Heidrich\inst{1}} %\orcidID{2222--3333-4444-5555}}

%\author{Rui Li\inst{1}
%	\and Darius R\"uckert\inst{1,2} 
%	\and Yuanhao Wang\inst{1} 
%	\and Ramzi Idoughi\inst{1} 
%	\and Wolfgang Heidrich\inst{1}} 

\author{Rui Li\inst{1},\,Darius R\"uckert\inst{2},\,Yuanhao Wang\inst{1},\,Ramzi Idoughi\inst{1},\,Wolfgang Heidrich\inst{1}} 

\authorrunning{Li et al.}
\institute{King Abdullah University of Science and Technology, Thuwal, Saudi Arabia \and
Friedrich-Alexander-Universit\"at Erlangen-Nürnberg, Erlangen, Germany
\email{rui.li@kaust.edu.sa, darius.rueckert@fau.de, yuanhao.wang@kaust.edu.sa, ramzi.idoughi@kaust.edu.sa, wolfgang.heidrich@kaust.edu.sa}\\
%\url{https://www.arthurlirui.com} \and
}
%\end{comment}
%******************
\maketitle

\begin{abstract}
Neural rendering with implicit neural networks has recently emerged
as an attractive proposition for scene reconstruction, achieving
excellent quality albeit at high computational cost. While the most
recent generation of such methods has made progress on the rendering
(inference) times, very little progress has been made on improving
the reconstruction (training) times.

In this work we present Neural Adaptive Scene Tracing (\nascent),
the first neural rendering method based on directly training a
hybrid explicit-implicit neural representation. \nascent uses a
hierarchical octree representation with one neural network per leaf
node and combines this representation with a two-stage sampling
process that concentrates ray samples where they matter most -- near
object surfaces. As a result, \nascent is capable of reconstructing
challenging scenes including both large, sparsely populated volumes
like UAV captured outdoor environments, as well as small scenes
with high geometric complexity. \nascent outperforms existing neural
rendering approaches in terms of both quality and training time.
\end{abstract}

\section{Introduction}

In recent years, inverse rendering methods based on implicit neural
networks such as NeRF~\cite{mildenhall2020nerf} and its variants (e.g.~\cite{yu2020pixelnerf}, \cite{liu2020neural}, \cite{Reiser2021ICCV}, \cite{martel2021acorn}, \cite{Lin_2021_ICCV}, \cite{meshry2019neural}, \cite{lindell2021autoint}) have
garnered a lot of interest in both computer graphics and computer
vision. These methods have led to a massive improvement in the quality
of 3D reconstruction and re-rendering tasks. Unfortunately, this
quality improvement comes at a high computational cost during both
training and inference (re-rendering), since the implicit network must
be evaluated at millions of points. This shortcoming has so far
precluded the use of implicit neural networks for the reconstruction
of very large scenes.

In parallel to the development of these neural inverse rendering
methods, we have also seen the introduction of {\em neural scene
  representations}~\cite{Yifan:IsoPoints:2021}, \cite{sitzmann2019siren},
  \cite{martel2021acorn}. These are not concerned with solving an inverse
problem, but instead take an existing image or volume, and compress it
into a compact neural network representation. In this space, the ACORN
system~\cite{martel2021acorn} has shown that hybrid explicit-implicit
representations based on hierarchical octree representations can yield
a improvements in terms of both the compute time and the
quality of fine details in representations of large images and
volumes.

\begin{figure*}[t]
	\centering
	\centering
	\includegraphics[width=0.9\linewidth]{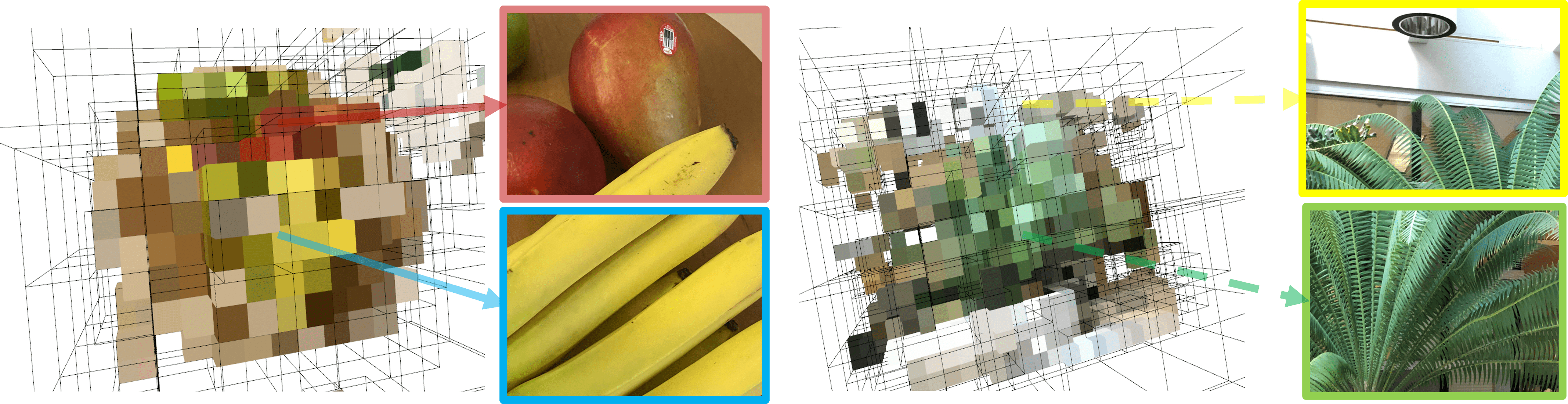}
	\caption{\nascent jointly optimizes a hybrid explicit-implicit
		representation consisting of an octree for 3D space
		partitioning, and structured networks in each active leaf
		node. Each network maps a spatial coordinate and a direction
		to a view-independent density and a view-dependent color.
		\nascent adaptively allocates more tree nodes to parts of
		the 3D space with higher scene complexity. Shown here are
		renderings of novel views, from the fruit and fern
		datasets. }
	\label{fig:teaser}
\end{figure*}

Here, we introduce Neural Adaptive Scene Tracing (\nascent), a hybrid
explicit-implicit neural representation that can be {\em trained
  directly} on scene reconstruction tasks
(Figure~\ref{fig:teaser}). \nascent uses an octree representation to
partition the space into regions according to scene complexity. Each
octree node has its own small-scale MLP to represent the node
contents. The fully differentiable rendering pipeline employs a
ray-based importance sampling scheme in this hierarchical
representation, with the importance being determined by an initial
node-based splatting approach that maximizes sample reuse across
views.

With this approach, \nascent achieves both high detail accuracy for
large scenes, as well as fast training and inference. The adaptive
representation works well for a large range of scene types and camera
positions, from complex small scale scenes with either full angular
coverage or light-field like directional coverage all the way to large
sparse volumes that arise in UAV-based capture of large-scale
environments.

Specifically, our contributions are: (1) we propose an octree-based
neural representation method that represents a scene as an octree with
a coordinate-based neural network inside each leaf node and can be
trained directly from 2D image data; (2) we also propose an octree
structure optimization method that jointly solves multiple neural
networks representation and computational resource allocation
problems; (3) our representation method can handle challenging cases
of large viewpoint change and dynamic camera range cases, e.g.
UAV-view terrain scanning.

% moved here for better placement

%\begin{enumerate}
	%\item A jointly optimization framework for octree-based sampling and implicit 3D scene reconstruction, viewpoint synthesis network architecture, 
	%\item an adaptive, hierarchical sampling, training, rendering pipeline, %for computational resource allocation.
	%\item enable free viewpoint traveling in implicit scene with highly photo-realistic details.
%\end{enumerate}

%Benefit of Multi-Light Weight Model:
%\begin{enumerate}
%	\item Enable high sampling rate
%	\item scale to efficient rendering for large scene
%	\item fast training and model distilling
%	\item become a building block for large scene
%\end{enumerate}

%Benefit of Heavy-Weight Model
%\begin{itemize}
%	\item can represent high capacity of scene content
%	\item high quality training and representation
%\end{itemize}

\section{Related Works}

\paragraph{3D Scene Reconstruction} is an active research topic in
computer graphics. The goal of 3D scene reconstruction is to infer the
3D geometry and texture of a real scene from active
measurements~\cite{kuhner2020large}, passive
imaging~\cite{aharchi2019review} or by combining
both~\cite{gurram20073d}. This task is fundamental in several
application fields such as scene understanding, object detection,
robot navigation, and industrial inspection. During the last decades,
several approaches have been proposed to reconstruct scenes from 2D
captured images~\cite{schonberger2016structure}, \cite{zollhofer2018state},
  \cite{aharchi2019review}, \cite{ham2019computer}, \cite{dahnert2021panoptic}. In our
work, we adopt a multi-view reconstruction approach, where a 3D model
of the scene is reconstructed from a set of 2D images taken from known
camera viewpoints~\cite{seitz2006comparison}. The traditional pipeline
first recovers camera pose for the multi-views system, and then
generates a sparse 3D points distribution of the scene by
Structure-from-Motion (SfM) technique. At this stage, a dense scene
reconstruction can be obtained by performing multi-view stereo
techniques. To enable a photo-realistic viewpoint change, a material
type or parametric reflection model can also be specified in the
rendering pipeline. Finally, a ray tracing can be performed using a
physically-based renderer to simulate the light propagation and the
camera imaging process. Recently, neural rendering techniques have
been applied with a huge success to the scene
reconstruction.

\paragraph{Neural Rendering} techniques have been a resounding success
in the computer graphics. They have been applied to achieve realistic
rendering of real scenes and improved the view
synthesis~\cite{eslami2018neural}, \cite{sitzmann2019scene},
  \cite{mildenhall2020nerf}, \cite{niemeyer2020differentiable}, \cite{schwarz2020graf},
  \cite{chan2021pi}, the relighting and material
editing~\cite{boss2021nerd}, \cite{srinivasan2021nerv}, \cite{xiang2021neutex},
  \cite{zhang2021physg}, the texture synthesis~\cite{oechsle2019texture},
  \cite{saito2019pifu}, \cite{chibane2020implicit}. Other applications of neural
rendering are discussed in the survey~\cite{tewari2020state}.

The Neural Radiance Fields (NeRF) work~\cite{mildenhall2020nerf} paved
the way to a new sub-domain in neural rendering. NeRF and its many
adaptations show impressive results in several graphics
tasks. However, the large number of samples needed per ray and the
requirement to evaluate the network for each sample is a real obstacle
for real-time applications. Several strategies have been explored to
speed up the neural rendering using NeRF-like networks. These
approaches include pruning~\cite{liu2020neural}, network
factorizations~\cite{Reiser2021ICCV},
caching~\cite{garbin2021fastnerf}, use of dynamic data
structures~\cite{liu2020neural}, \cite{yu2021plenoctrees}, and directly
learning the integral along a ray~\cite{lindell2021autoint}.
Most of these approaches improve only the rendering performance, but
not the training. In this work we specifically target accelerations of
the training time by direct training on a hierarchical representation.

%, and the
%combining implicit-explicit representation in a multi-scale
%structure~\cite{martel2021acorn}.

\paragraph{3D Scene Representation} is of paramount importance in the
reconstruction process. Historically, several ways have been used for
the representation of the geometry of the scene, including regular 3D
grids of voxels representing discrete occupancy, point clouds, polygon
meshes, set of depth maps, or a function of the distance to the
closest surface~\cite{seitz2006comparison}. More recently, several
neural representation have been proposed. They can be classified into
explicit, implicit and hybrid representations. The explicit methods
describe the scene based on a collection of primitives like
voxels~\cite{sitzmann2019deepvoxels}, point
clouds~\cite{aliev2020neural}, meshes~\cite{hedman2018deep}, or
multi-plane images~\cite{zhou2018stereo}, \cite{flynn2019deepview}. The
rendering using these representations is fast, but their huge
requirements in terms of memory, make them challenging to scale.

On the other hand, coordinate-based networks have been introduced to
represent scenes in an implicit fashion using neural
network~\cite{eslami2018neural}, \cite{park2019deepsdf}, \cite{mildenhall2020nerf},
  \cite{sitzmann2019siren}, \cite{xian2021space}, \cite{chan2021pi}. These implicit neural
representations leverage a Multi-Layer Perceptron (MLP) to learn a
mapping from continuous coordinates to physical properties such as
density, field, occupancy or radiance distribution. Despite the
impressive results of these representation approaches, they suffer
from both a large training time and large rendering time, since the
network has to be evaluated for each voxel of the grid. A recent
exception is the ACORN system proposed by Martel et
al.~\cite{martel2021acorn}. It utilizes a hybrid
implicit-explicit multi-scale representation in order to combine the
computationally efficiency of explicit representations with the memory
scalability of implicit approaches. ACORN is also designed to prune
empty space in an optimized fashion, and its shows excellent
performance in representing fine scale detail on large object
domains. However, like several other works~\cite{Yifan:IsoPoints:2021},
  \cite{sitzmann2019siren}, ACORN is purely a neural representation, not a
neural rendering method. That is, these approaches can be used to
compress existing volumes into neural representations, but they cannot
in a straightforward way be used for solving scene reconstruction
problems.

Our neural representation is inspired by the hierarchical
representation of ACORN, but with several crucial adaptations that
make \nascent highly suitable for scene reconstruction tasks.

%
%In this section, we will give a brief review about scene reconstruction, 3D scene representation, viewpoint-dependent appearance modeling from multi-view images, especially related neural-based methods in terms of training strategies, sampling, representation, rendering fields.
%
%\paragraph{3D Scene Representation}
%Traditional ways to represent 3D scene are first generate sparse 3D points by structure-from-motion and recover camera pose for multi-view systems,  and then dense scene reconstruction can be performed by multi-view stereo. Generated 3D models can be represented by a set of 3D points with estimated normal and textures, i.e., point cloud or mesh. To enable a photo-realistic viewpoint change, a material type or parametric reflection model can also be specified in the rendering pipeline, then perform a ray tracing, by a physically-based renderer,  to simulate the light propagation and camera imaging process.
%However, point cloud or mesh assume the scene are roughly Lambertian, i.e., view invariant or follows physical reflection rules, which can not faithfully recover a true viewpoints of scene.
%
%LLFF\cite{mildenhall2019llff}, PlenOct\cite{yu2021plenoctrees} BARF\cite{Lin_2021_ICCV}
%
%Neural network 
%
%\paragraph{3D Scene Reconstruction}
%3D geometry and texture in real scene can be recovered from multi-view images by taking camera observations.
%\paragraph{Neural Rendering}
%
%Neural Representation
%\cite{lindell2021autoint} \cite{Wizadwongsa2021NeX} \cite{Yifan:IsoPoints:2021}
%
%Neural Scene Reconstruction 
%\cite{meshry2019neural}

\begin{figure*}[t]
	\centering
	\includegraphics[width=0.95\linewidth]{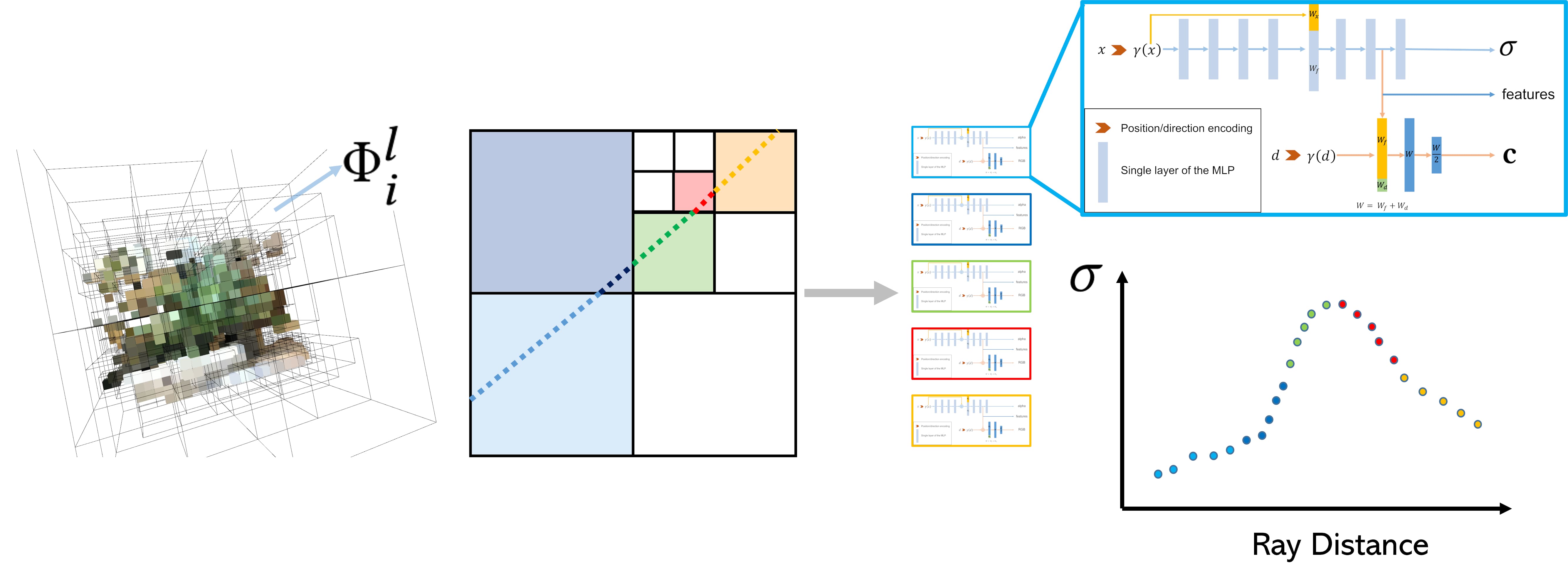}
	\caption{System diagram of \nascent. The architecture is an
          explicit-implicit neural representation for the 3D scene,
          consisting of an octree partitioning of space and a separate
          lightweight MLP for each leaf node of the octree. The same
          network architecture and hyper-parameters are used for all
          octree nodes, which concentrates the model parameters in
          regions of high complexity. This adaptive representation is
          combined with an adaptive sampling scheme and differentiable
          rendering described in the text.}
	\label{fig:pipeline}
\end{figure*}

\section{Method}

\nascent uses a hybrid explicit-implicit neural representation based
on a hierarchical octree data structure (\secref{sec:model}) in which
each leaf node has its own neural network, see
\figref{fig:pipeline}. This model is evaluated with a two-step
sampling approach that concentrates most samples in regions of high
geometric complexity as well as near object surfaces
(\secref{sec:sampling}). The samples are then composited front-to-back
(\secref{sec:imageformation}) to render images in a differentiable
fashion. In this way we can both optimize the neural networks in the
leaf nodes as well as adaptively refine the hierarchical model
structure (\secref{sec:modelupdate}, \secref{sec:pretrain}). The
details of the individual steps are discussed in the following.

\subsection{Hybrid Scene Model}
\label{sec:model}

\nascent uses a hybrid explicit-implicit scene model $\model$, that
maps a sample location $\x$ and a viewing direction $\dir$ to an RGB
color $\rgb$ and the density or opacity $\density$ of the sample:
\begin{equation}
  \model=\model_0^0: (\x,\dir) \rightarrow (\rgb,\density).
  \label{eq:model}
\end{equation}

The explicit part of the representation is somewhat inspired by the
hierarchical structure of ACORN~\cite{martel2021acorn}, however with a number of
important differences. Specifically, the model $\model_i^\level$ is
recursively defined as either a leaf node represented as a neural
network, or a subdivided node with exactly 8 child nodes in standard
octree fashion:

\begin{equation}
  \model_i^\level(\x,\dir) =
  \begin{cases}
    \mlp_i^\level: (\x,\dir) \rightarrow (\rgb,\density)&,\text{if
      leaf node}\\
    \bigcup\{\model_{i,1}^{l+1},\dots,\model_{i,8}^{l+1}\}&,\text{else}
  \end{cases}.
  \label{eq:model_hierarchy}
\end{equation}

Note that unlike previous hybrid neural representations like
ACORN~\cite{martel2021acorn}, \nascent does not use a global neural network, but
instead individual lightweight networks for the leaf nodes of the
octree representation.

The neural networks for each leaf node have the same MLP architecture,
depicted in \figref{fig:pipeline}. The network consists of a multi-layered
view-independent part and a single view-dependent layer. Note that
only the RGB color $\rgb$ depends on the viewing direction $\dir$,
while the density $\density$ is view independent. This allows us to
re-use calculated densities across multiple views (see sampling
process below).

The number of layers and neurons per layer in the view-dependent part
are hyper parameters, however unless otherwise noted, all experiments
in this paper use 8 layers with 64 neurons each. The view dependent
layer has 256 neurons. Positional encoding is used for both the
position $\x$ and the direction $\dir$ with 10 and 4 frequencies,
respectively. As the activation function, we use randomized leaky ReLU
(RReLU) with a negative lower ($-0.3$) and upper ($-0.1$).  Unless
otherwise noted, we limit the maximum octree level to 5.  The learning
rate starts at $5\cdot10^{-4}$ and is reduced by a factor of 0.1 every
10 epochs.

\subsection{Image Formation}
\label{sec:imageformation}

Like most recent neural inverse rendering works, \nascent targets
scenes that primarily consist of opaque surfaces. Such scenes are
represented well by the front-to-back compositing model introduced by
NeRF~\cite{mildenhall2020nerf}, which we replicate in the following for
completeness. Given a set of samples $\{\x_i\}_i$ along a ray
$\ray$ with direction $\dir_\ray$ and the associated color and density
values $(\rgb_i,\density_i) = \model(\x_i,\dir_\ray)$, the
corresponding image pixel is given as
\begin{equation}
  \img(\ray) = \sum_i \trans_i(1-e^{-\density_i\sweight_i})\rgb_i,
  \quad \text{where} \quad
  \trans_i = \exp\left[-\textstyle\sum_{j=1}^{i-1}\density_j\sweight_j\right].
\label{eq:composite}
\end{equation}

Here, $\trans_i$ is the cumulative transparency along the ray segment
leading up to sample $i$, and $\sweight_i$ is a sample weight based
on the length of the ray segment between successive samples similar to
NeRF~\cite{mildenhall2020nerf}, but computed independently for each octree node, so that
empty or low resolution nodes do not inflate the weight of the first
sample in the next node.

Note that this image formation model requires the samples to be
ordered front-to-back, since $T_i$ in \eqnref{eq:composite} requires
summation over all samples $j$ closer than $i$. This is
straightforward to achieve in non-adaptive representations like
NeRF~\cite{mildenhall2020nerf} or kiloNeRF~\cite{Reiser2021ICCV}, but requires extra book keeping
efforts in our adaptive, hierarchical approach. Furthermore, any
samples located behind an opaque surface will have zero contribution
to the pixel value, and will therefore also not contribute to the
gradient. Such samples can therefore be culled to reduce the
computational burden.

\subsection{Two-step Sampling and Ray-tracing}
\label{sec:sampling}
To address these issues we employ a two-step sampling process. First,
we use stratified regular sampling in the octree nodes to obtain an
estimate of the importance of volume regions to each ray. Then, we
apply a ray-based importance sampling scheme along each ray using the
information gathered in the first pass. 

\paragraph*{Stratified node-based sample generation}
Considering \eqref{eq:composite}, an important observation is that
$t_i$, the accumulated transparency along the first part of the ray
segment, can act as an effective importance function for the sampling
process, along with the hierarchical model structure itself, which
refines around regions of high complexity. Furthermore, this
cumulative transparency depends only on the density of the samples,
but not their color, and the densities independent of ray
direction. This makes it possible to re-use samples across different
views. 

To exploit this observation, we generate samples on stratified grids
within each octree leaf node. The number of samples is the same for
each leaf node ($64^3$ in our implementation), so that the evaluations
of the networks $\mlp_i^\level$ can be batched in a straightforward
fashion, while the adaptive nature of the octree naturally adjusts the
sampling density to the local scene complexity.

%\begin{figure}[h]
%	\centering
%	\includegraphics[width=0.7\linewidth]{figure/tbd}
%	\caption{View dependant part.}
%	\label{fig:subnetworks}
%\end{figure}

In this first sampling stage, we only evaluate the view-independent
part of the network, yielding the densities $\density_i$, which can be
re-used for all camera views. Furthermore, since these densities are
only used for importance sampling in the second stage, we do not need
to generate gradient information for this stage.  This makes the
process efficient despite the large number of samples generated.

\paragraph*{Sample sorting and ray compositing}
For each view, the samples generated in this fashion are projected
into the image plane, and associated with a pixel and the
corresponding ray $\ray$ (with ray id for each ray). Next, we need to sort the samples belonging
to each ray in depth. Instead of solving a large number of small
sorting problems, it is more efficient to sort all samples
simultaneously. To this end, we assign a global sorting key $\z_g$ to
each sample, which is given as
\begin{equation}\label{eqn:sort_by_z}
  \z_g = \ray \cdot\z_{\max} + \z_s,
\end{equation}
where $\z_s$ is the sample depth relative to the camera, $\z_{\max}$
is the maximum scene depth defines by the user, and $\ray$ is an
integer ray ID. Each sample is associated with the ray corresponding
to the pixel it projects to in a nearest-neighbor sense.

Sorting according to this global key will therefore bring all samples
into a global order in which successive groups of samples correspond
to the same ray, and each group is sorted by depth. The groups are
padded to the same maximum length, and then composited in parallel
according to \eqnref{eq:composite}.

\paragraph*{Ray-based importance sampling}

In the second sampling stage, we generate the actual ray-based
sampling pattern that is used for differentiable image rendering. When
the sorted stratified samples are given, we estimate the cumulative
density distribution (i.e., accumulative sum of $\sigma$) in each
block that similar to NeRF's hierarchical sampling
scheme~\cite{mildenhall2020nerf} (i.e., stratified sampling based on
spatial ray distance), but only evaluate the density distribution
within one node. Then, we apply importance sampling to reallocate the
samples according to the cumulative density distribution interval
(i.e., uniform sampling based on the CDF), assuming that the steep
slopes in the CDF indicate true
surfaces.

\subsection{Optimization of Hybrid Model}
\label{sec:modelupdate}

\begin{figure*}[ht]
	\centering
	\def \scale {0.18}
	\def \scaleB {1}
	\subfigure[init lv 0]{
		\begin{minipage}[t]{\scale\linewidth}
			\includegraphics[width=\scaleB\linewidth]{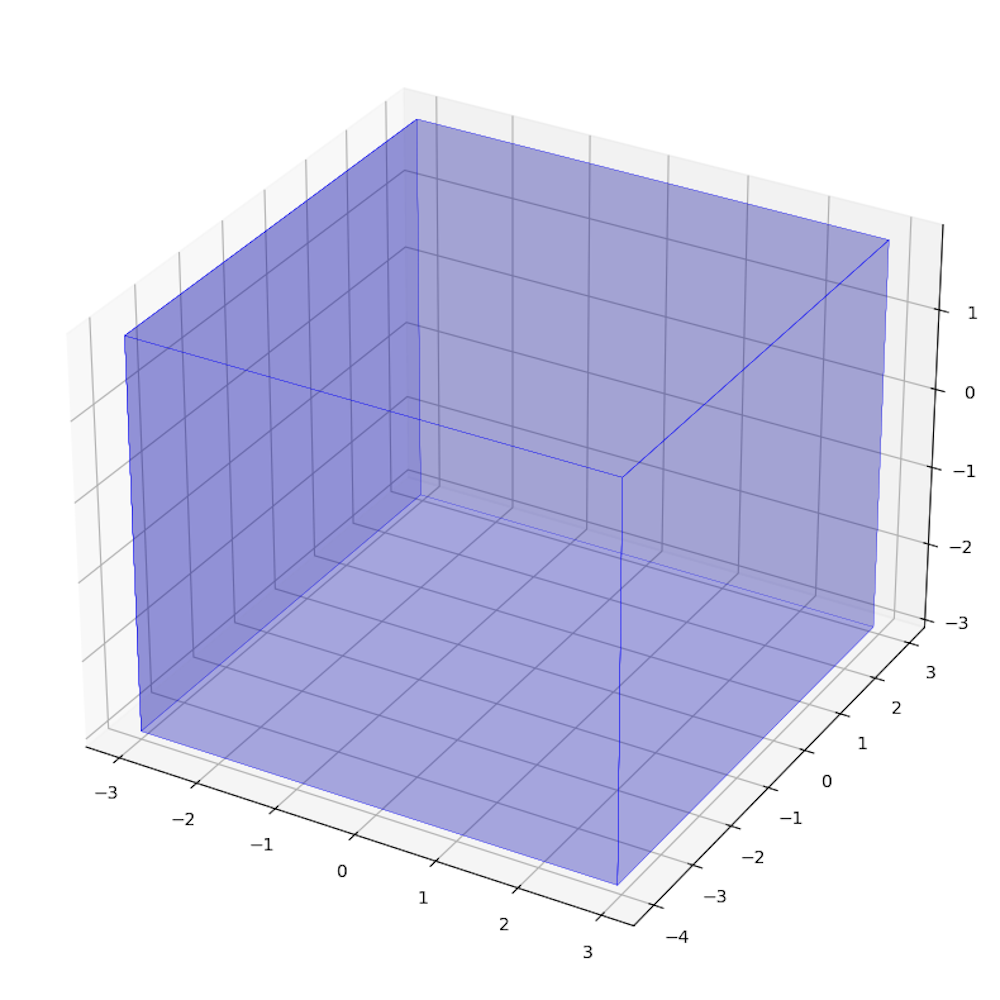}
	\end{minipage}}
	\subfigure[init lv 1]{
		\begin{minipage}[t]{\scale\linewidth}
			\includegraphics[width=\scaleB\linewidth]{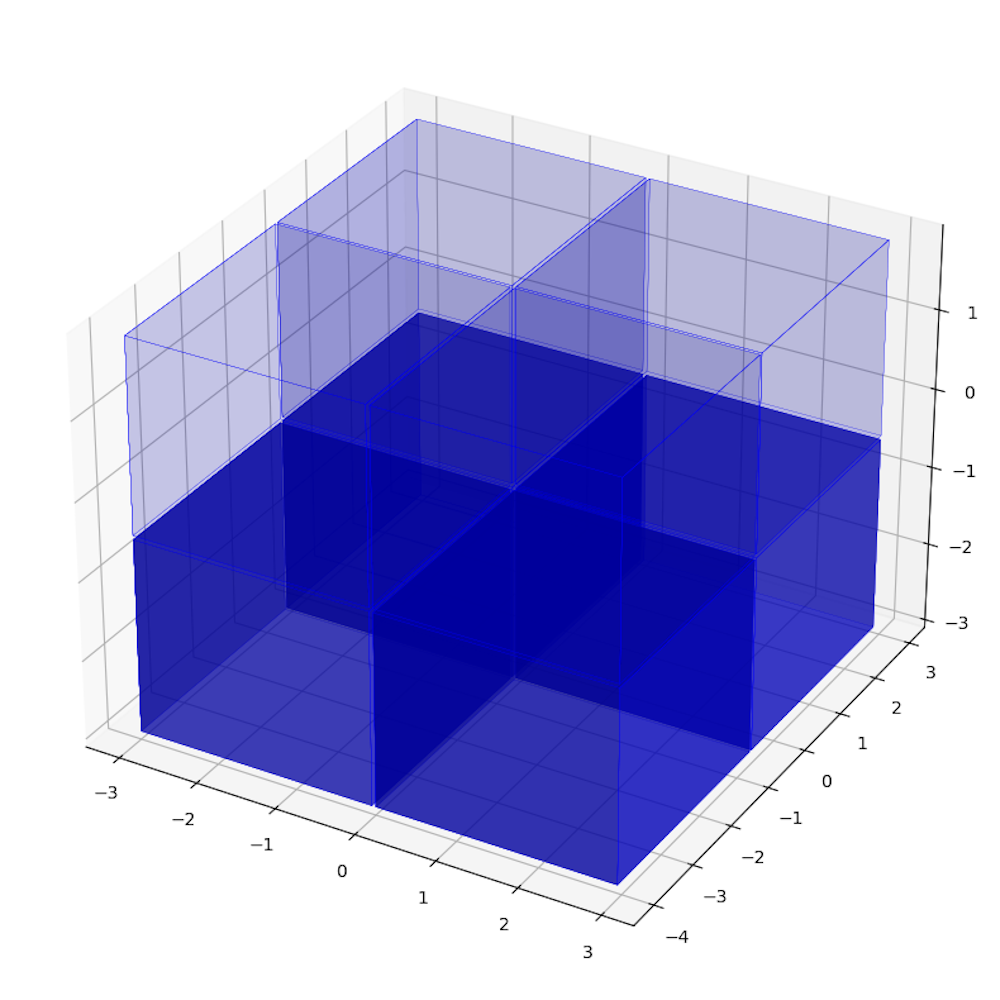}
	\end{minipage}}
	\subfigure[init lv 2]{
		\begin{minipage}[t]{\scale\linewidth}
			\includegraphics[width=\scaleB\linewidth]{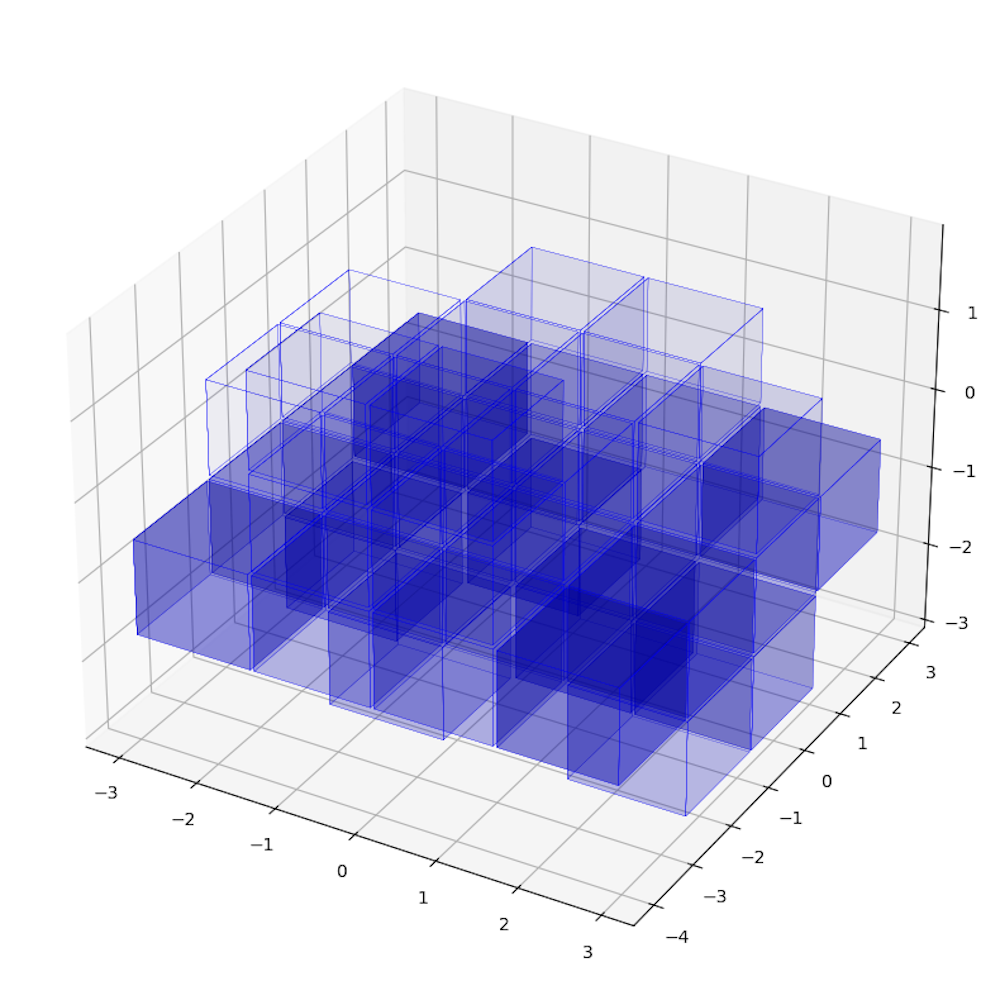}
	\end{minipage}}
	\subfigure[init lv 3]{
		\begin{minipage}[t]{\scale\linewidth}
			\includegraphics[width=\scaleB\linewidth]{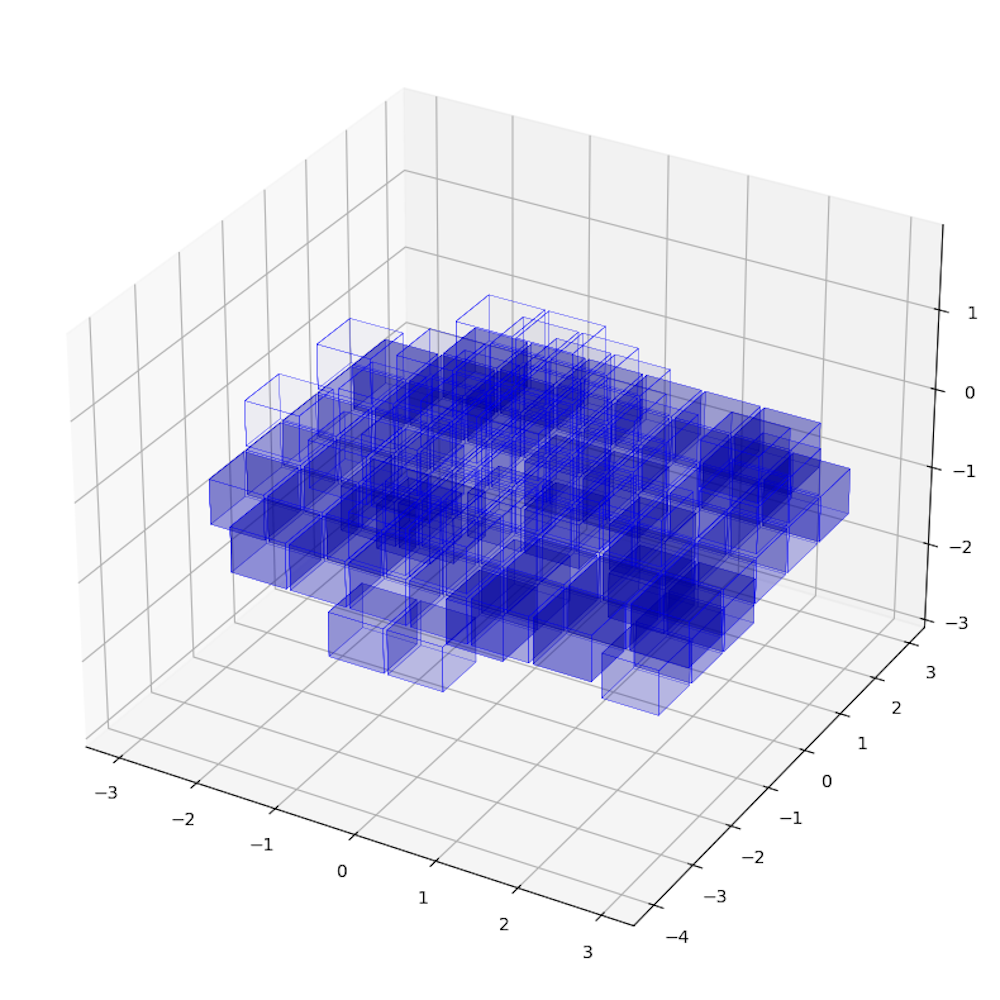}
	\end{minipage}} 
	\subfigure[init lv 4]{
		\begin{minipage}[t]{\scale\linewidth}
			\includegraphics[width=\scaleB\linewidth]{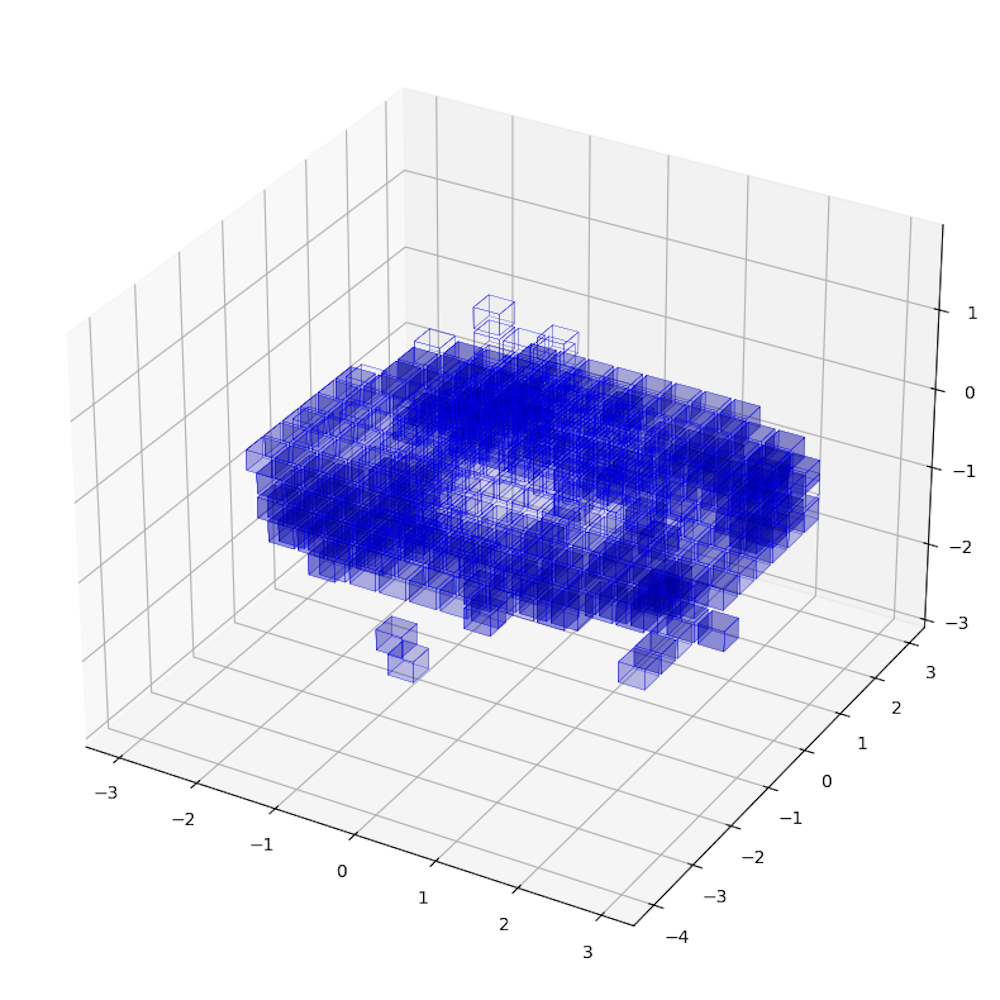}
	\end{minipage}} 
	
	\subfigure[deactivate]{
		\begin{minipage}[t]{\scale\linewidth}
			\includegraphics[width=\scaleB\linewidth]{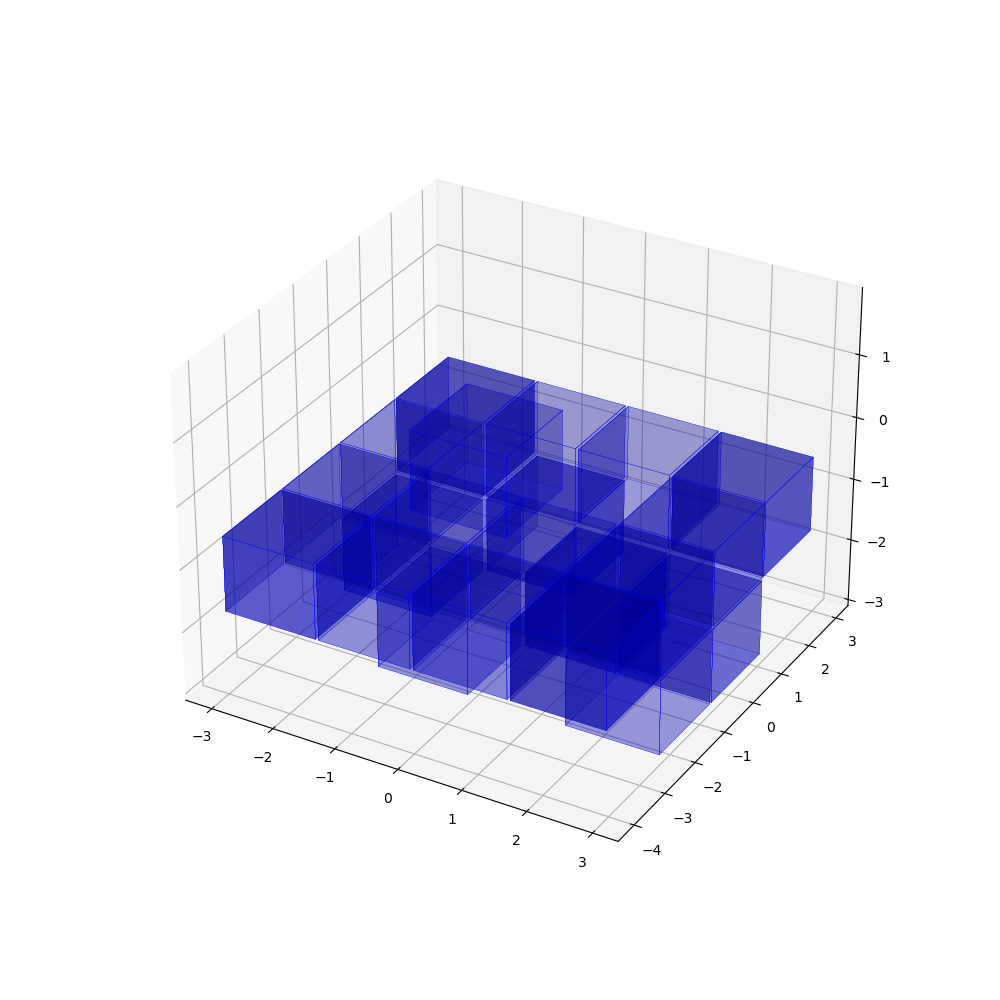}
	\end{minipage}}
	\subfigure[lv 2$\rightarrow$ lv 3]{
		\begin{minipage}[t]{\scale\linewidth}
			\includegraphics[width=\scaleB\linewidth]{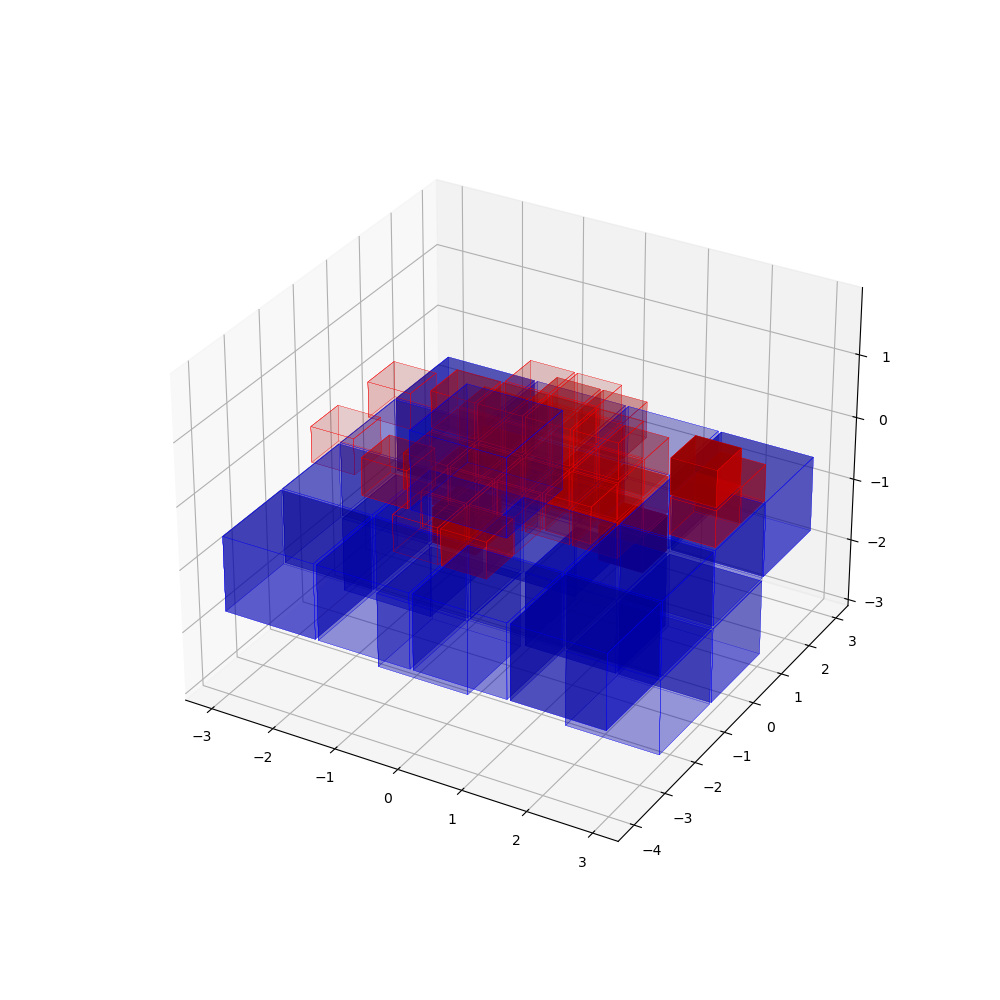}
	\end{minipage}}
	\subfigure[lv 3$\rightarrow$ lv 4]{
		\begin{minipage}[t]{\scale\linewidth}
			\includegraphics[width=\scaleB\linewidth]{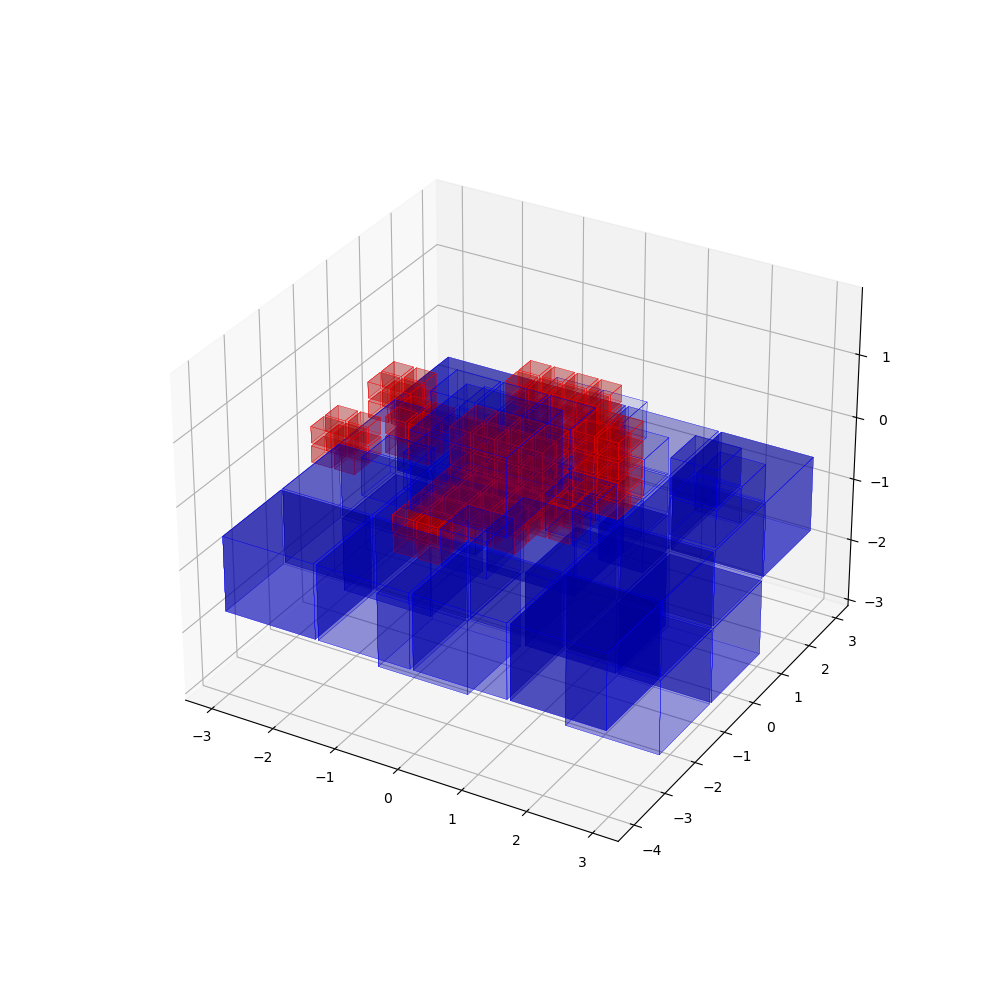}
	\end{minipage}}
	\subfigure[block merge]{
		\begin{minipage}[t]{\scale\linewidth}
			\includegraphics[width=\scaleB\linewidth]{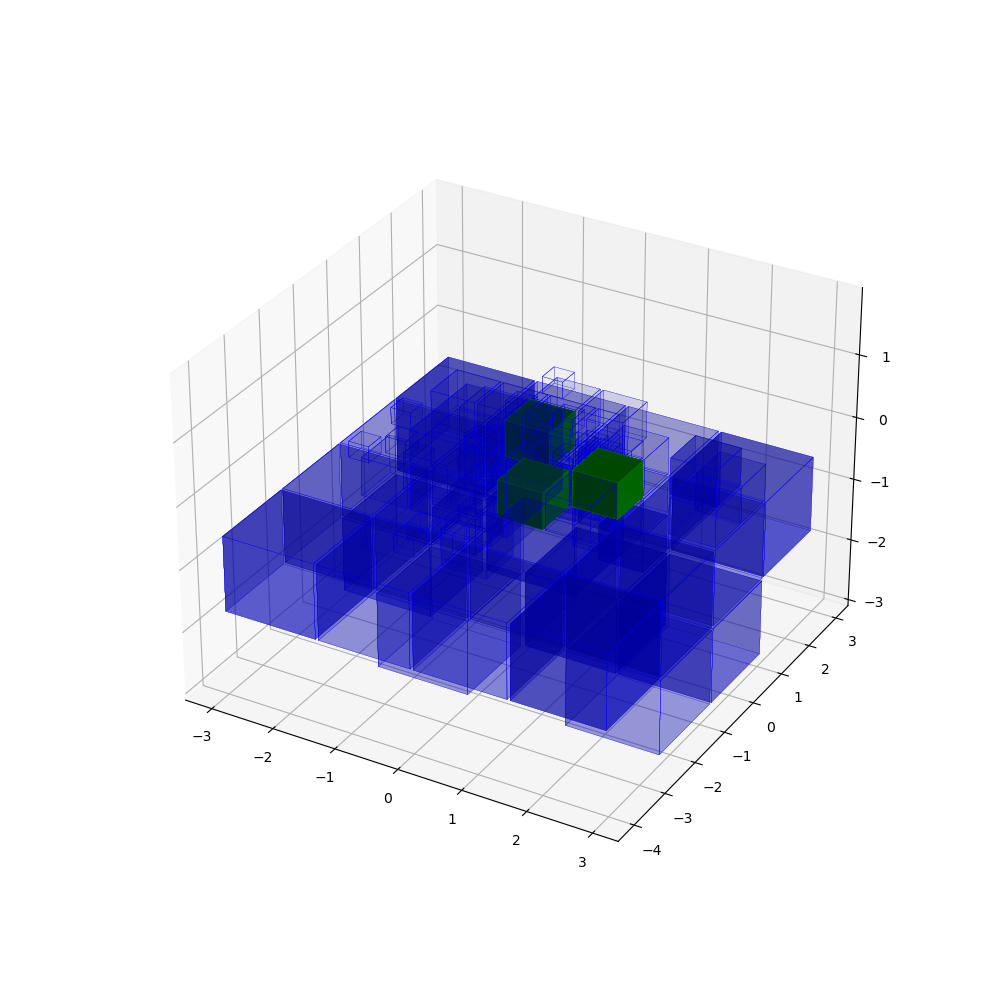}
	\end{minipage}} 
	\subfigure[view]{
		\begin{minipage}[t]{\scale\linewidth}
			\includegraphics[width=\scaleB\linewidth]{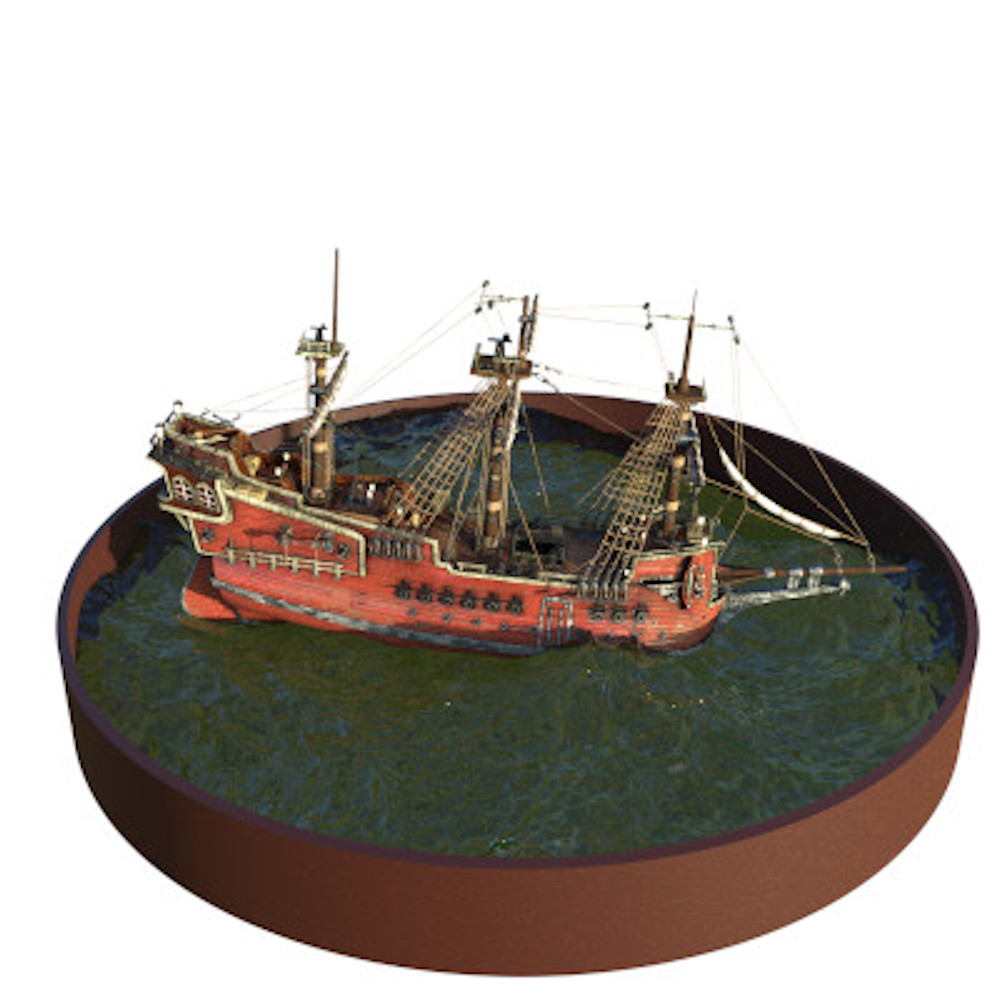}
	\end{minipage}} \\
	\caption{Octree structure update. (a) to (e) show the initial
          training by using a fully subdivided octree to a given
          level, with empty nodes culled. (f) only shows active and
          keep unchanged blocks in level 2, (g) shows block splitting
          for level 2 to level 3, (h) shows block splitting for level
          3 to level 4, (i) shows a block merge to prune the octree
          for simplification. }
	\label{fig:octree_opt}
\end{figure*}

The full model $\model$ consists not only of the neural networks in
the leaf nodes, but also of the octree structure itself. To optimize
this octree structure, we solve an optimization problem with a mixed
integer program, similar to the method proposed by
ACORN~\cite{martel2021acorn}. However, while ACORN is trained directly
from a known reference volume, the volume is initially unknown in our
inverse rendering setting. We therefore have to devise a different
cost function to decide which octree nodes should be subdivided,
merged or deactivated. 

%Our octree optimization procedure considers weighted average density
%$\alpha$ inside the node and projected rendering error $\beta$ in the
%node. 

%\begin{equation}\label{eqn:tree_opt}
%	\min \sum_{i} (1-\alpha^{\intercal}_i) I_i + \beta^{\intercal}_i I_i,
%\end{equation}

%where the cost function encourages active nodes to be non-empty (high
%average density) with low projected rendering error.

Specificially, our octree optimization procedure considers both the
weighted average density within each node, as well as the aggregated
reprojection error within each node.  If a weighted average density in
a block is less than a threshold (0.01), the block will simply be set
to inactive, and will not join the later computation. If a parent node
and child node are both active, our algorithm will choose the node
with smaller size, i.e. the child node has priority. Please refer to
the supplemental materials and the code for more details.
\figref{fig:octree_opt} illustrates the evolution of the octree
structure from initial levels to full octree optimization stage.  

\subsection{Model Updates by Pre-training}
\label{sec:pretrain}

Every time the octree structure changes, the networks for the old leaf
nodes are replaced with new networks for the new leafs. For example,
when a leaf node is subdivided, the corresponding network
$\mlp_i^\level$ is replaced by eight new networks
$\mlp_{i,1}^{\level+1},\dots,\mlp_{i,8}^{\level+1}$ responsible for
the different quadrants. Conversely, when nodes are merged, eight
networks at level $\level$ get replaced by a single network at level
$\level-1$.

After such structural changes, we directly pre-train the new
network(s) using stratified samples from the previous network(s). This
allows the model to quickly return to a similar quality than before
the structure change without the need for costly ray-tracing and
compositing operations.
After this pre-training, the normal ray-tracing-based training
resumes.

\section{Experiments}

For evaluation and both qualitative and quantitative comparison
against state-of-the-art methods, we apply our method to several
publicly available datasets that have been used by competing methods
before, e.g. \textbf{Synthetic-NeRF}~\cite{mildenhall2020nerf},
\textbf{LLFF-NeRF}~\cite{mildenhall2020nerf}, \textbf{DTU Robot Image
  Data Sets}~\cite{jensen2014large}.
We also conduct extensive ablation studies for various parameter
choices, e.g., sub-network architecture and the number of block
levels.
In addition to the results in this document, we also refer to the
supplemental material for more results.

\subsection{UAV-view Terrain Scanning and Reconstruction}
In addition to existing standard datasets we also introduce a new
UAV-based scene. UAV remote sensing data has usually much sparser view
points, with little overlap between neighboring views. Moreover, the
standoff distance is often large compared to the scene scale, so that
parallax is limited. 

%To demonstrate the ability to reconstruct challenging remote sensing
%scenes from UAV-based views, we start by comparing \nascent to comparison
%methods on a new UAV dataset containing images with a large standoff
%distance, sparse viewpoints, low overlap between views, as well as
%sparse volume occupancy that tests the methods ability to adapt the
%sampling process.

This setting is quite challenging for previous neural rendering
methods were mainly designed for rendering dense viewpoints with
similar camera viewing angles and highly overlapping scene content,
and then represent scene by single network~\cite{mildenhall2020nerf},
\cite{martel2021acorn}, \cite{Lin_2021_ICCV} or multiple
sub-networks~\cite{Reiser2021ICCV}. However, a non-adaptive single
network structure will have representation capacity problems for
training and rendering a large unbounded scene, multiple
sub-network~\cite{Reiser2021ICCV} will also require a pre-trained
single network for better initial performance.  Our method contains
the optimization of octree structure and sub-network training, thus,
the network in each block is only handling representation and
reconstruction tasks locally, and could also scale to larger scenes if
needed.

%% This is repetitive...!
%To capture dataset, we used a UAV to take photo of a  ground scene and apply our methods, visual comparison and real scene rendering results are shown in \figref{fig:uav_kaust}.
%Note that previous neural rendering methods~\cite{mildenhall2020nerf, Reiser2021ICCV, liu2020neural}, require dense viewpoint sampling with comparatively near and fixed viewing distance. thus, has limitation to represent scene with large viewpoint change and may suffer network capacity problem to represent large scene. Specifically, taking photo by a flying UAV and render novel viewpoint or reconstruct large 3D scene is a promising applications for neural rendering, however, due to the limitation of single network capacity and highly dynamic ray marching distance, directly training and rendering neural radiance field by single network or sampling point globally is a time-consuming and challenging task. 
Our proposed method is scalable and represents scene content by
multiple networks in an octree structure. Therefore the overall
representational capacity of the model depends on both the number of
octree cells as well as the number of parameters in the networks. Both
of these are hyper parameters that we analyze in detail
below. However, even very lightweight per-node networks are capable of
producing higher quality representations compared to competing
approaches.

\begin{figure*}
	\def \scale {0.18}
	\def \scaleB {1.0}
	\centering
	\subfigure[GT]{
		\begin{minipage}[t]{\scale\linewidth}
			\includegraphics[width=\scaleB\linewidth]{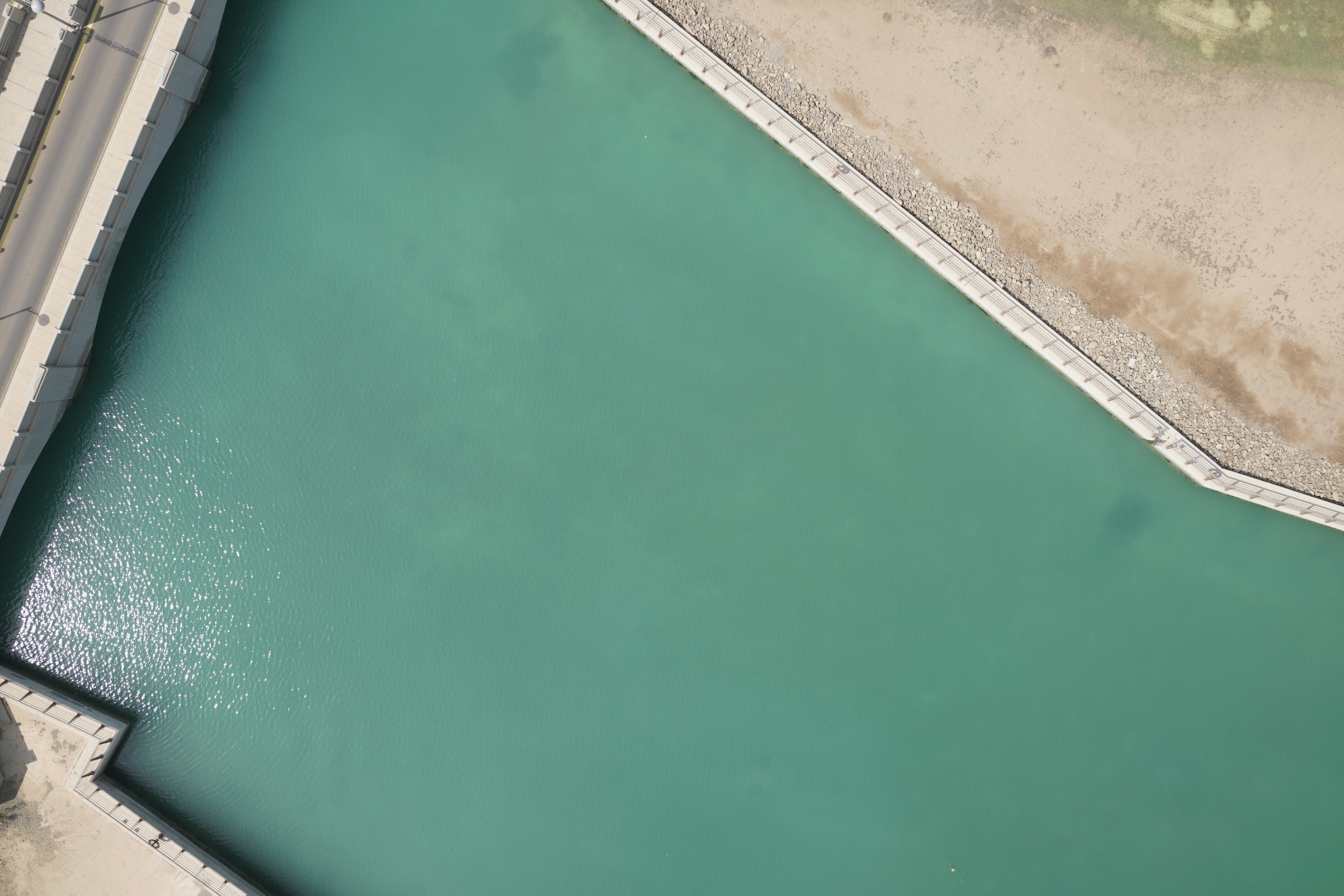}
			\color{white}
			\includegraphics[draft, width=\scaleB\linewidth]{figure/uav/kaust_s03/gt/Island_Mosque_Flight_01_00449}
			\includegraphics[width=\scaleB\linewidth]{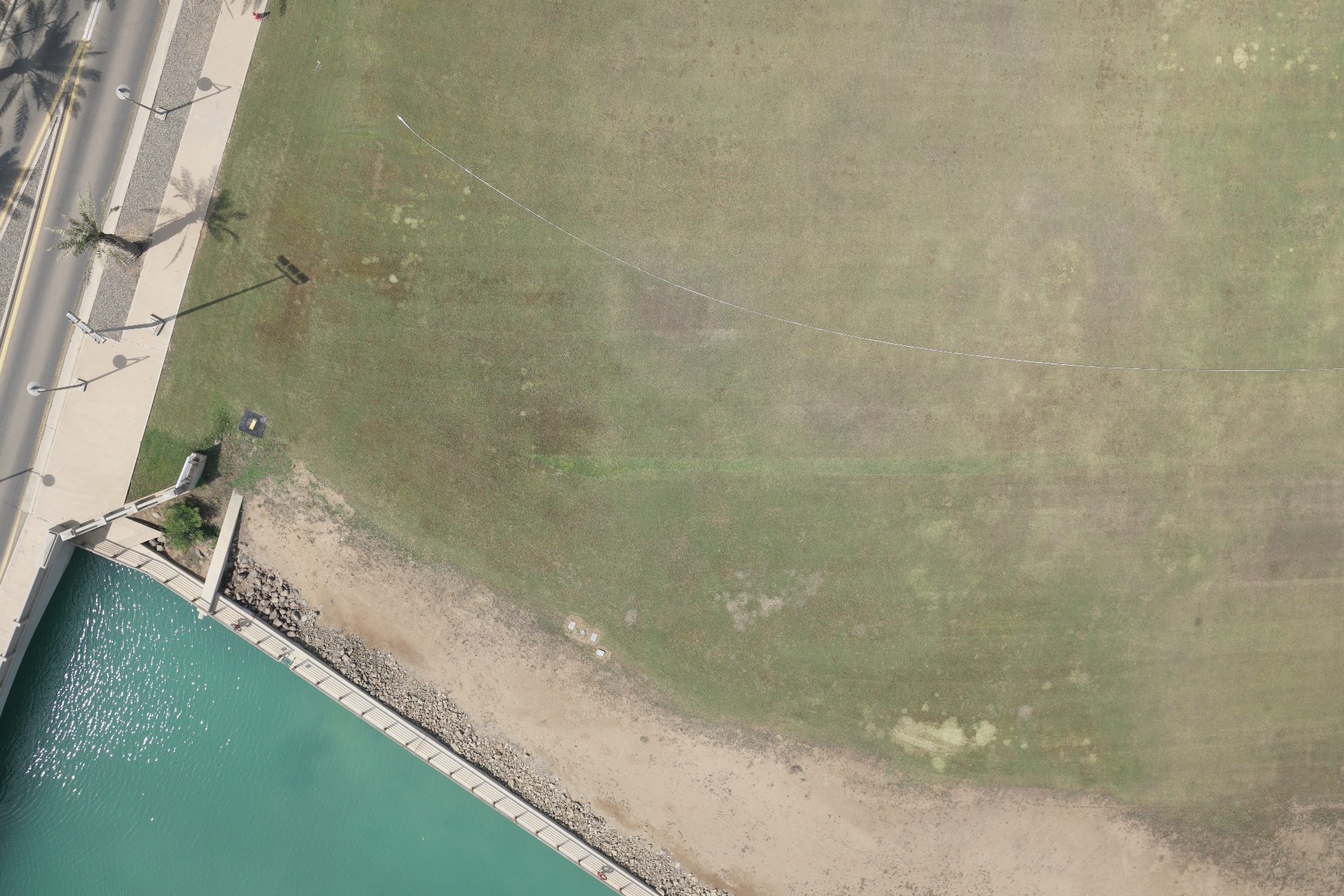}
			\color{white}
			\includegraphics[draft, width=\scaleB\linewidth]{figure/uav/kaust_s03/gt/Island_Mosque_Flight_01_00454}
	\end{minipage}}
	\subfigure[NeRF]{
		\begin{minipage}[t]{\scale\linewidth}
			\includegraphics[width=\scaleB\linewidth]{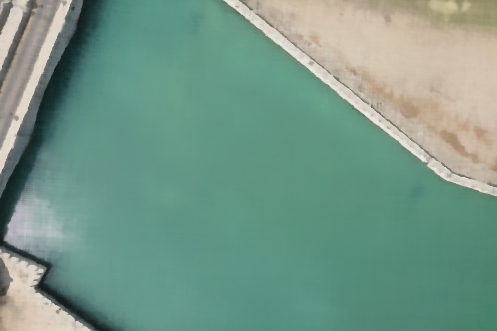}
			\includegraphics[width=\scaleB\linewidth]{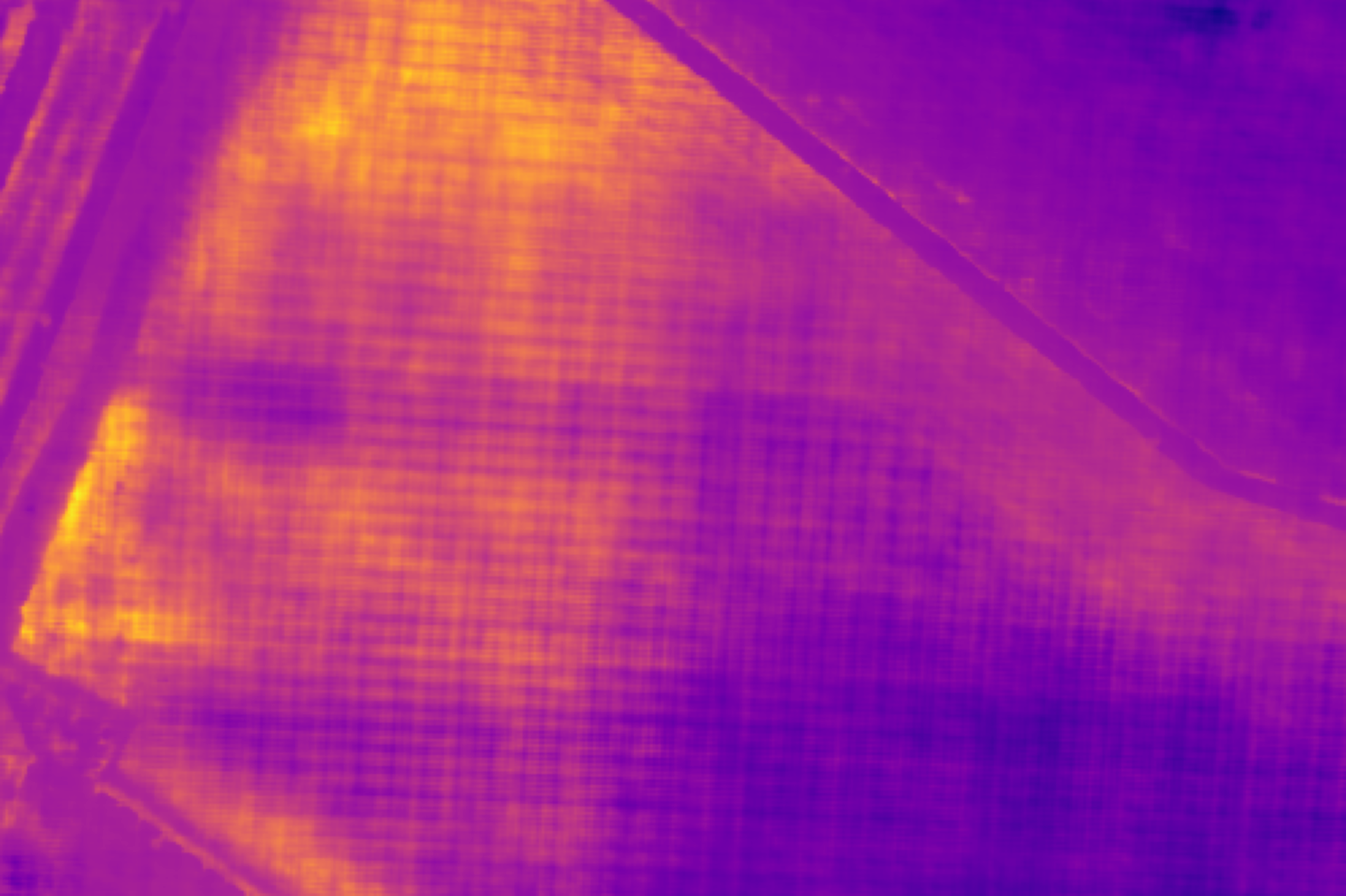}
			\includegraphics[width=\scaleB\linewidth]{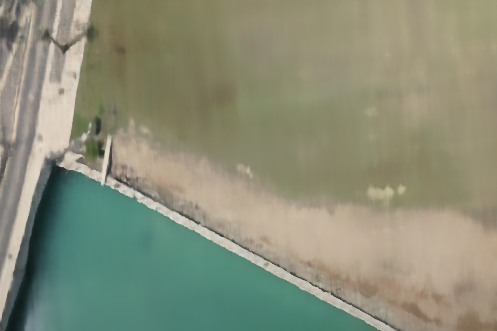}
			\includegraphics[width=\scaleB\linewidth]{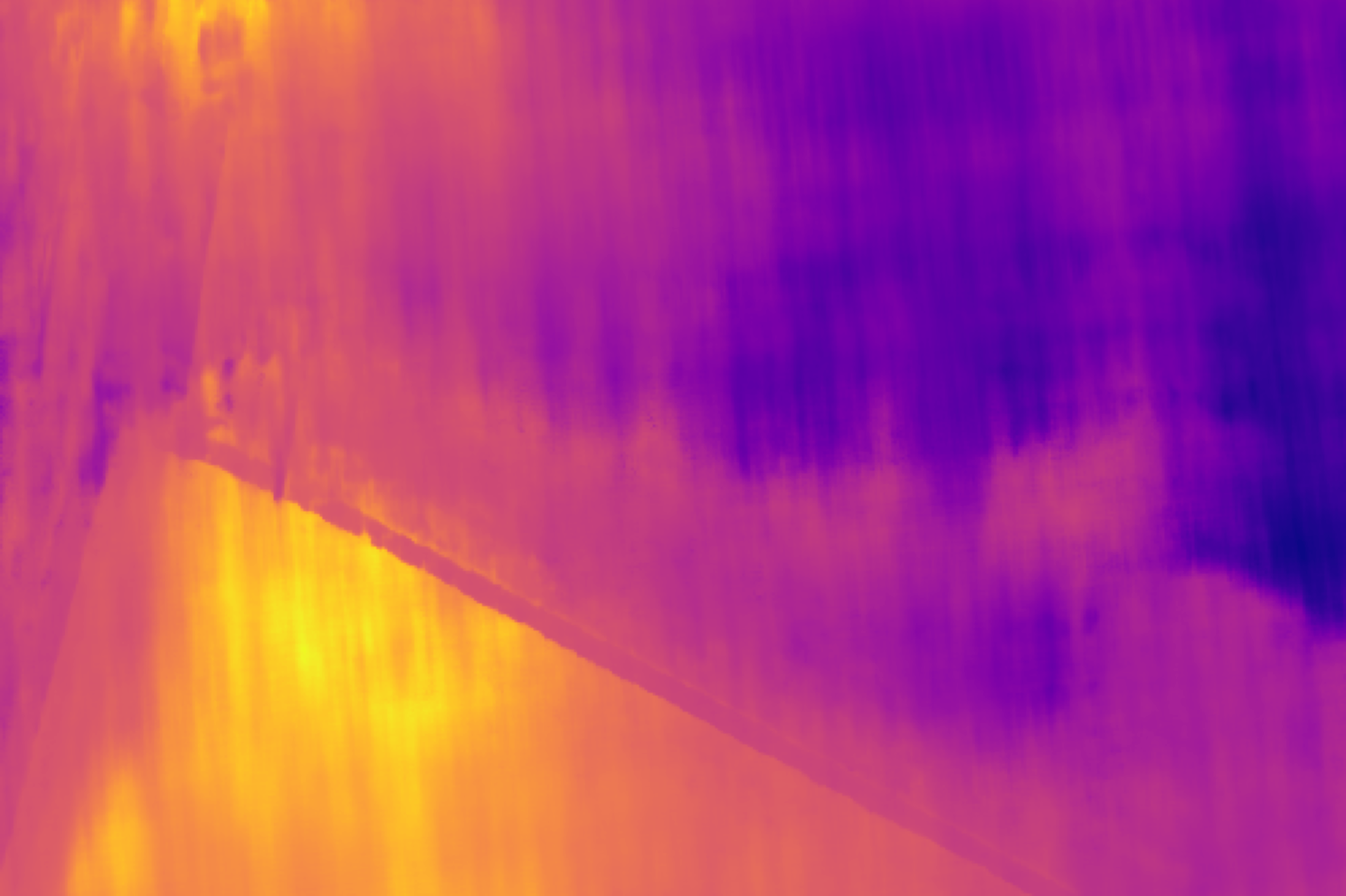}
	\end{minipage}}
	\subfigure[KiloNeRF]{
		\begin{minipage}[t]{\scale\linewidth}
			\includegraphics[width=\scaleB\linewidth]{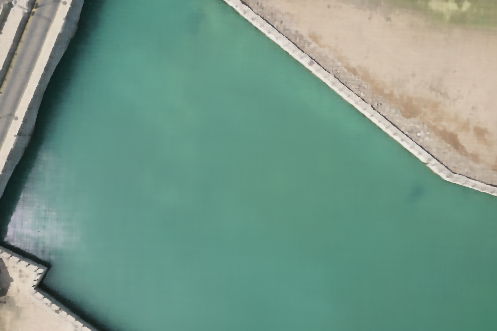}
			\includegraphics[width=\scaleB\linewidth]{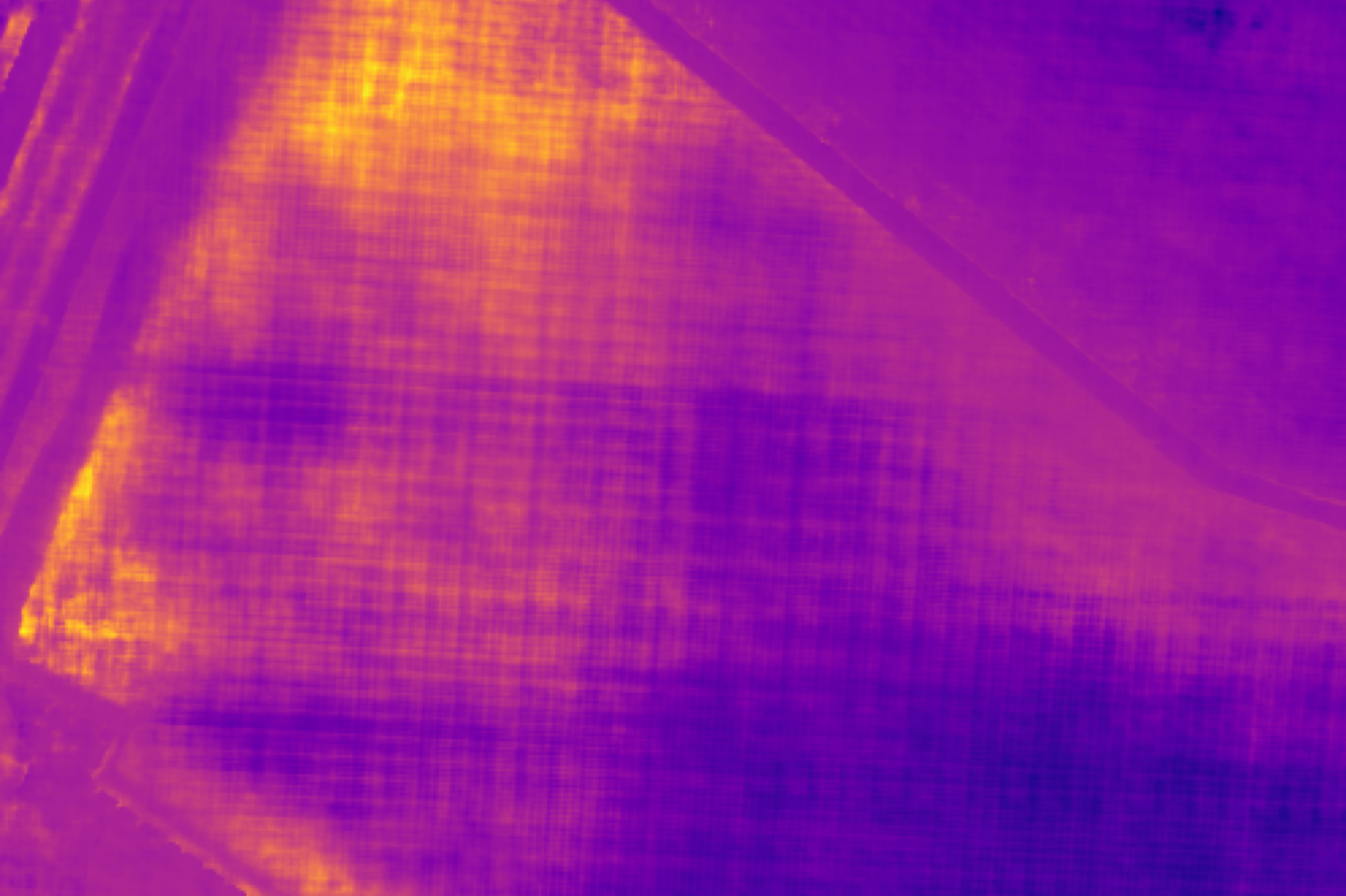}
			\includegraphics[width=\scaleB\linewidth]{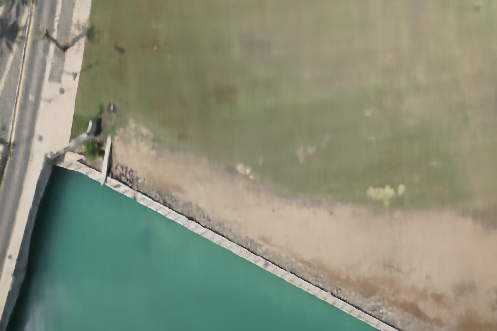}
			\includegraphics[width=\scaleB\linewidth]{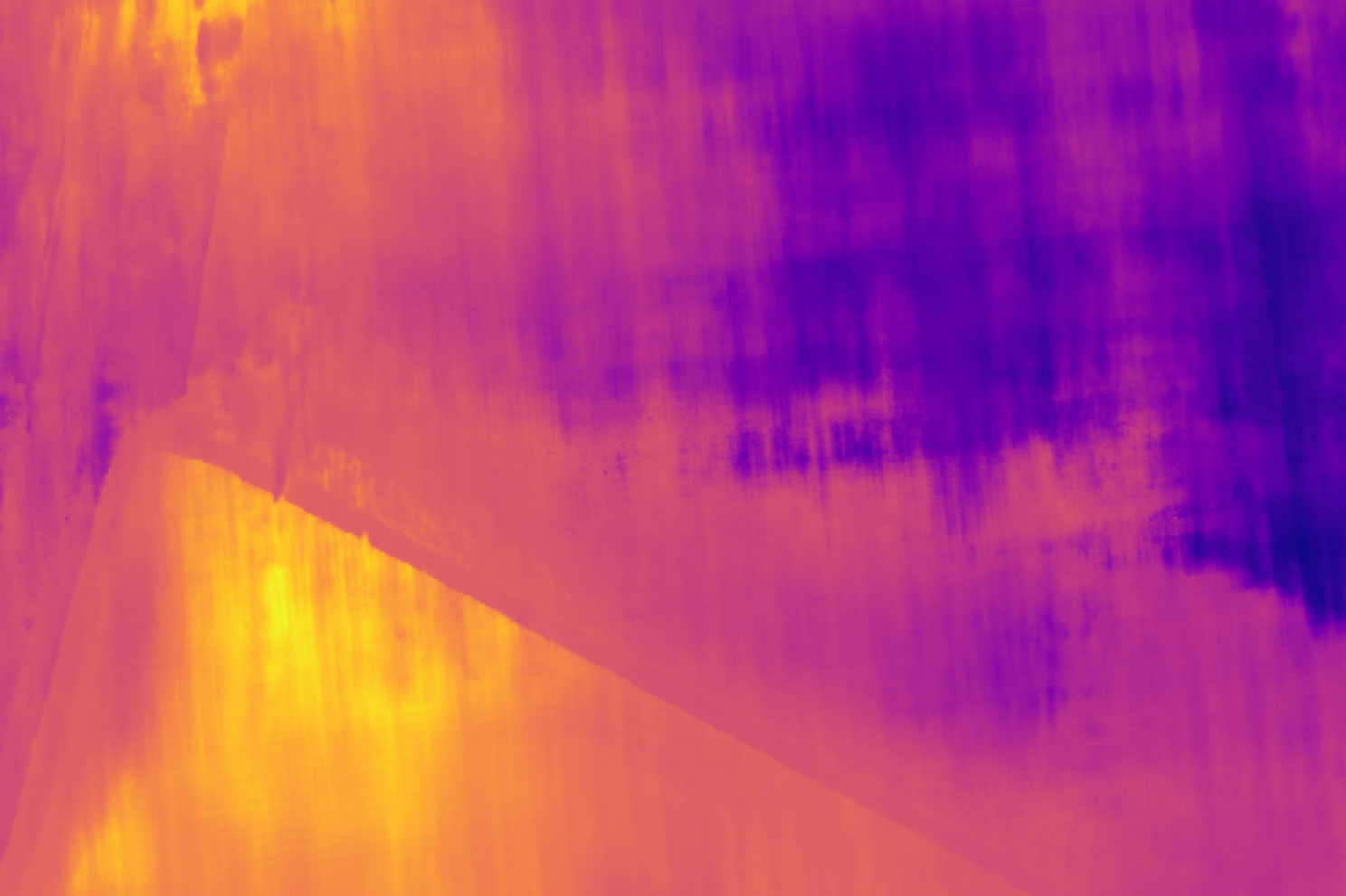}
	\end{minipage}}
	\subfigure[MipNeRF]{
		\begin{minipage}[t]{\scale\linewidth}
			\includegraphics[width=\scaleB\linewidth]{figure/uav/kaust_s03/kilo/002}
			\includegraphics[width=\scaleB\linewidth]{figure/uav/kaust_s03/kilo/002_depth_depth_plasma}
			\includegraphics[width=\scaleB\linewidth]{figure/uav/kaust_s03/kilo/005}
			\includegraphics[width=\scaleB\linewidth]{figure/uav/kaust_s03/kilo/005_depth_depth_plasma}
	\end{minipage}}
	\subfigure[our]{
		\begin{minipage}[t]{\scale\linewidth}
			\includegraphics[width=\scaleB\linewidth]{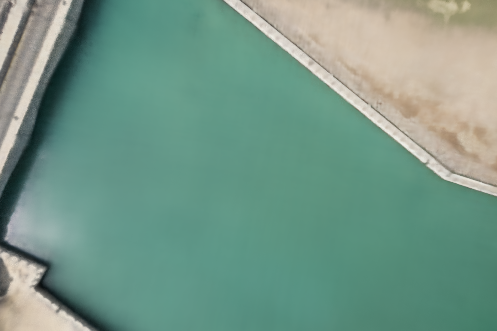}
			\includegraphics[width=\scaleB\linewidth]{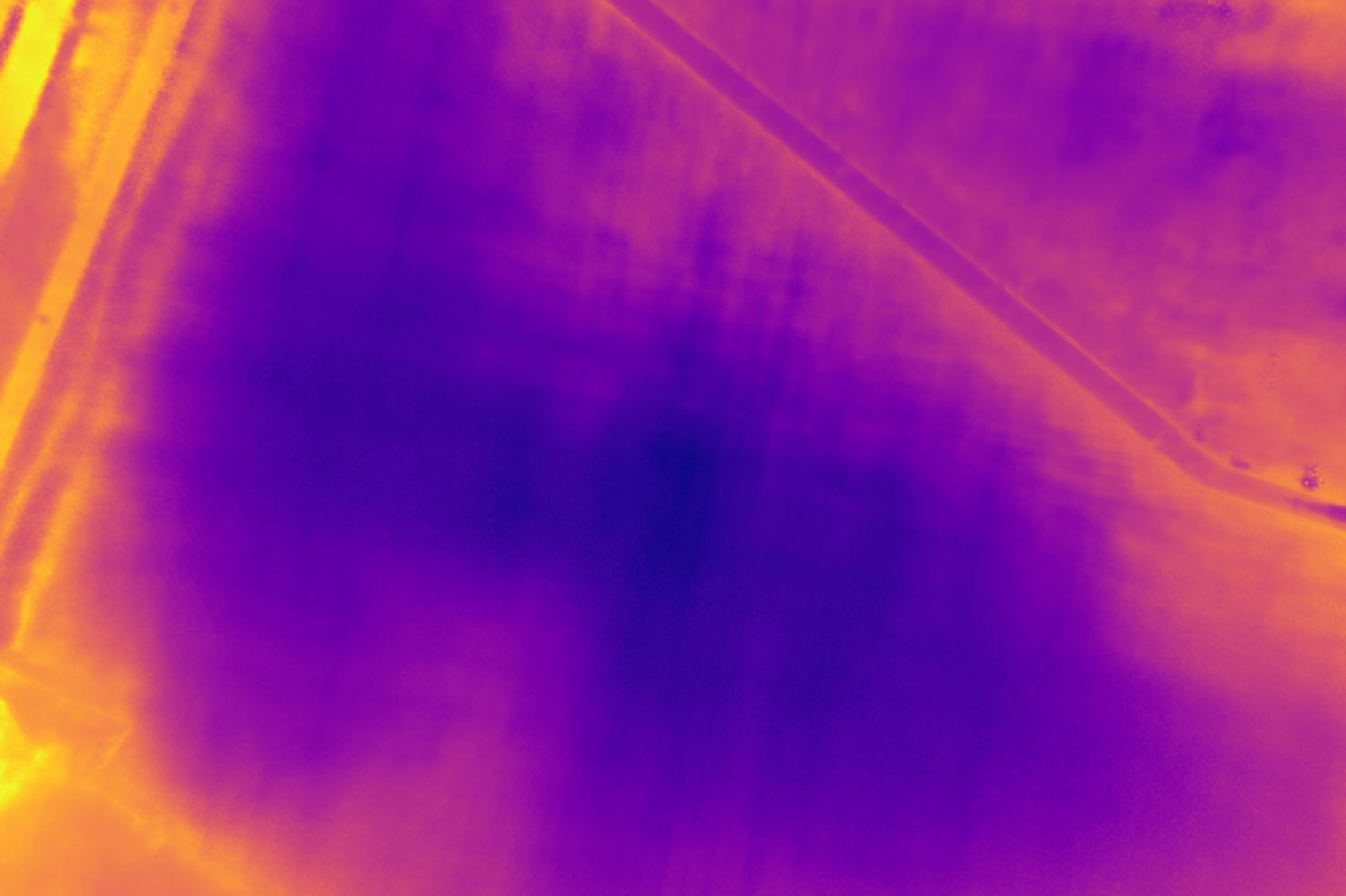}
			\includegraphics[width=\scaleB\linewidth]{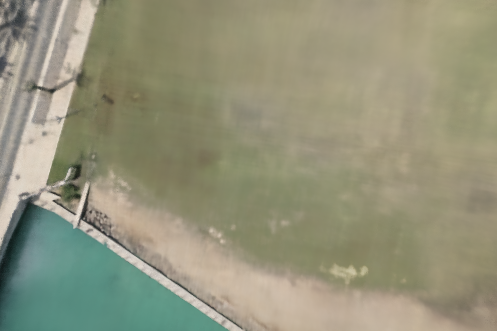}
			\includegraphics[width=\scaleB\linewidth]{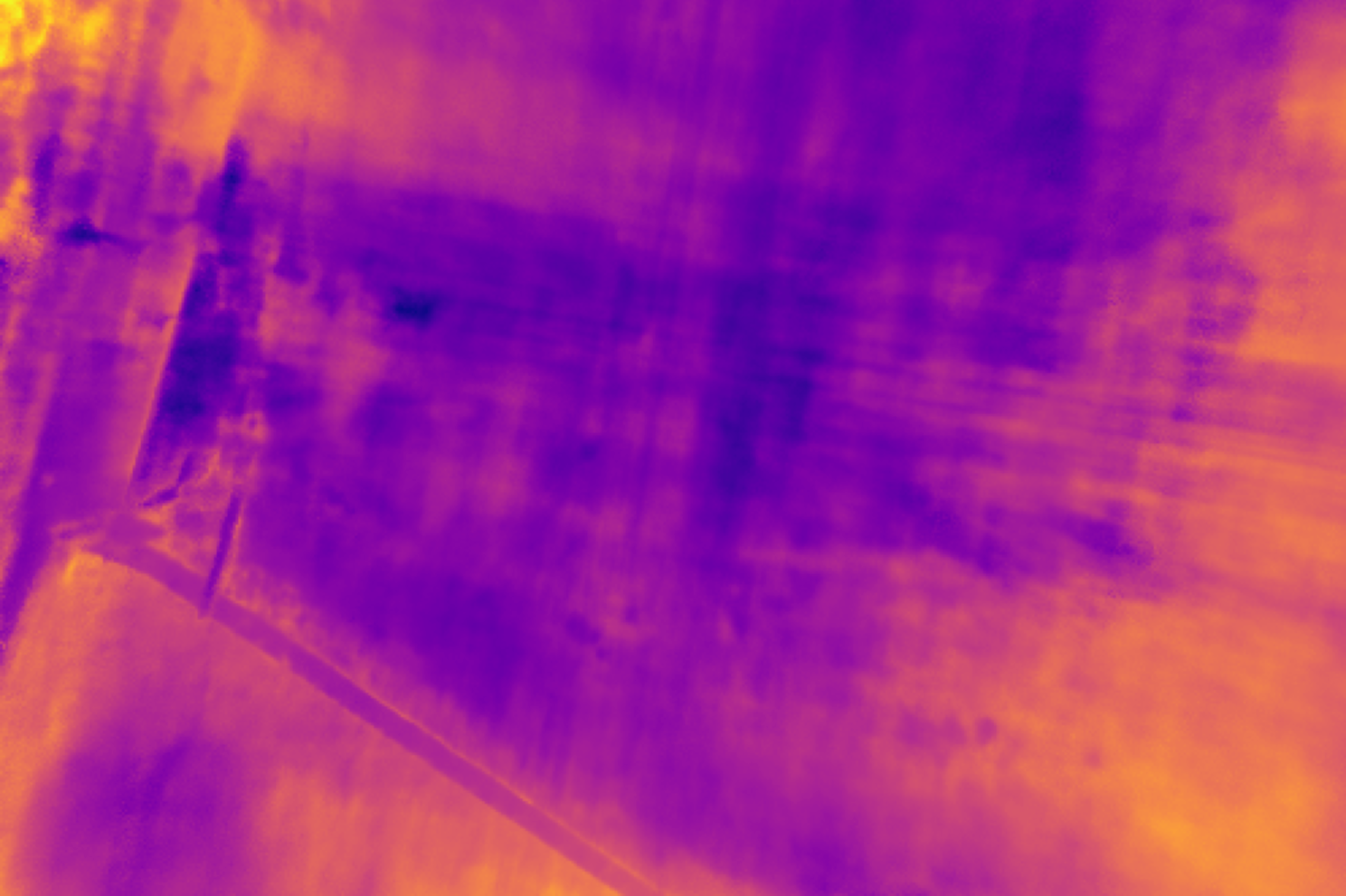}
	\end{minipage}}
	\caption{Ground scene reconstruction from UAV data
          (re-rendering from novel view point and false-color
          rendition of the reconstructed depth map). Note the
          improved detail in our depth map compared to both NeRF and
          KiloNeRF, which indicates better leaning of the 3D scene
          density and also results in better detail preservation in
          the re-rendering. We compare our method against NeRF~\cite{mildenhall2020nerf}, KiloNeRF~\cite{Reiser2021ICCV}, MipNeRF~\cite{barron2021mipnerf}.}
	\label{fig:uav_kaust}
\end{figure*}

\subsection{Visual Comparison on Public Datasets}
We demonstrate the performance of our method by rendering novel views
of synthetic and real scene dataset~\cite{mildenhall2020nerf} by
visualizing novel views in test set as well as the rendered depth map
of the scene. Visually, it is difficult to see differences between any
of the recent methods for {\bf view interpolation} -- camera positions
close to the training positions. However, differences become apparent
for {\bf view extrapolation}, where the novel camera position is far
from any of the input cameras. In this document we therefore focus on
this view extrapolation scenario for the visual results; the
supplemental material has more results.

For comparison methods, we choose those neural rendering methods that
can support both sperical and front view scene rendering, including
NeRF~\cite{mildenhall2020nerf}, KiloNeRF~\cite{Reiser2021ICCV} and
MipNeRF~\cite{barron2021mipnerf}.

\begin{figure*}[ht]
	\def \scale {0.22}
	\def \scaleB {0.9}
	\centering
	\subfigure[NeRF~\cite{mildenhall2020nerf}]{
		\begin{minipage}[t]{\scale\linewidth}
			\includegraphics[width=\scaleB\linewidth]{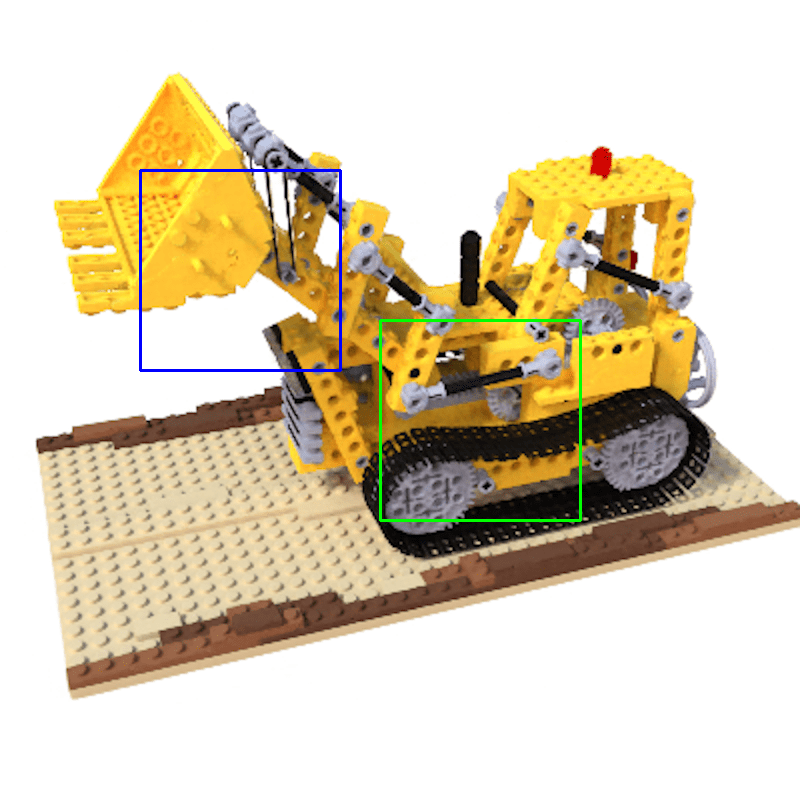}
			\includegraphics[width=\scaleB\linewidth]{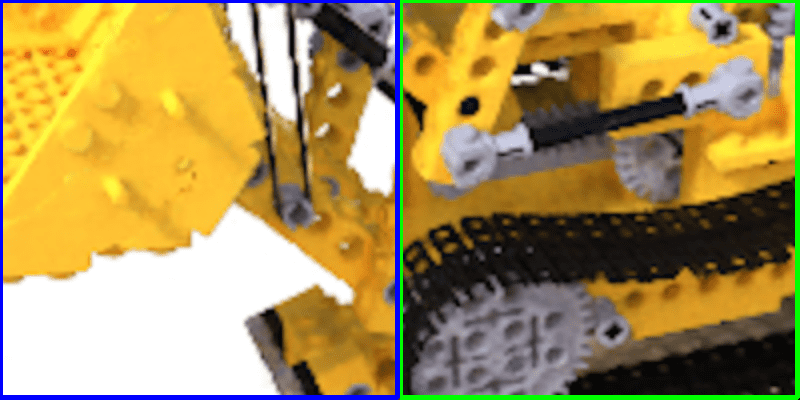}
			\includegraphics[width=\scaleB\linewidth]{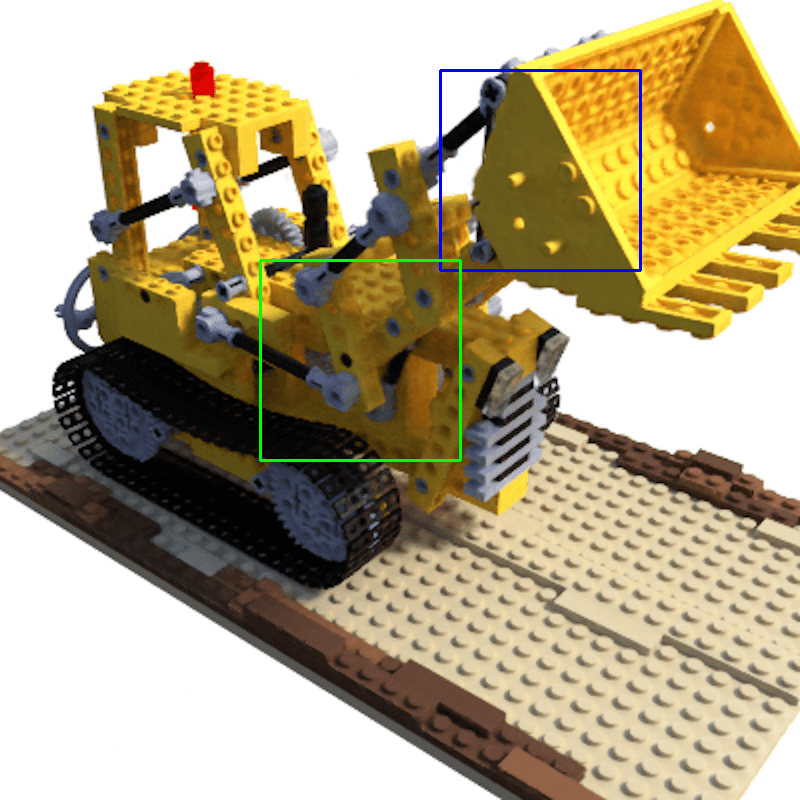}
			\includegraphics[width=\scaleB\linewidth]{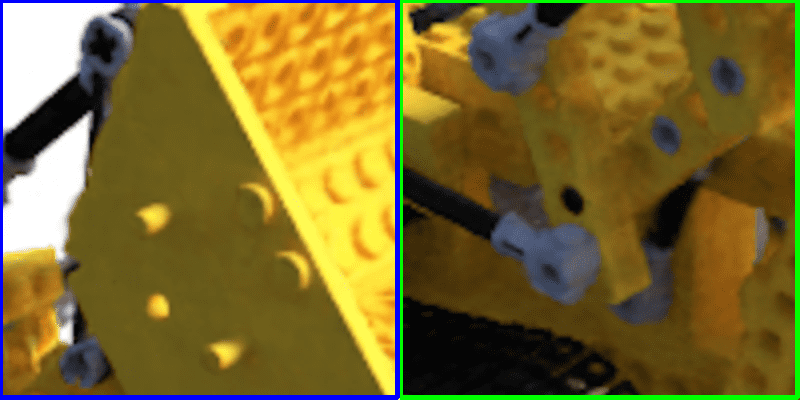}
	\end{minipage}}
	\subfigure[KiloNeRF~\cite{Reiser2021ICCV}]{
		\begin{minipage}[t]{\scale\linewidth}
			\includegraphics[width=\scaleB\linewidth]{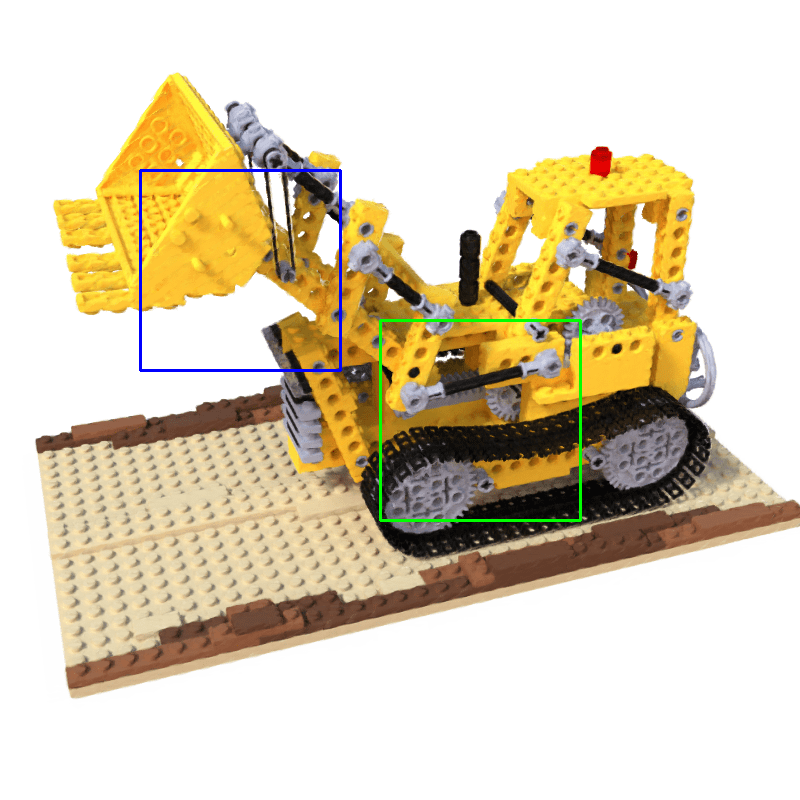}
			\includegraphics[width=\scaleB\linewidth]{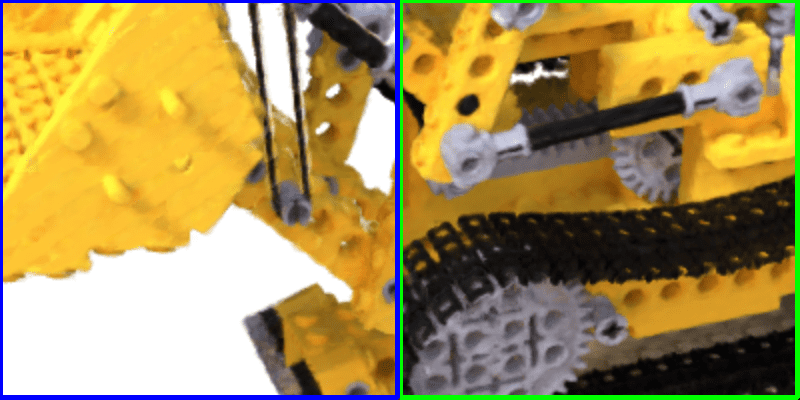}
			\includegraphics[width=\scaleB\linewidth]{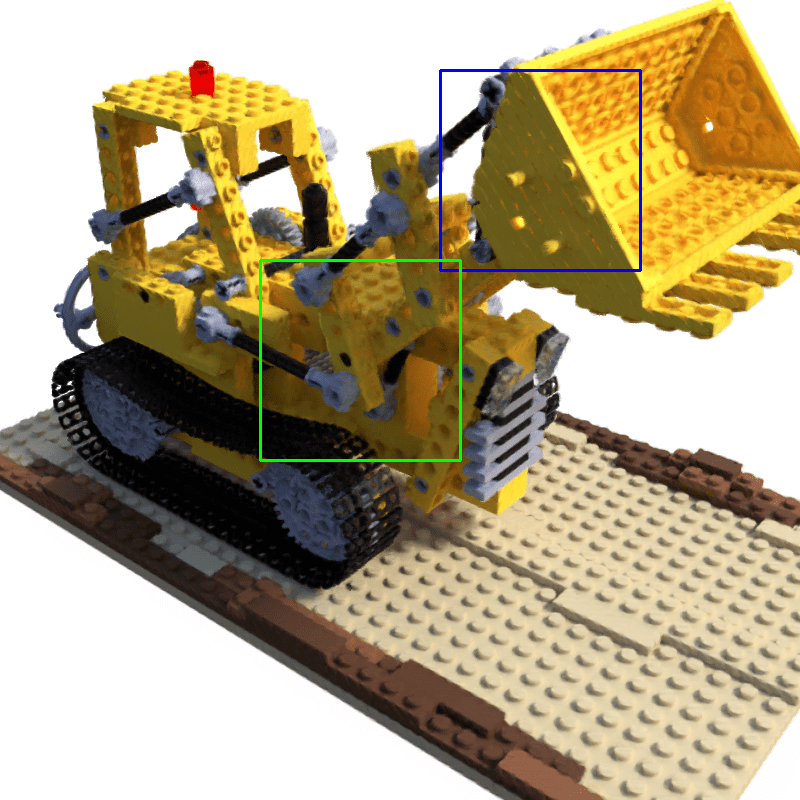}
			\includegraphics[width=\scaleB\linewidth]{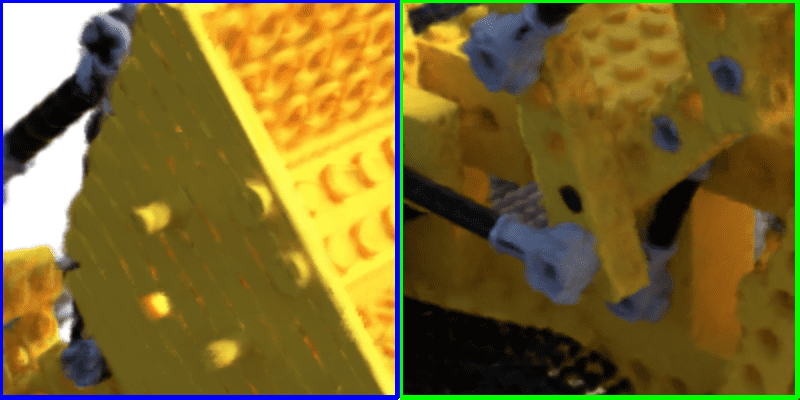}
	\end{minipage}}
	\subfigure[MipNeRF~\cite{barron2021mipnerf}]{
		\begin{minipage}[t]{\scale\linewidth}
			\includegraphics[width=\scaleB\linewidth]{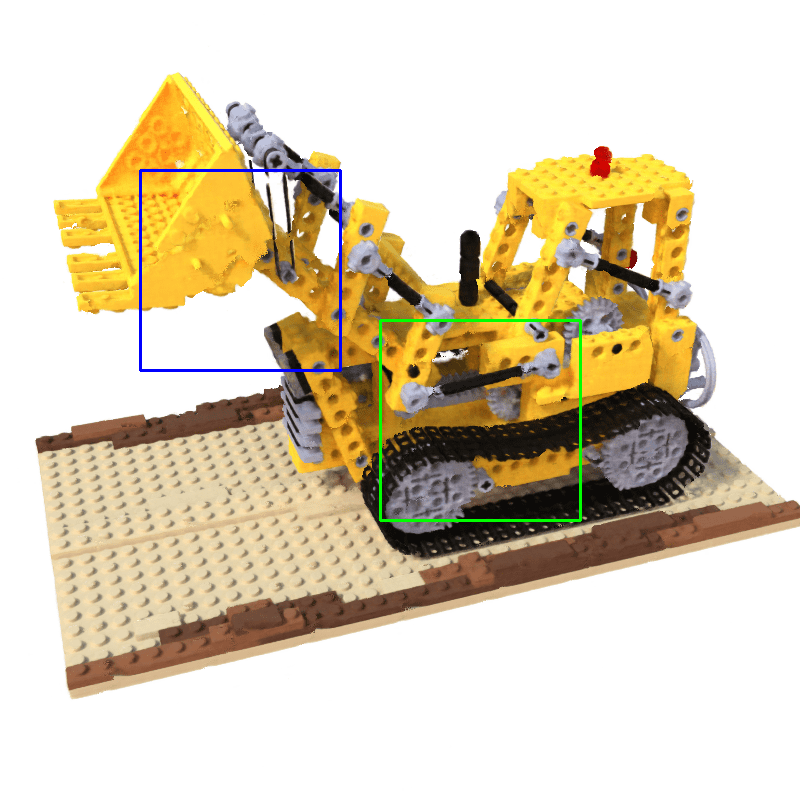}
			\includegraphics[width=\scaleB\linewidth]{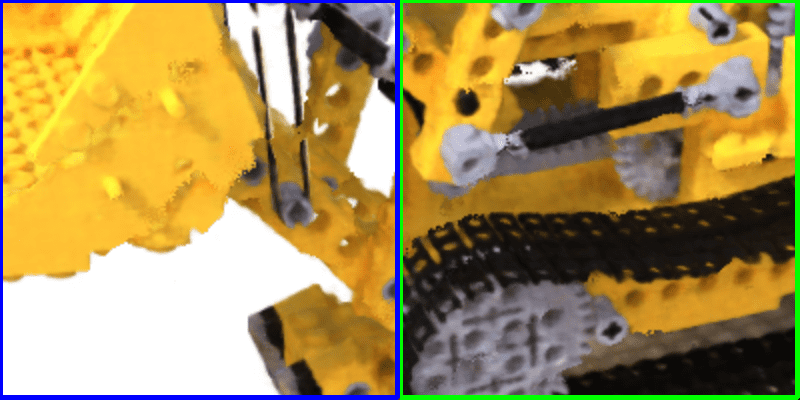}
			\includegraphics[width=\scaleB\linewidth]{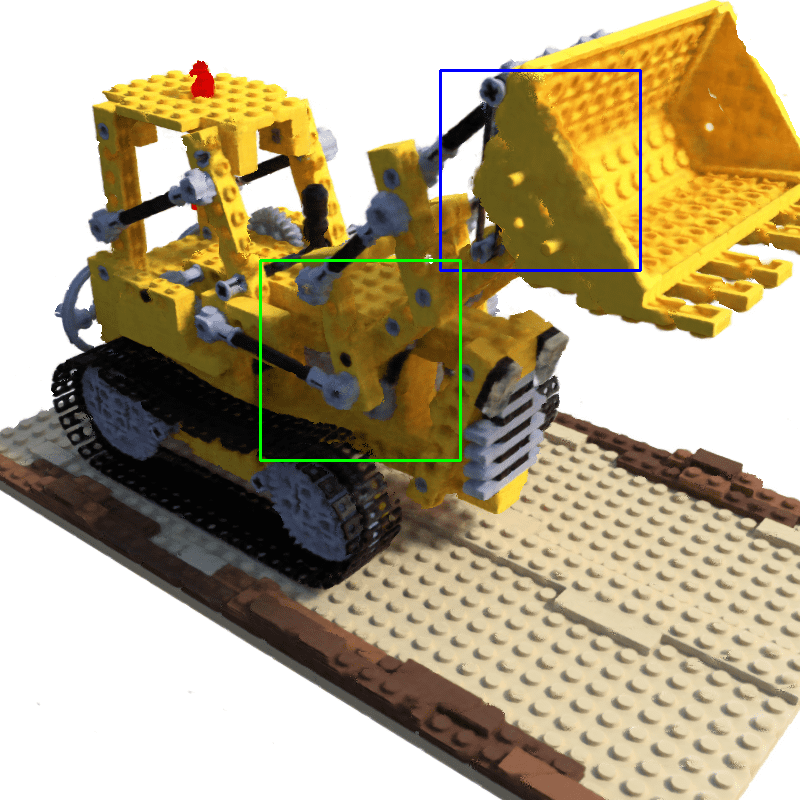}
			\includegraphics[width=\scaleB\linewidth]{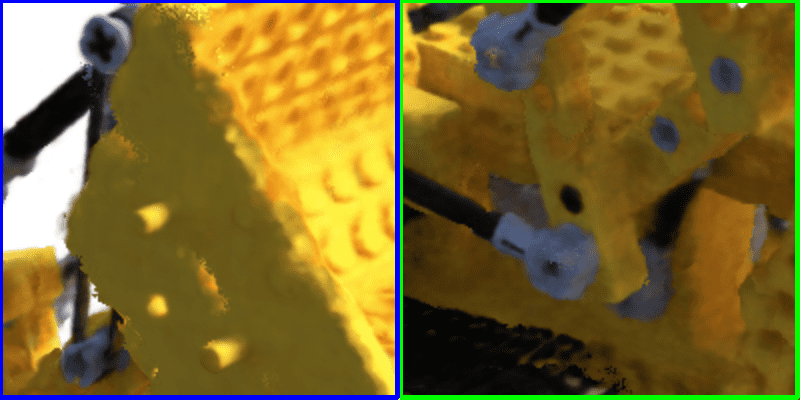}
	\end{minipage}}
	\subfigure[Our]{
		\begin{minipage}[t]{\scale\linewidth}
			\includegraphics[width=\scaleB\linewidth]{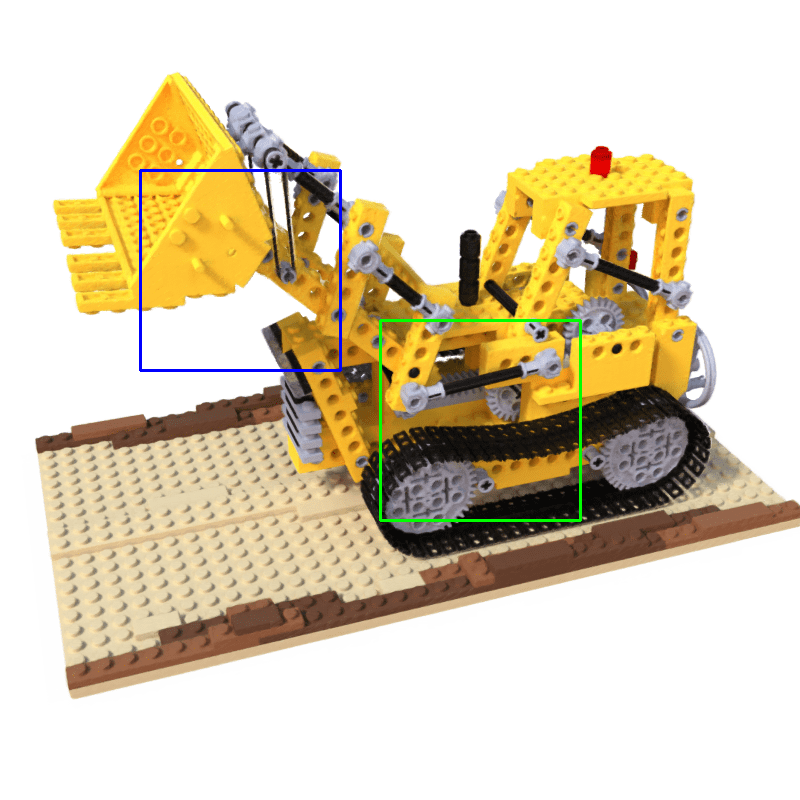}
			\includegraphics[width=\scaleB\linewidth]{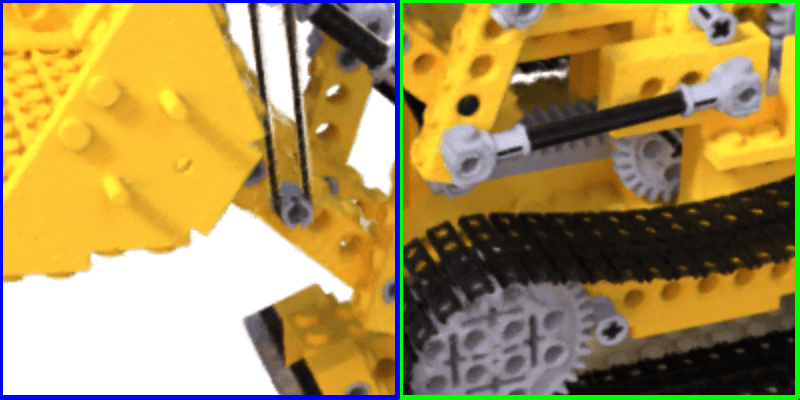}
			\includegraphics[width=\scaleB\linewidth]{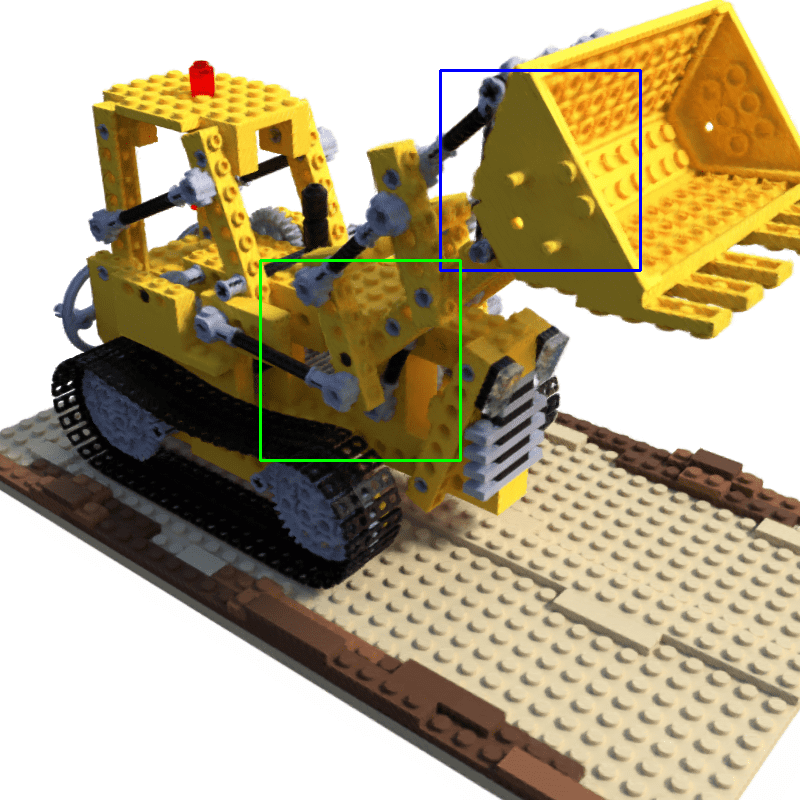}
			\includegraphics[width=\scaleB\linewidth]{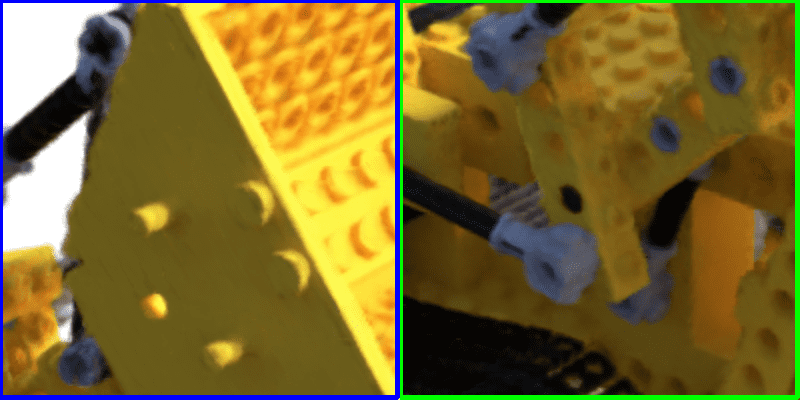}
	\end{minipage}}
	\caption{Novel View Comparison on Synthetic Dataset~\cite{mildenhall2020nerf}. We render viewpoints from near to far for visualizing viewpoint change and the influence of geometry in rendering.}
	\label{fig:extrem_view}
\end{figure*}

\paragraph{Synthetic Dataset} 
\figref{fig:extrem_view} visualizes results for extrapolated
viewpoints on the synthetic Lego model.
NeRF~\cite{mildenhall2020nerf} tend to produce slightly patchy colors
in flat areas since incorrect geometry exists in the density field.
Also, a single large model is computationally expensive, and therefore
limits the number of samples for a ray. KiloNeRF~\cite{Reiser2021ICCV}
uses NeRF's model as a teacher to learn a set of small networks for a
space partitioning into a regular grid, with the goal of improving the
inference (rendering) efficiency and enabling better sampling
rates. However, the networks for the individual grid cells are not
consistent at cell boundaries, and so light leaks can easily happen in
a novel views of the scene, since all the samples along the ray
contain zero density for a true surface. Moreover, KiloNeRF also
inherits defects from the original NeRF model. MipNeRF has issues at
depth discontinuities for these extreme view points, which also
indicates that it did not learn an accurate 3D representation.
Our method trains the composite model from scratch and enables
efficient rendering while avoiding the artifacts of the comparison
methods.

\begin{figure*}[h]
	\def \scale {0.21}
	\def \scaleB {1.0}
	\centering
	\subfigure[NeRF~\cite{mildenhall2020nerf}]{
		\begin{minipage}[t]{\scale\linewidth}
			\includegraphics[width=\scaleB\linewidth]{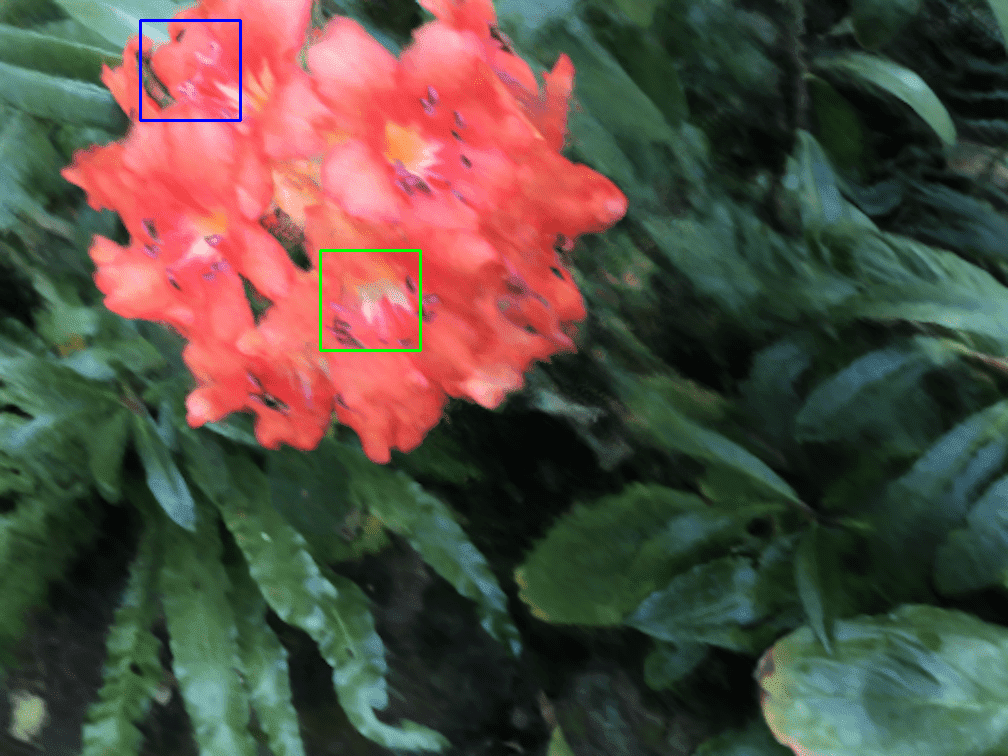}
			\includegraphics[width=\scaleB\linewidth]{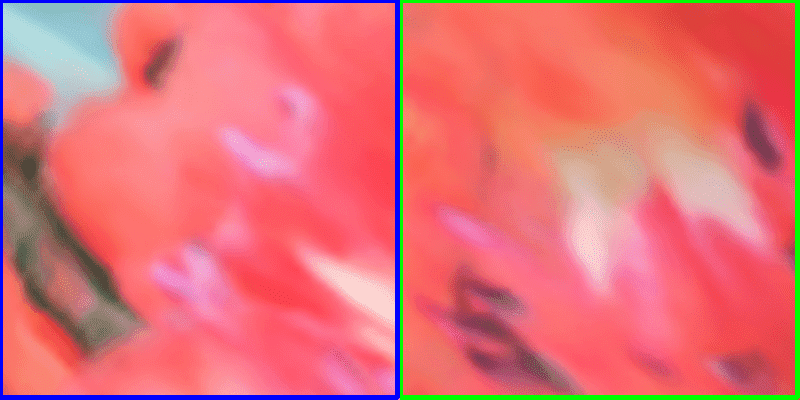}
			\includegraphics[width=\scaleB\linewidth]{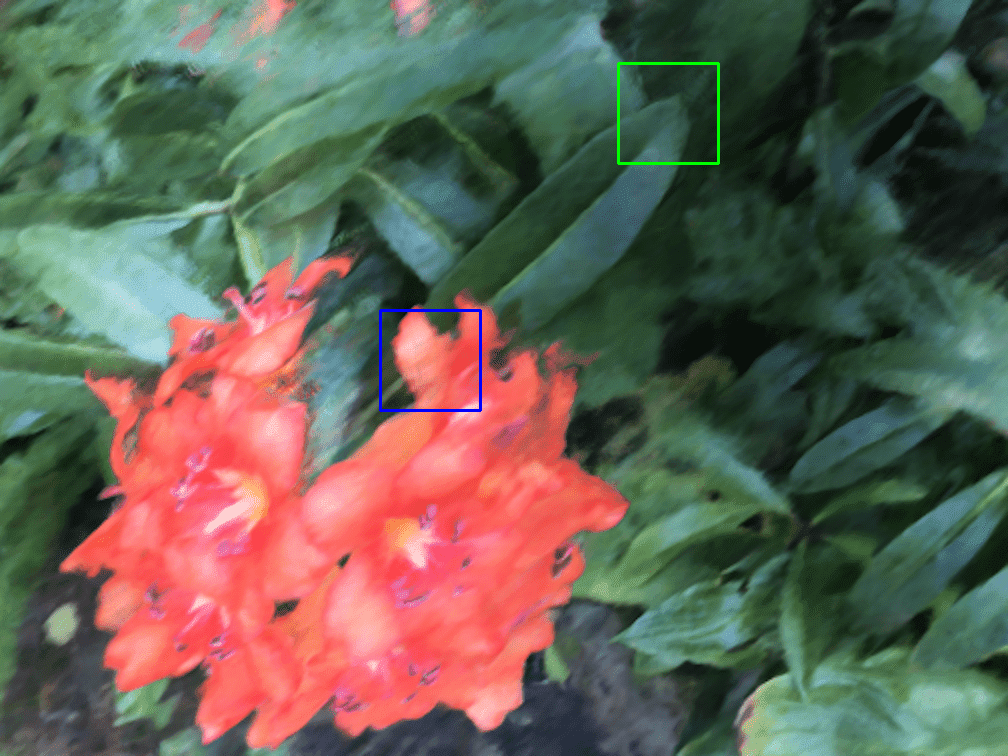}
			\includegraphics[width=\scaleB\linewidth]{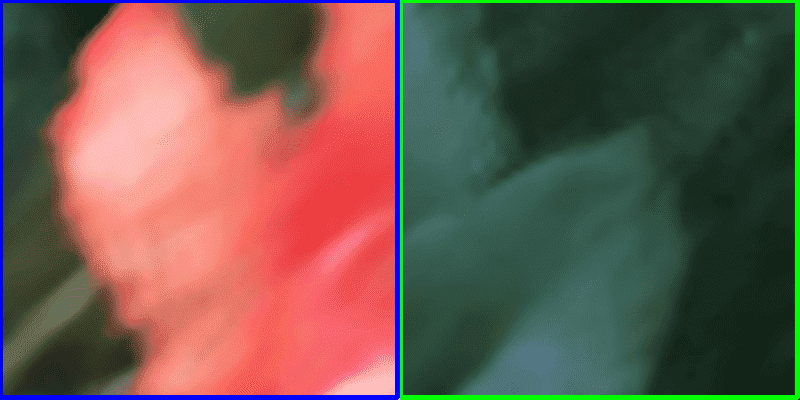}
	\end{minipage}}
	\subfigure[KiloNeRF~\cite{Reiser2021ICCV}]{
		\begin{minipage}[t]{\scale\linewidth}
			\includegraphics[width=\scaleB\linewidth]{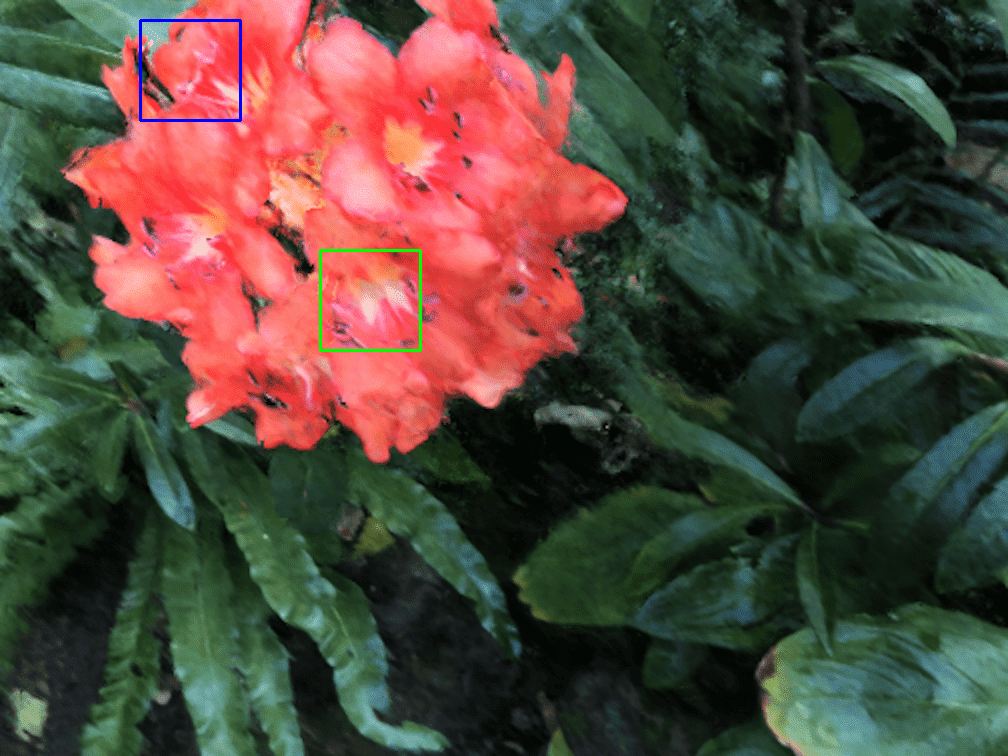}
			\includegraphics[width=\scaleB\linewidth]{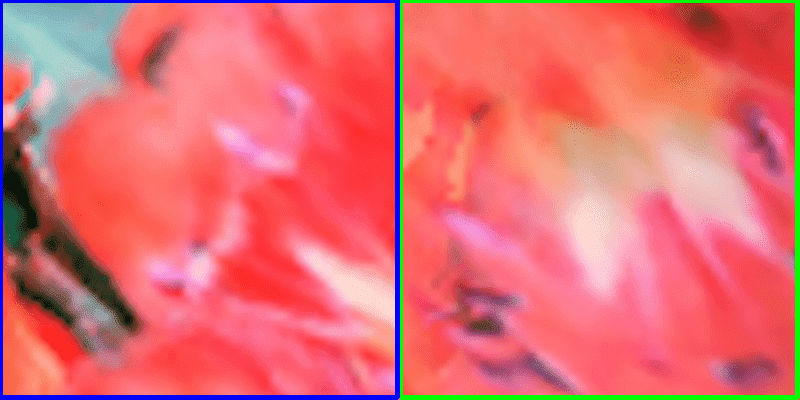}
			\includegraphics[width=\scaleB\linewidth]{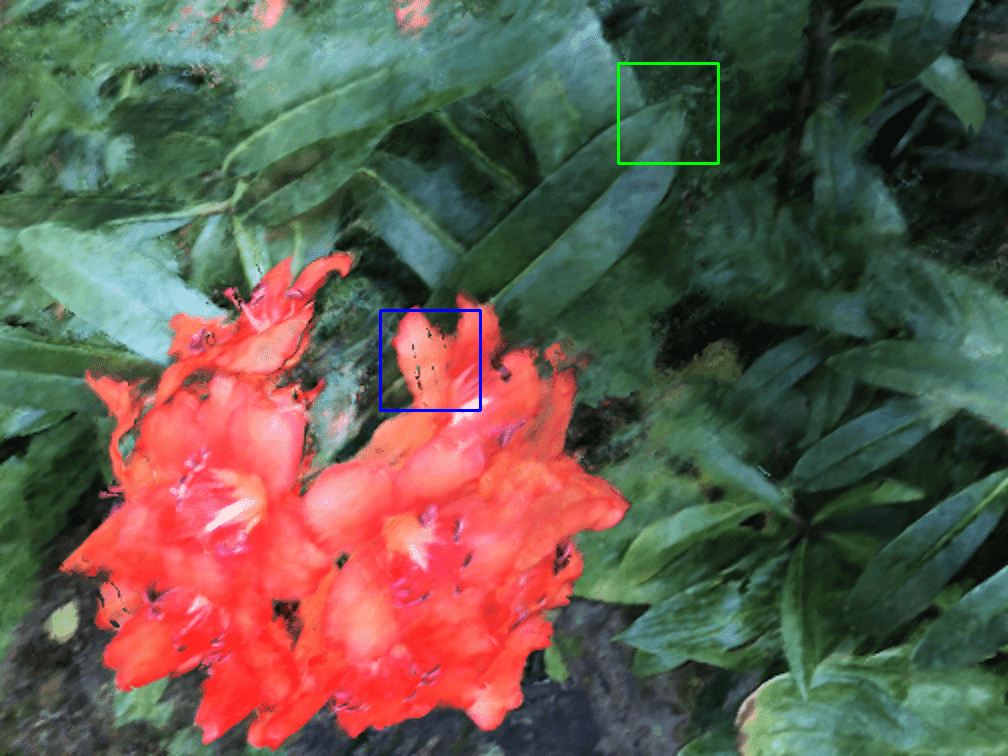}
			\includegraphics[width=\scaleB\linewidth]{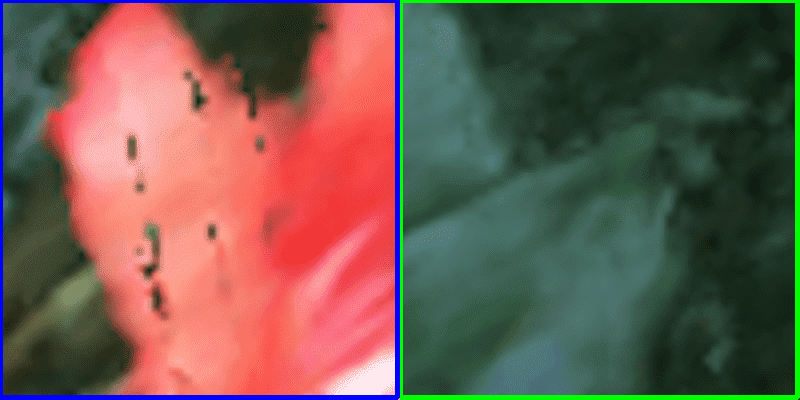}
	\end{minipage}}
	\subfigure[MipNeRF~\cite{barron2021mipnerf}]{
		\begin{minipage}[t]{\scale\linewidth}
			\includegraphics[width=\scaleB\linewidth]{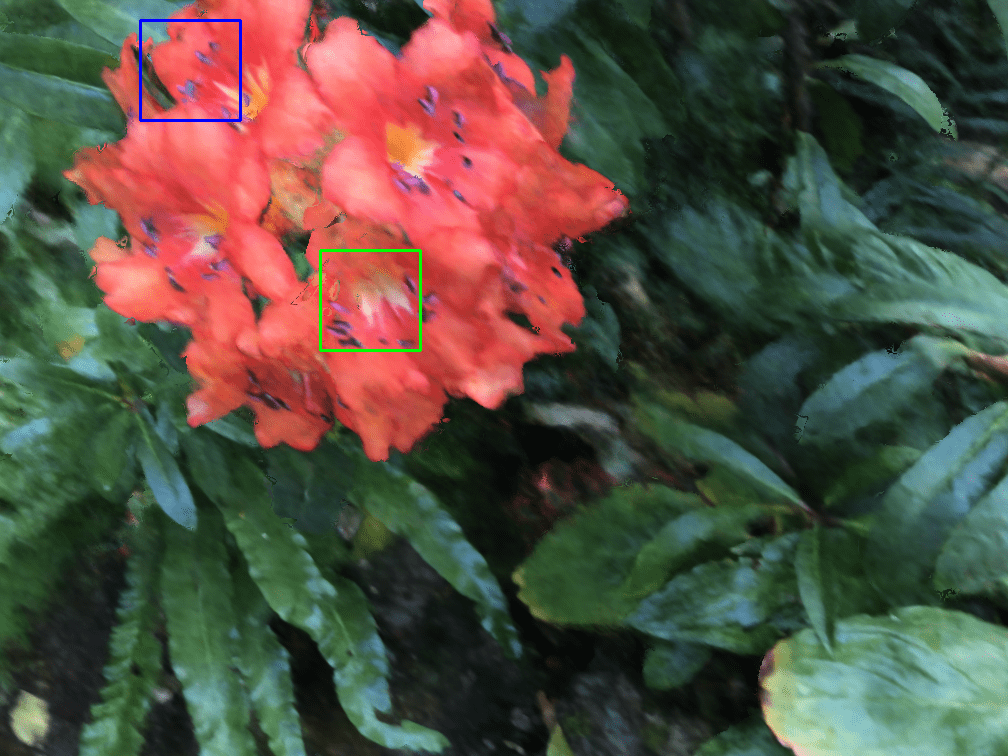}
			\includegraphics[width=\scaleB\linewidth]{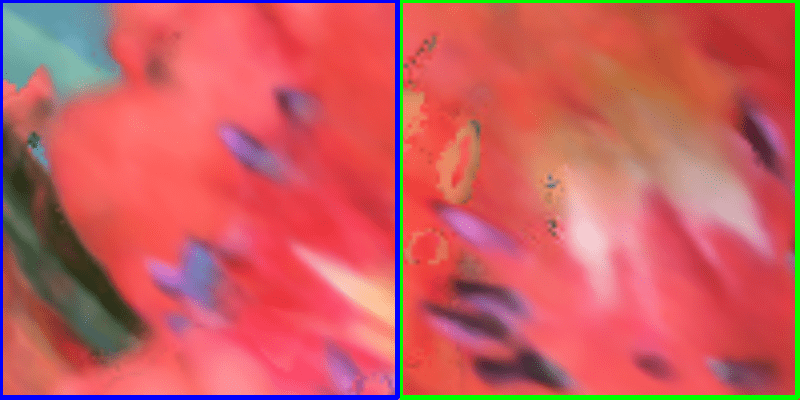}
			\includegraphics[width=\scaleB\linewidth]{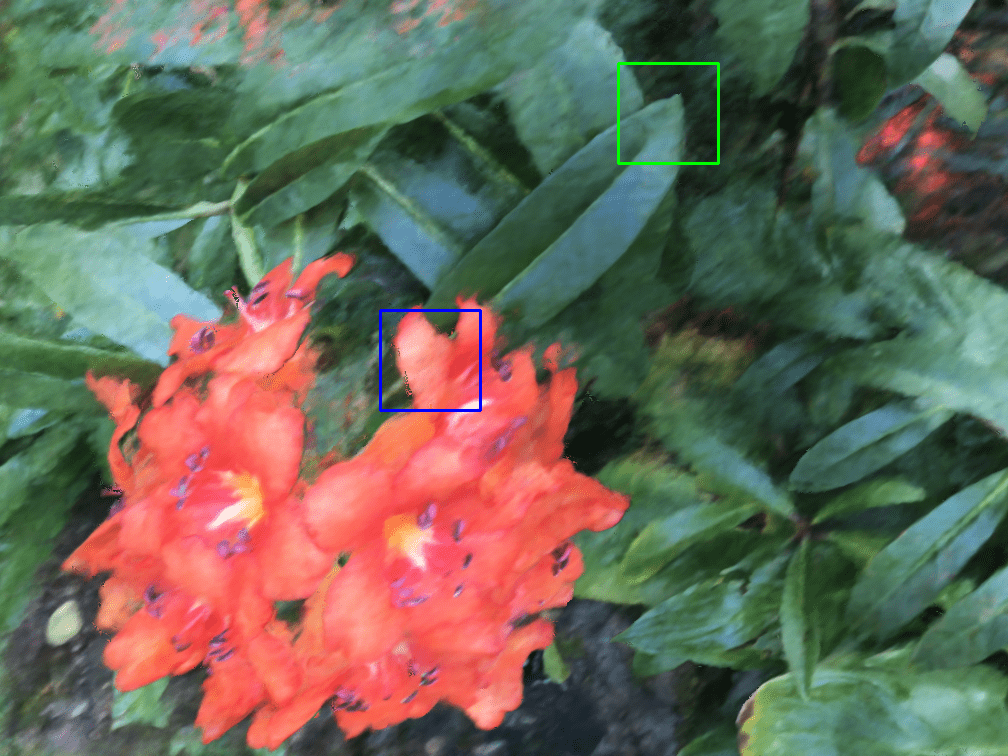}
			\includegraphics[width=\scaleB\linewidth]{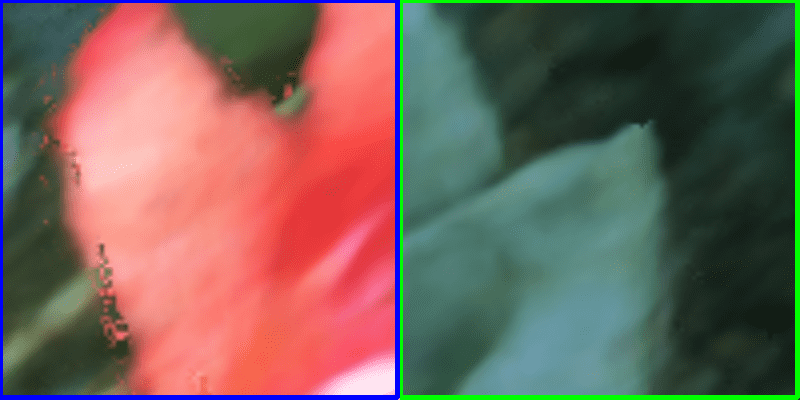}
	\end{minipage}}
	\subfigure[Our]{
		\begin{minipage}[t]{\scale\linewidth}
			\includegraphics[width=\scaleB\linewidth]{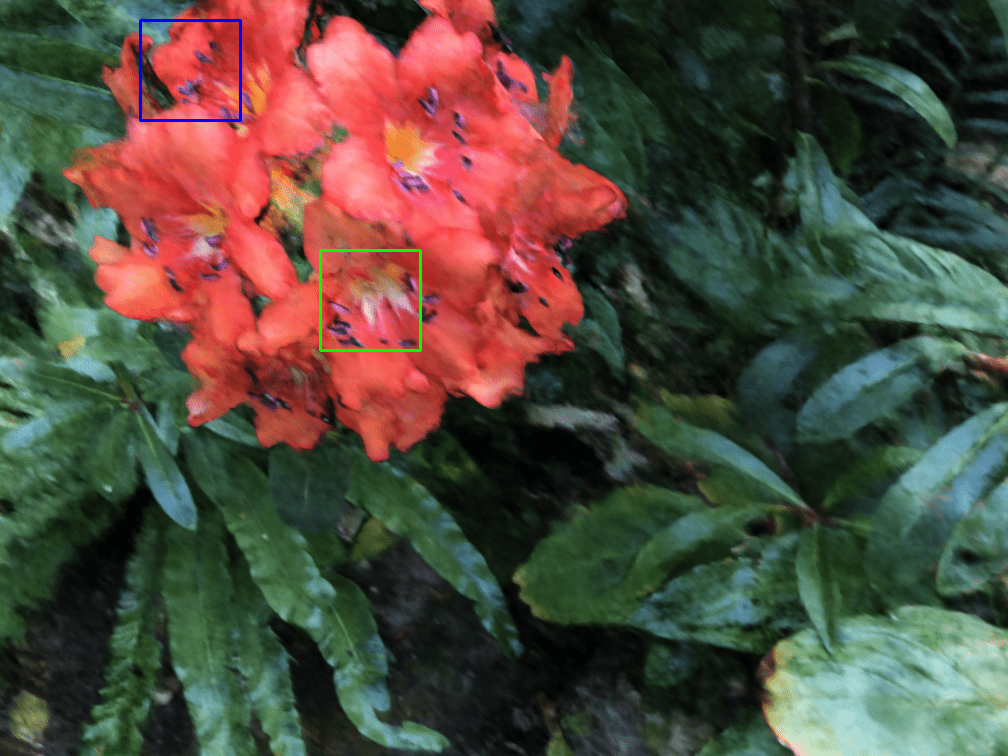}
			\includegraphics[width=\scaleB\linewidth]{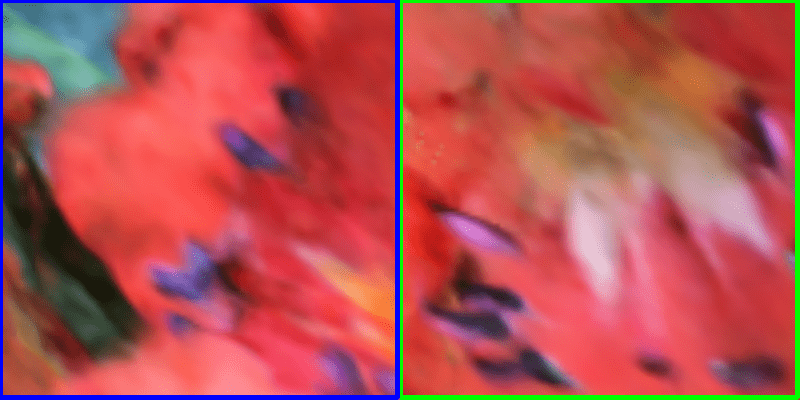}
			\includegraphics[width=\scaleB\linewidth]{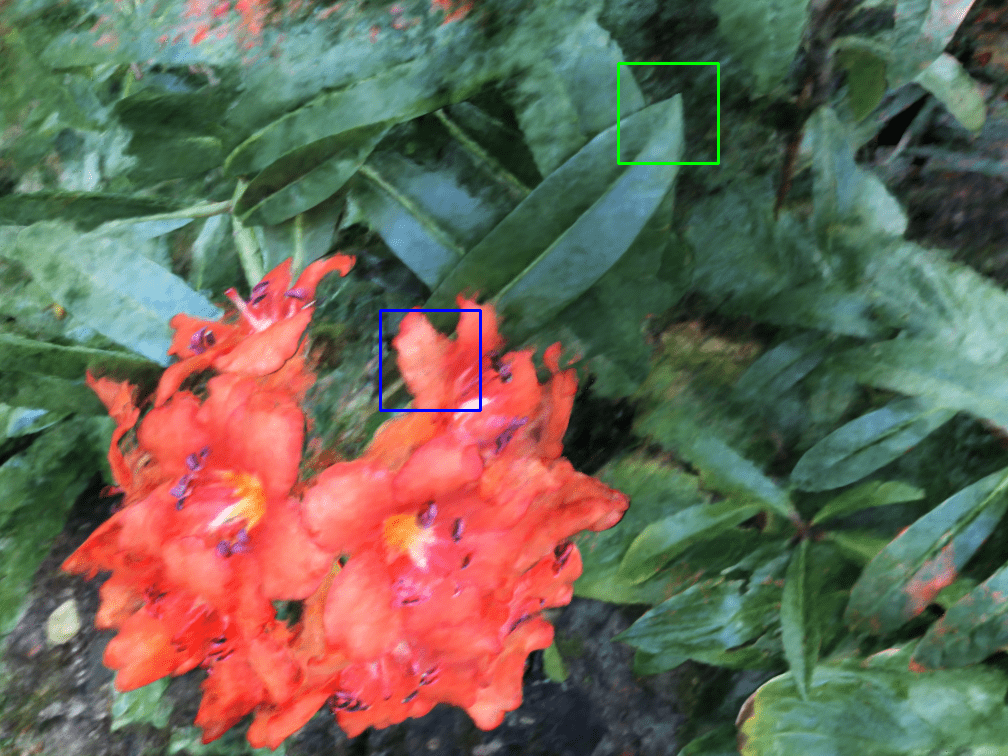}
			\includegraphics[width=\scaleB\linewidth]{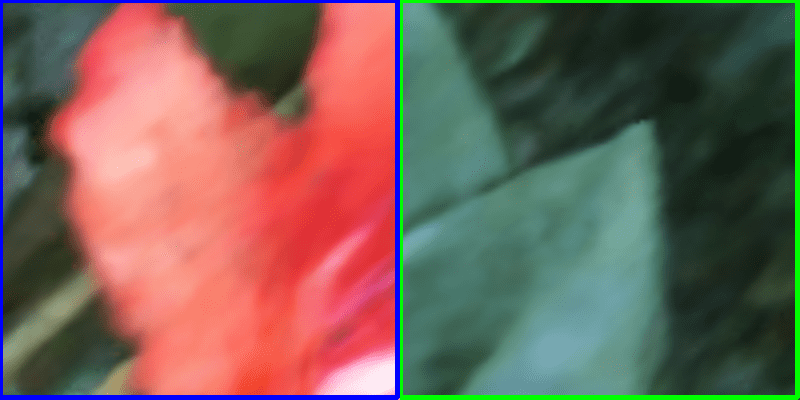}
	\end{minipage}}
	\caption{Novel View Comparison on Real Scene Dataset~\cite{mildenhall2020nerf}. We render extrapolated viewpoints that far away from view sampling in the training dataset, to show the rendering performance for challenging large viewpoint change.}
	\label{fig:extreme_llff_view}
\end{figure*}

\paragraph{Real Scene Dataset}
\figref{fig:extreme_llff_view} shows an extrapolated viewpoint for a
light field dataset, which confirms the findings on the synthetic
data.  NeRF~\cite{mildenhall2020nerf} and
KiloNeRF~\cite{Reiser2021ICCV} exhibit reduced color accuracy in
flower's androecium (see row 2), while our method can faithfully
recover color in fine area due to a better jointly trained geometry
and color representation. Moreover, NeRF~\cite{mildenhall2020nerf} and
KiloNeRF~\cite{Reiser2021ICCV} tend to lose shape details in the
flower and leaves under strong view point changes. MipNeRF produces
sharper results but again also has boundary artifacts at depth
discontinuities, indicating an inaccurate density field. On the other
hand the octree structure of \nascent manages to learn a very detailed
density field that preserves fine structures over extreme viewpoint
changes.

\subsection{Quantitative Comparisons}
In \tabref{tab:quantitative_results}, we compare our reconstruction
results quantitatively against other state-of-the-art works using
PSNR, SSIM, and LPIPS~\cite{zhang2018perceptual} as metrics. Note that
these comparisons are for the {\em view interpolation} scenario since
the datasets do not contain comparison views that are far from the
training data.
The datasets used here are Synthetic-NeRF~\cite{mildenhall2020nerf},
RealScene-LLFF~\cite{mildenhall2019llff}, and the new UAV dataset.
Extensive experiments show that our method is highly competitive on
all datasets. The most contented dataset is the LLFF dataset, where
\nascent loses to NSVF~\cite{liu2020neural} in terms of PSNR and SSIM,
but wins according to LPIPS.  LLFF~\cite{mildenhall2019llff} and
PixelNeRF are only competitive on the narrow baseline light field
data, whereas the other methods show more even performance on all
datasets.

Our proposed method excels at the new UAV dataset, since UAV
viewpoints have a large view of field, sparse viewpoint and long-range
distance, the traditional sampling scheme in NeRF-related methods will
waste a large amount of samples in empty space, or hard to sample proper
candidates of the ground surface due to limited sampling points along the ray
direction. Moreover, NSVF cannot be evaluated on this data because it only reconstructs
bounded scenes with extremely high training time in UAV dataset.
Our octree-based sampling scheme can achieve uniform
sampling inside tree blocks, smaller blocks even have a finer sampling
step, in order to enable a better searching scheme for thin objects.

As the ablation studies in the next section demonstrate, we have the
ability to further improve the quality by using a more powerful network
configuration in each octree node, albeit at a performance cost.

\begin{table*}[]
  \vspace{-3mm}
	\caption{Quantitative Evaluation on Synthetic-NeRF\cite{mildenhall2020nerf}, RealScene-LLFF\cite{mildenhall2019llff}, UAV dataset.}
	\begin{tabular}{cccccccccc}
		\hline
		& \multicolumn{3}{c}{Synthetic-NeRF~\cite{mildenhall2020nerf}}                                                      & \multicolumn{3}{c}{LLFF~\cite{mildenhall2019llff}}                                                                & \multicolumn{3}{c}{UAV dataset}                                                                           \\ \hline
		Method                        & \multicolumn{1}{l}{PSNR$\uparrow$} & \multicolumn{1}{l}{SSIM$\uparrow$} & LPIPS$\downarrow$               & \multicolumn{1}{l}{PSNR$\uparrow$} & \multicolumn{1}{l}{SSIM$\uparrow$} & LPIPS$\downarrow$               & \multicolumn{1}{l}{PSNR$\uparrow$} & \multicolumn{1}{l}{SSIM$\uparrow$} & LPIPS$\downarrow$               \\ \hline
		LLFF         & 26.05                              & 0.893                              & 0.160                        & 25.03                              & 0.793                              & 0.243                        & 23.70                              & 0.834                              & 0.260                        \\ \hline
		NeRF         & 31.01                              & 0.947                              & 0.081                        & 27.15                              & 0.828                              & 0.192                        & {\color[HTML]{000000} 24.98}       & 0.853                              & 0.201                        \\ \hline
		PixelNeRF    & 26.20                              & 0.940                              & 0.080                        & 25.89                              & {\color[HTML]{FE996B} 0.898}       & 0.187                        & 24.69                              & 0.824                              & 0.201                        \\ \hline
		NSVF         & {\color[HTML]{FE996B} 31.75}       & {\color[HTML]{FE996B} 0.954}       & {\color[HTML]{FE0000} 0.048} & -       & -       & - & -                                  & -                                  & -                            \\ \hline
		KiloNeRF     & 30.95                              & 0.937                              & 0.080                        & 26.15                              & 0.828                              & 0.192                        & {\color[HTML]{FE996B} 25.78}       & {\color[HTML]{FE996B} 0.864}       & {\color[HTML]{FE996B} 0.198} \\ \hline
		Our(W64-D8)  & {\color[HTML]{FD6864} 31.85}       & {\color[HTML]{FD6864} 0.967}       & {\color[HTML]{FD6864} 0.049} & {\color[HTML]{FE996B} 27.79}       & {\color[HTML]{FE996B} 0.898}       & {\color[HTML]{FD6864} 0.114} & {\color[HTML]{FD6864} 30.48}       & {\color[HTML]{FD6864} 0.931}       & {\color[HTML]{FD6864} 0.115} \\
		Our(W128-D8) & {\color[HTML]{FE0000} 31.94}       & {\color[HTML]{FE0000} 0.969}       & {\color[HTML]{FE0000} 0.048} & {\color[HTML]{FE0000} 28.19}       & {\color[HTML]{FD6864} 0.903}       & {\color[HTML]{FE0000} 0.113} & {\color[HTML]{FE0000} 30.50}       & {\color[HTML]{FE0000} 0.932}       & {\color[HTML]{FE0000} 0.113} \\ \hline
	\end{tabular}
	\label{tab:quantitative_results}
\end{table*}

\subsection{Training Efficiency Comaprison}
Training time for the Synthetic-NeRF dataset is shown in
Table~\ref{tab:total_time_comparison}. At the
default parameter settings detailed in \secref{sec:model}, \nascent
has faster training times than the competing methods and competitive
rendering times compared to the fastest existing neural rendering
methods. Details of the performance/speed trade-off are provided in
the next section, and more results can be found in the supplement.
\begin{table}[]
	\centering
	\begin{tabular}{lccccc}
		\hline
		& NeRF & KiloNeRF & MipNeRF & Our(W64D8) & Our(W128D8) \\ \hline
		Tot. Time(h) & 6.5  & 18.5     & 5.3     & 4.2        & 11.6        \\ \hline
	\end{tabular}
	\caption{Comparison of total training time for Synthetic-NeRF dataset.}
	\label{tab:total_time_comparison}
\end{table}

%\paragraph{Training Speed Comparison}
%We compare our method for training an model.
%
%\paragraph{Rendering Speed Comparison}
%We compare rendering speed in training an model.

\subsection{Ablation Study}

\begin{figure*}[h]
	\centering
	\def \scale {0.09}
	\def \scaleB {1.0}
	\subfigure[GT]{
		\begin{minipage}[t]{\scale\linewidth}
			\includegraphics[width=\scaleB\linewidth]{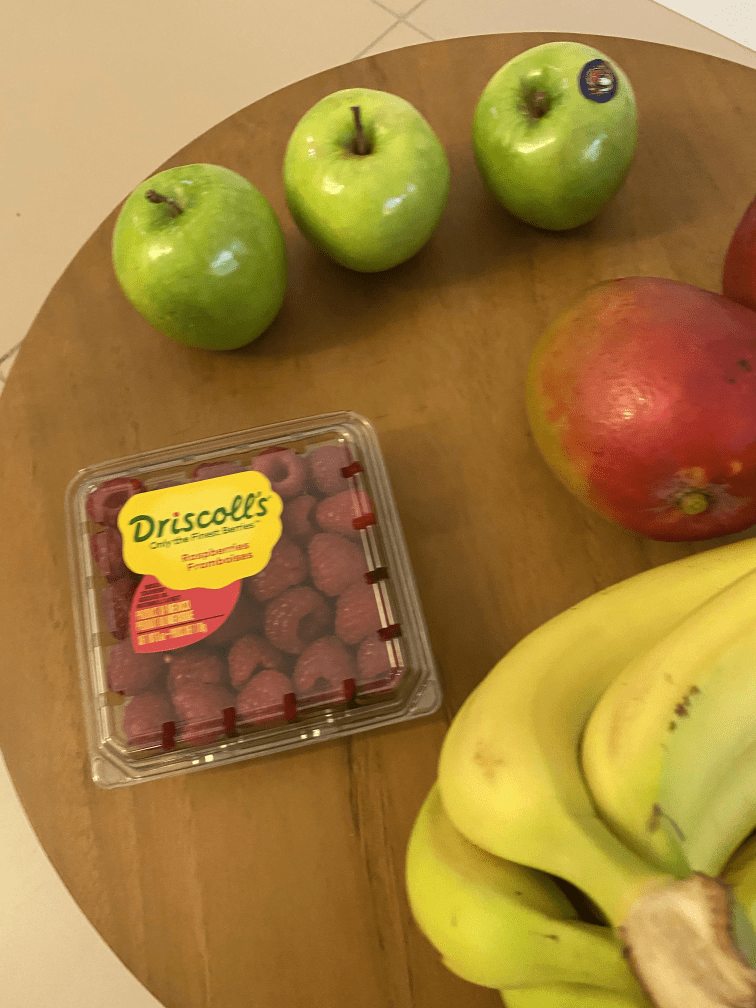}
			\color{white}
			\includegraphics[draft, width=\scaleB\linewidth]{figure/ablation/network/GT/000018}
			\includegraphics[width=\scaleB\linewidth]{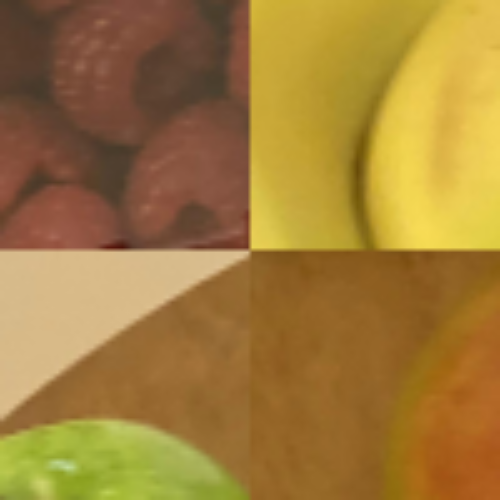}
			\color{white}
			\includegraphics[draft, width=\scaleB\linewidth]{figure/ablation/network/GT/000018_details}
	\end{minipage}}
	\subfigure[W64D4]{
		\begin{minipage}[t]{\scale\linewidth}
			\includegraphics[width=\scaleB\linewidth]{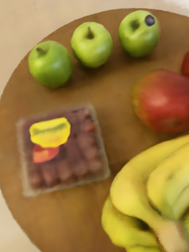}
			\includegraphics[width=\scaleB\linewidth]{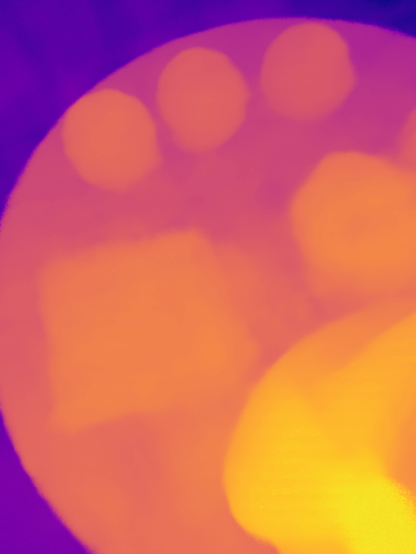}
			\includegraphics[width=\scaleB\linewidth]{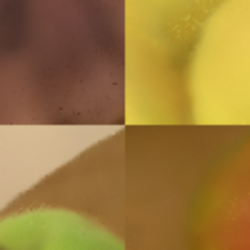}
			\includegraphics[width=\scaleB\linewidth]{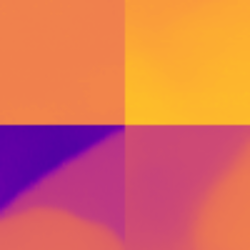}
	\end{minipage}}
	\subfigure[W128D4]{
		\begin{minipage}[t]{\scale\linewidth}
			\includegraphics[width=\scaleB\linewidth]{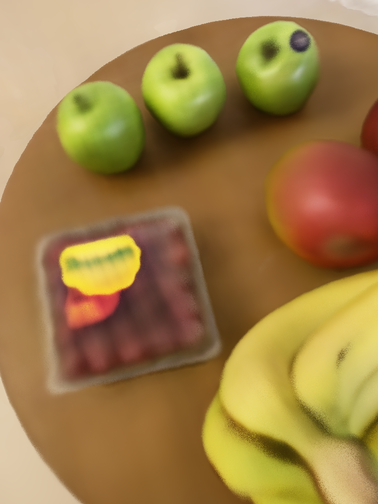}
			\includegraphics[width=\scaleB\linewidth]{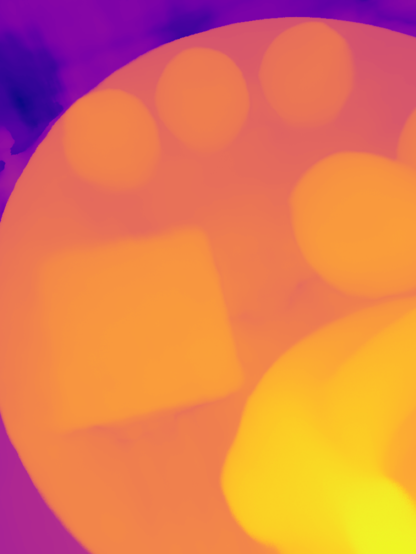}
			\includegraphics[width=\scaleB\linewidth]{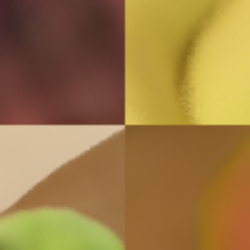}
			\includegraphics[width=\scaleB\linewidth]{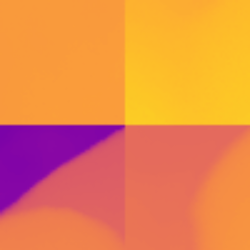}
	\end{minipage}}
	\subfigure[W256D4]{
		\begin{minipage}[t]{\scale\linewidth}
			\includegraphics[width=\scaleB\linewidth]{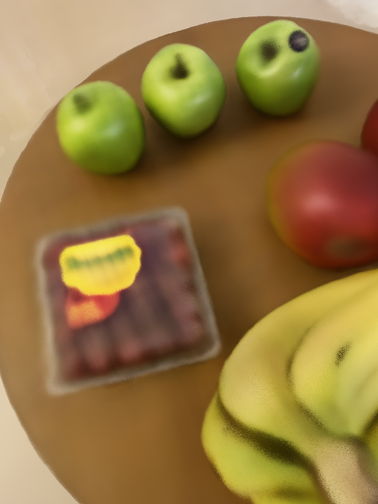}
			\includegraphics[width=\scaleB\linewidth]{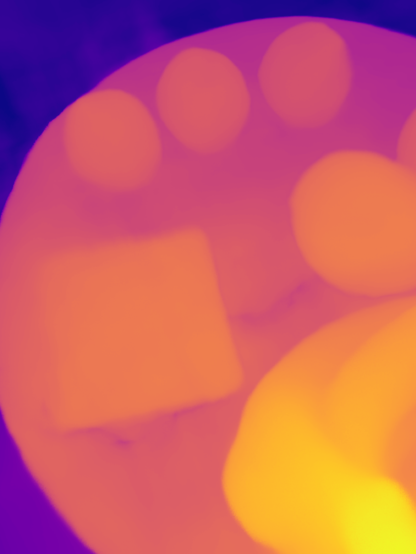}
			\includegraphics[width=\scaleB\linewidth]{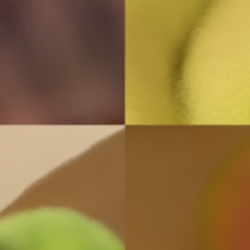}
			\includegraphics[width=\scaleB\linewidth]{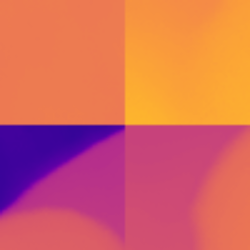}
	\end{minipage}}
	\subfigure[W512D4]{
		\begin{minipage}[t]{\scale\linewidth}
			\includegraphics[width=\scaleB\linewidth]{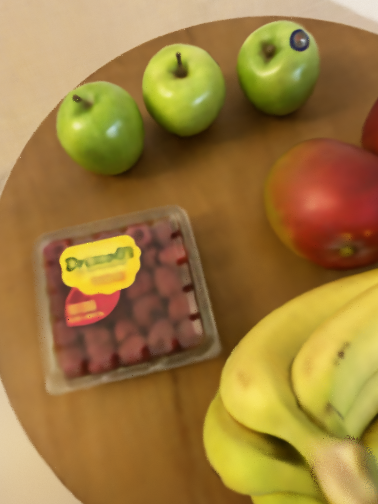}
			\includegraphics[width=\scaleB\linewidth]{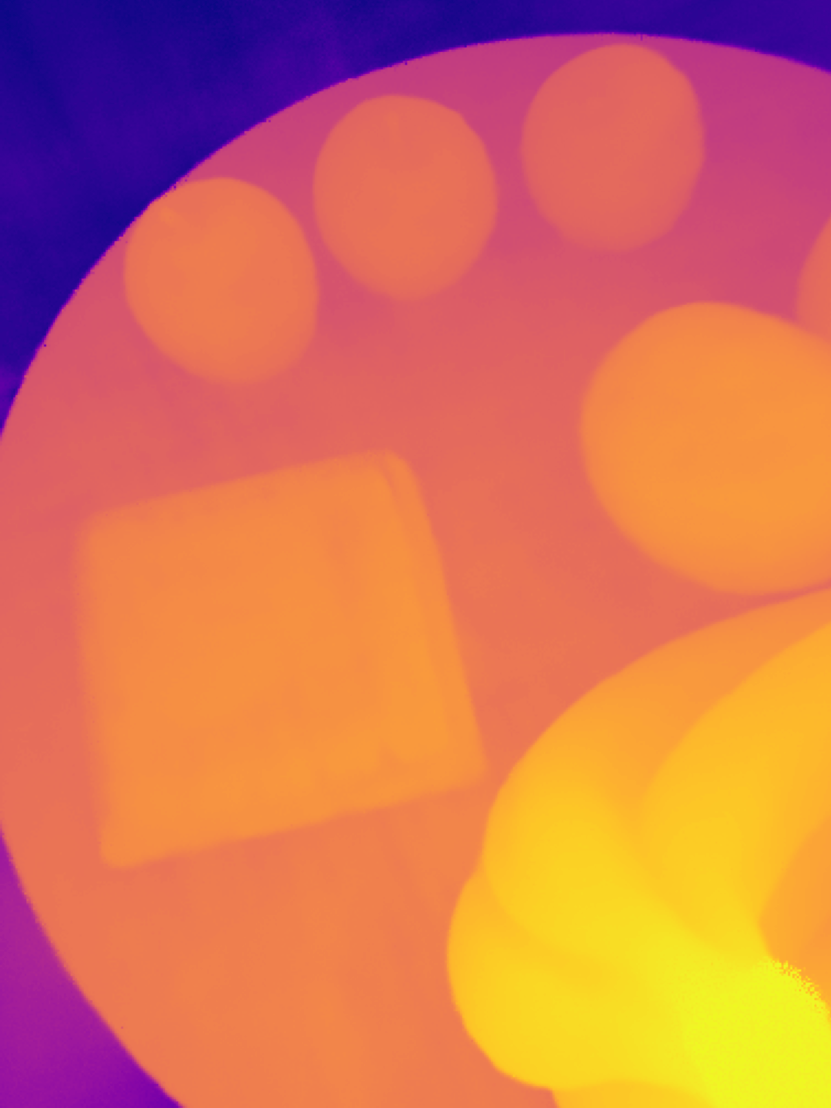}
			\includegraphics[width=\scaleB\linewidth]{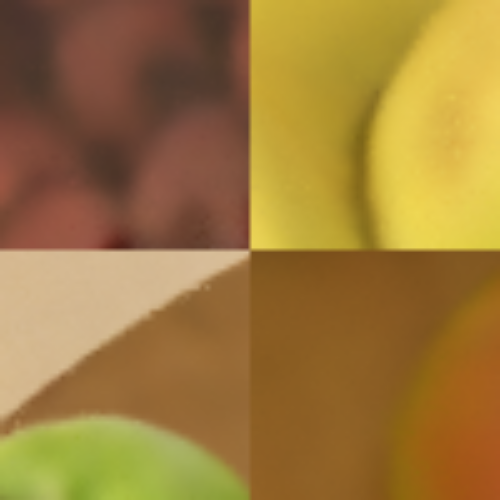}
			\includegraphics[width=\scaleB\linewidth]{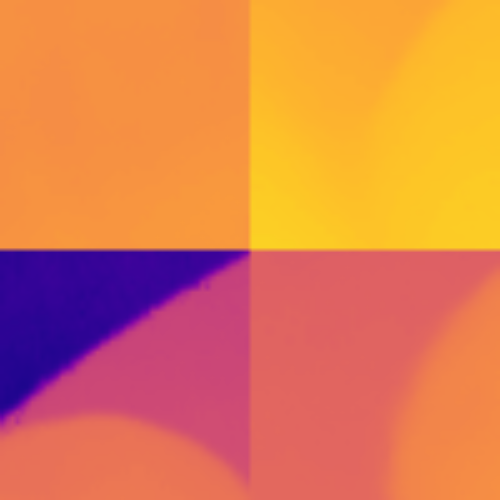}
	\end{minipage}}
	\subfigure[W64D8]{
		\begin{minipage}[t]{\scale\linewidth}
			\includegraphics[width=\scaleB\linewidth]{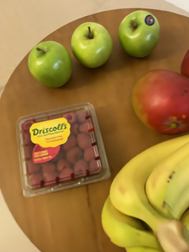}
			\includegraphics[width=\scaleB\linewidth]{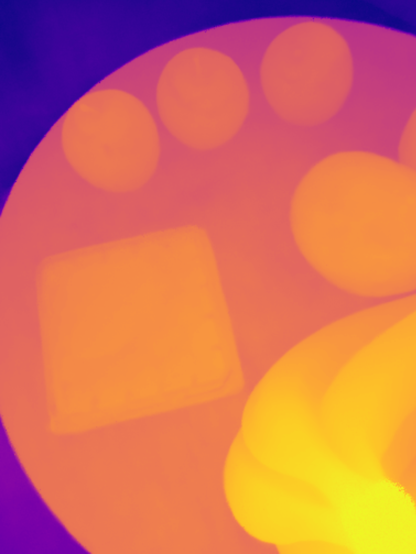}
			\includegraphics[width=\scaleB\linewidth]{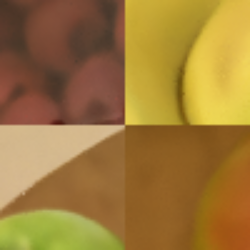}
			\includegraphics[width=\scaleB\linewidth]{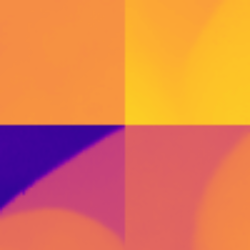}
	\end{minipage}}
	\subfigure[W128D8]{
		\begin{minipage}[t]{\scale\linewidth}
			\includegraphics[width=\scaleB\linewidth]{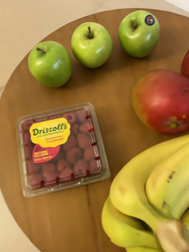}
			\includegraphics[width=\scaleB\linewidth]{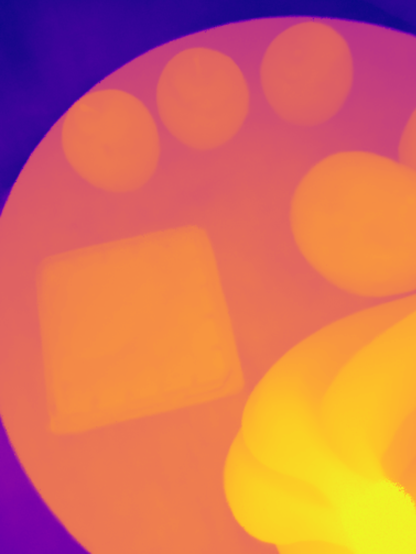}
			\includegraphics[width=\scaleB\linewidth]{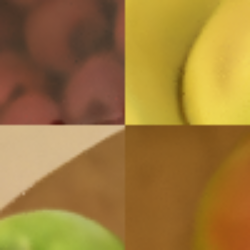}
			\includegraphics[width=\scaleB\linewidth]{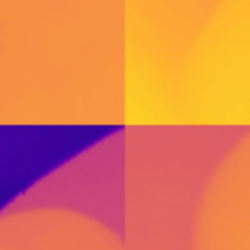}
	\end{minipage}}
	\subfigure[W256D8]{
		\begin{minipage}[t]{\scale\linewidth}
			\includegraphics[width=\scaleB\linewidth]{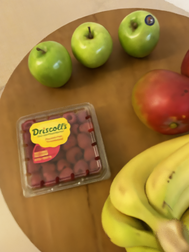}
			\includegraphics[width=\scaleB\linewidth]{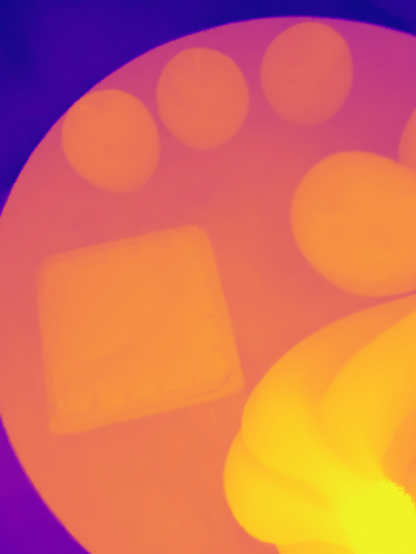}
			\includegraphics[width=\scaleB\linewidth]{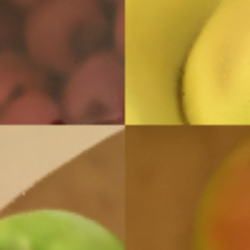}
			\includegraphics[width=\scaleB\linewidth]{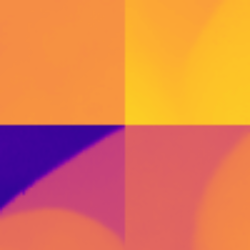}
	\end{minipage}}
	\subfigure[W512D8]{
		\begin{minipage}[t]{\scale\linewidth}
			\includegraphics[width=\scaleB\linewidth]{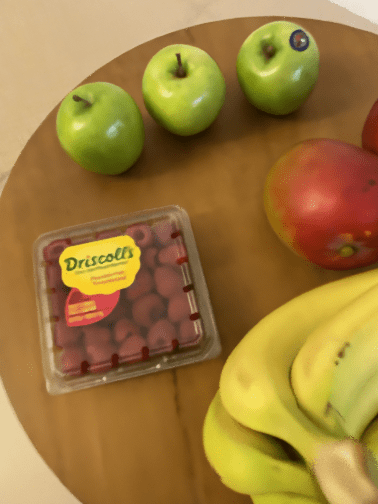}
			\includegraphics[width=\scaleB\linewidth]{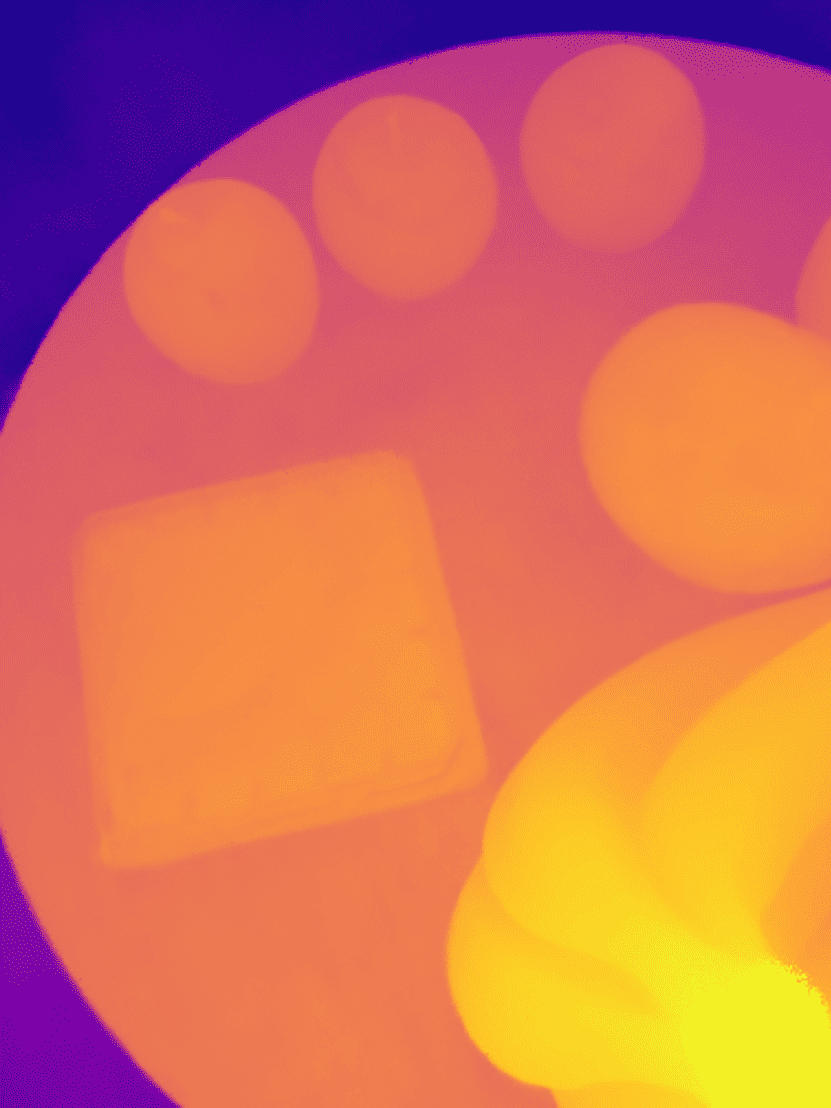}
			\includegraphics[width=\scaleB\linewidth]{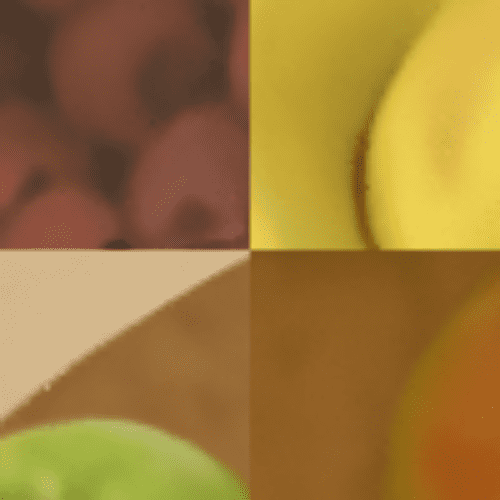}
			\includegraphics[width=\scaleB\linewidth]{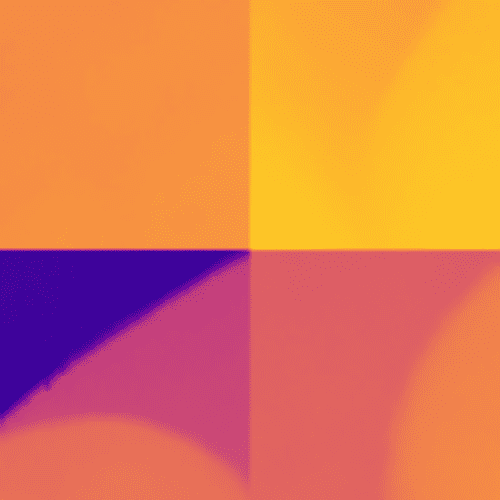}
	\end{minipage}}
	\caption{Ablation study for various types of sub-networks architecture. The network width are $\{64, 128, 256, 512\}$, the network depth are $\{4, 8\}$.}
	\label{fig:network_types}
\end{figure*}

\paragraph{Network Architecture}
We perform an ablation study on the hyper-parameters of the implicit
networks for each octree node in \tabref{tab:network_ablation}.
%\todo{mention fruit dataset somewhere}
For these use the fruit dataset that contains various zoom-in and
zoom-out views to perform ablation study.  Note that W and D refer to
width and depth of the network.  In \figref{fig:network_types}, we
also show novel view rendering results for various sub-networks for a
training epochs of 20. Experiments show that the higher approximation
power of larger networks improves the image quality, although at
significantly higher computational cost. Our default parameters
(W64-D8) are on the lower end of the quality scale but provide
excellent training and rendering times, and still provide better
quality than the comparison methods, as shown above.

\begin{table*}[]
	\centering
	\caption{Ablation study on unit network architecture. We fix
          an optimized octree and replace network architecture in each
          node to show rendering performance, test on the fruit
          dataset ($1008\times756$). \label{tab:network_ablation} }
%	\begin{tabular}{lccccc}
%		\hline
%		\multicolumn{1}{c}{Network} & PSNR$\uparrow$ & SSIM$\uparrow$ & LPIPS$\downarrow$ & Training/epoch & Rendering/frame \\ \hline
%		W64-D4                      & 25.78          & 0.921          & 0.150             & 5 min          & 10 s            \\ \hline
%		W64-D8                      & 27.82          & 0.941          & 0.110             & 6 min          & 12 s            \\ \hline
%		W128-D4                     & 26.35          & 0.919          & 0.105             & 18 min         & 45 s            \\ \hline
%		W128-D8                     & 27.96          & 0.941          & 0.103             & 20 min         & 47 s            \\ \hline
%		W256-D4                     & 28.25          & 0.903          & 0.095             & 25 min         & 55 s            \\ \hline
%		W256-D8                     & 29.49          & 0.951          & 0.092             & 35 min         & 1.5 min         \\ \hline
%		W512-D4                     & 28.55          & 0.931          & 0.082             & $\sim$1.2 h    & 1.3 min         \\ \hline
%		W512-D8                     & 31.25          & 0.961          & 0.077             & $\sim$1.6 h    & 2.5 min         \\ \hline
%	\end{tabular}
\begin{tabular}{lccccc}
	\hline
	\multicolumn{1}{c}{Network} & PSNR$\uparrow$ & SSIM$\uparrow$ & LPIPS$\downarrow$ & Train/epoch & Render/frame \\ \hline
	W64-D4                      & 26.55          & 0.921          & 0.104             & 4 min          & 10 s            \\ \hline
	W64-D8                      & 29.21          & 0.952          & 0.093             & 6 min          & 12 s            \\ \hline
	W128-D4                     & 27.85          & 0.922          & 0.105             & 18 min         & 45 s            \\ \hline
	W128-D8                     & 29.36          & 0.953          & 0.093             & 20 min         & 47 s            \\ \hline
	W256-D4                     & 28.65          & 0.922          & 0.105             & 25 min         & 55 s            \\ \hline
	W256-D8                     & 29.89          & 0.958          & 0.091             & 35 min         & 1.5 min         \\ \hline
	%W512-D4                     & 28.55          & 0.931          & 0.082             & $\sim$1.2 h    & 1.3 min         \\ \hline
	%W512-D8                     & 31.25          & 0.961          & 0.077             & $\sim$1.6 h    & 2.5 min         \\ \hline
\end{tabular}
\end{table*}

\begin{table}[]
	\centering
	\caption{Ablation study on octree levels.	\label{tab:level_ablation}
}
	\begin{tabular}{lccc}
		\hline
		\multicolumn{1}{c}{Level (No.~$\mlp$)} & PSNR$\uparrow$ & SSIM$\uparrow$ & LPIPS$\downarrow$ \\ \hline
		level 0 (1)                 & 20.11          & 0.852          & 0.220             \\ \hline
		level 1 (8)                 & 23.42          & 0.871          & 0.180             \\ \hline
		level 2 (64)                & 27.82          & 0.941          & 0.110             \\ \hline
		level 3 (512)               & 29.96          & 0.958          & 0.091             \\ \hline
		level 4 (2048)              & 30.20          & 0.959          & 0.080             \\ \hline
		level all (2633)            & 30.75          & 0.961          & 0.078             \\ \hline
	\end{tabular}
\end{table}

\paragraph{Octree Structure}
We also compare rendering performance for different granularity of the
octree, i.e., the number of octree levels. In general, finer scale
octree will have smaller block size and higher representation capacity
with higher quantity of sub-networks, therefore, there is a trade-off
between the number of sub-networks and the representation
capacity. The network architecture is $W64-D8$, and use same dataset
as \tabref{tab:network_ablation}. \tabref{tab:level_ablation} shows
that a reduction of the octree levels (level 0, 1) has poor
performance in rendering, and are thus, and is thus only used for
initial training when initializing the system. Level 4 has the best
performance with the highest number of networks, but will also lead to
the highest computation and storage burden, and is therefore, only
active in regions of high complexity. In general, we start training in
level 0 or 1 for a warm initialization and initial octree structure,
and level 2, 3 are active levels during the main training and
rendering process.

\section{Conclusions}

In this paper, we have presented Neural Adaptive Scene Tracing
(\nascent), a hybrid explicit-implicit neural rendering
approach that can be trained directly in the 2D image data. The model
representation consists of a hierarchical and adaptive octree
structure with a per-node implicit network. We use this model in
combination with an optimized two-stage sampling process that
maximizes the re-use of view-independent data in order to reduce
the number of neural network evaluations. This, together with a strong
spatial clustering of the samples near interesting object surfaces,
enables improved training times as well as superior results
compared to other neural rendering approaches.

The ablation studies show that the quality of the reconstructions can
be further improved by utilizing more powerful networks in each node,
albeit at significantly increased training and rendering times. We
believe this topic merits further investigation. For example one may
choose different network hyper parameters for nodes in different
regions, based on either a heuristic or neural architecture search.
This could further improve the quality while bounding the increase in
compute time.

\nascent is implemented in PyTorch, and the source code and UAV
dataset will be made available at the time of publication.

\bibliographystyle{splncs04}
\bibliography{egbib}

\begin{thebibliography}{10}
\providecommand{\url}[1]{\texttt{#1}}
\providecommand{\urlprefix}{URL }
\providecommand{\doi}[1]{https://doi.org/#1}

\bibitem{aharchi2019review}
Aharchi, M., Kbir, M.A.: A review on {3D} reconstruction techniques from {2D}
  images. In: The Proceedings of the Third International Conference on Smart
  City Applications. pp. 510--522. Springer (2019)

\bibitem{aliev2020neural}
Aliev, K.A., Sevastopolsky, A., Kolos, M., Ulyanov, D., Lempitsky, V.: Neural
  point-based graphics. In: Computer Vision--ECCV 2020: 16th European
  Conference, Glasgow, UK, August 23--28, 2020, Proceedings, Part XXII 16. pp.
  696--712. Springer (2020)

\bibitem{barron2021mipnerf}
Barron, J.T., Mildenhall, B., Tancik, M., Hedman, P., Martin-Brualla, R.,
  Srinivasan, P.P.: Mip-nerf: A multiscale representation for anti-aliasing
  neural radiance fields. In: ICCV (2021)

\bibitem{boss2021nerd}
Boss, M., Braun, R., Jampani, V., Barron, J.T., Liu, C., Lensch, H.: Nerd:
  {N}eural reflectance decomposition from image collections. In: Proceedings of
  the IEEE/CVF International Conference on Computer Vision. pp. 12684--12694
  (2021)

\bibitem{chan2021pi}
Chan, E.R., Monteiro, M., Kellnhofer, P., Wu, J., Wetzstein, G.: pi-gan:
  {P}eriodic implicit generative adversarial networks for 3d-aware image
  synthesis. In: Proceedings of the IEEE/CVF Conference on Computer Vision and
  Pattern Recognition. pp. 5799--5809 (2021)

\bibitem{chibane2020implicit}
Chibane, J., Pons-Moll, G.: Implicit feature networks for texture completion
  from partial 3d data. In: European Conference on Computer Vision. pp.
  717--725. Springer (2020)

\bibitem{dahnert2021panoptic}
Dahnert, M., Hou, J., Nie{\ss}ner, M., Dai, A.: Panoptic 3d scene
  reconstruction from a single rgb image. Advances in Neural Information
  Processing Systems  \textbf{34} (2021)

\bibitem{eslami2018neural}
Eslami, S.A., Rezende, D.J., Besse, F., Viola, F., Morcos, A.S., Garnelo, M.,
  Ruderman, A., Rusu, A.A., Danihelka, I., Gregor, K., et~al.: Neural scene
  representation and rendering. Science  \textbf{360}(6394),  1204--1210 (2018)

\bibitem{flynn2019deepview}
Flynn, J., Broxton, M., Debevec, P., DuVall, M., Fyffe, G., Overbeck, R.,
  Snavely, N., Tucker, R.: Deepview: {V}iew synthesis with learned gradient
  descent. In: Proceedings of the IEEE/CVF Conference on Computer Vision and
  Pattern Recognition. pp. 2367--2376 (2019)

\bibitem{garbin2021fastnerf}
Garbin, S.J., Kowalski, M., Johnson, M., Shotton, J., Valentin, J.: {FastNeRF}:
  High-fidelity neural rendering at 200fps. arXiv preprint arXiv:2103.10380
  (2021)

\bibitem{gurram20073d}
Gurram, P., Lach, S., Saber, E., Rhody, H., Kerekes, J.: 3d scene
  reconstruction through a fusion of passive video and lidar imagery. In: 36th
  Applied Imagery Pattern Recognition Workshop (aipr 2007). pp. 133--138. IEEE
  (2007)

\bibitem{ham2019computer}
Ham, H., Wesley, J., Hendra, H.: Computer vision based {3D} reconstruction: {A}
  review. International Journal of Electrical and Computer Engineering
  \textbf{9}(4), ~2394 (2019)

\bibitem{hedman2018deep}
Hedman, P., Philip, J., Price, T., Frahm, J.M., Drettakis, G., Brostow, G.:
  Deep blending for free-viewpoint image-based rendering. ACM Transactions on
  Graphics (TOG)  \textbf{37}(6),  1--15 (2018)

\bibitem{jensen2014large}
Jensen, R., Dahl, A., Vogiatzis, G., Tola, E., Aan{\ae}s, H.: Large scale
  multi-view stereopsis evaluation. In: 2014 IEEE Conference on Computer Vision
  and Pattern Recognition. pp. 406--413. IEEE (2014)

\bibitem{kuhner2020large}
K{\"u}hner, T., K{\"u}mmerle, J.: Large-scale volumetric scene reconstruction
  using lidar. In: 2020 IEEE International Conference on Robotics and
  Automation (ICRA). pp. 6261--6267. IEEE (2020)

\bibitem{Lin_2021_ICCV}
Lin, C.H., Ma, W.C., Torralba, A., Lucey, S.: {BARF: B}undle-adjusting neural
  radiance fields. In: Proceedings of the IEEE/CVF International Conference on
  Computer Vision (ICCV). pp. 5741--5751 (October 2021)

\bibitem{lindell2021autoint}
Lindell, D.B., Martel, J.N., Wetzstein, G.: {AutoInt: A}utomatic integration
  for fast neural volume rendering. In: CVPR (2021)

\bibitem{liu2020neural}
Liu, L., Gu, J., Lin, K.Z., Chua, T.S., Theobalt, C.: Neural sparse voxel
  fields. NeurIPS  (2020)

\bibitem{martel2021acorn}
Martel, J.N., Lindell, D.B., Lin, C.Z., Chan, E.R., Monteiro, M., Wetzstein,
  G.: {ACORN: A}daptive coordinate networks for neural representation. ACM
  Trans. Graph. (SIGGRAPH)  (2021)

\bibitem{meshry2019neural}
Meshry, M., Goldman, D.B., Khamis, S., Hoppe, H., Pandey, R., Snavely, N.,
  Martin-Brualla, R.: Neural rerendering in the wild. In: CVPR. pp. 6878--6887
  (2019)

\bibitem{mildenhall2019llff}
Mildenhall, B., Srinivasan, P.P., Ortiz-Cayon, R., Kalantari, N.K.,
  Ramamoorthi, R., Ng, R., Kar, A.: Local light field fusion: Practical view
  synthesis with prescriptive sampling guidelines. ACM Transactions on Graphics
  (TOG)  (2019)

\bibitem{mildenhall2020nerf}
Mildenhall, B., Srinivasan, P.P., Tancik, M., Barron, J.T., Ramamoorthi, R.,
  Ng, R.: Nerf: Representing scenes as neural radiance fields for view
  synthesis. In: ECCV (2020)

\bibitem{niemeyer2020differentiable}
Niemeyer, M., Mescheder, L., Oechsle, M., Geiger, A.: Differentiable volumetric
  rendering: Learning implicit 3d representations without 3d supervision. In:
  Proceedings of the IEEE/CVF Conference on Computer Vision and Pattern
  Recognition. pp. 3504--3515 (2020)

\bibitem{oechsle2019texture}
Oechsle, M., Mescheder, L., Niemeyer, M., Strauss, T., Geiger, A.: Texture
  fields: {L}earning texture representations in function space. In: Proceedings
  of the IEEE/CVF International Conference on Computer Vision. pp. 4531--4540
  (2019)

\bibitem{park2019deepsdf}
Park, J.J., Florence, P., Straub, J., Newcombe, R., Lovegrove, S.: Deepsdf:
  Learning continuous signed distance functions for shape representation. In:
  Proceedings of the IEEE/CVF Conference on Computer Vision and Pattern
  Recognition. pp. 165--174 (2019)

\bibitem{Reiser2021ICCV}
Reiser, C., Peng, S., Liao, Y., Geiger, A.: {KiloNeRF: S}peeding up neural
  radiance fields with thousands of tiny mlps. In: International Conference on
  Computer Vision (ICCV) (2021)

\bibitem{saito2019pifu}
Saito, S., Huang, Z., Natsume, R., Morishima, S., Kanazawa, A., Li, H.: Pifu:
  {Pixel-aligned} implicit function for high-resolution clothed human
  digitization. In: Proceedings of the IEEE/CVF International Conference on
  Computer Vision. pp. 2304--2314 (2019)

\bibitem{schonberger2016structure}
Schonberger, J.L., Frahm, J.M.: Structure-from-motion revisited. In:
  Proceedings of the IEEE conference on computer vision and pattern
  recognition. pp. 4104--4113 (2016)

\bibitem{schwarz2020graf}
Schwarz, K., Liao, Y., Niemeyer, M., Geiger, A.: Graf: {G}enerative radiance
  fields for 3d-aware image synthesis. arXiv preprint arXiv:2007.02442  (2020)

\bibitem{seitz2006comparison}
Seitz, S.M., Curless, B., Diebel, J., Scharstein, D., Szeliski, R.: A
  comparison and evaluation of multi-view stereo reconstruction algorithms. In:
  2006 IEEE computer society conference on computer vision and pattern
  recognition (CVPR'06). vol.~1, pp. 519--528. IEEE (2006)

\bibitem{sitzmann2019siren}
Sitzmann, V., Martel, J.N., Bergman, A.W., Lindell, D.B., Wetzstein, G.:
  Implicit neural representations with periodic activation functions. In: Proc.
  NeurIPS (2020)

\bibitem{sitzmann2019deepvoxels}
Sitzmann, V., Thies, J., Heide, F., Nie{\ss}ner, M., Wetzstein, G., Zollhofer,
  M.: Deepvoxels: {Learning} persistent 3d feature embeddings. In: Proceedings
  of the IEEE/CVF Conference on Computer Vision and Pattern Recognition. pp.
  2437--2446 (2019)

\bibitem{sitzmann2019scene}
Sitzmann, V., Zollh{\"o}fer, M., Wetzstein, G.: Scene representation networks:
  {C}ontinuous 3d-structure-aware neural s cene representations. arXiv preprint
  arXiv:1906.01618  (2019)

\bibitem{srinivasan2021nerv}
Srinivasan, P.P., Deng, B., Zhang, X., Tancik, M., Mildenhall, B., Barron,
  J.T.: {NeRV}: Neural reflectance and visibility fields for relighting and
  view synthesis. In: Proceedings of the IEEE/CVF Conference on Computer Vision
  and Pattern Recognition. pp. 7495--7504 (2021)

\bibitem{tewari2020state}
Tewari, A., Fried, O., Thies, J., Sitzmann, V., Lombardi, S., Sunkavalli, K.,
  Martin-Brualla, R., Simon, T., Saragih, J., Nie{\ss}ner, M., et~al.: State of
  the art on neural rendering. In: Computer Graphics Forum. vol.~39, pp.
  701--727. Wiley Online Library (2020)

\bibitem{xian2021space}
Xian, W., Huang, J.B., Kopf, J., Kim, C.: Space-time neural irradiance fields
  for free-viewpoint video. In: Proceedings of the IEEE/CVF Conference on
  Computer Vision and Pattern Recognition. pp. 9421--9431 (2021)

\bibitem{xiang2021neutex}
Xiang, F., Xu, Z., Hasan, M., Hold-Geoffroy, Y., Sunkavalli, K., Su, H.:
  {NeuTex: Neural} texture mapping for volumetric neural rendering. In:
  Proceedings of the IEEE/CVF Conference on Computer Vision and Pattern
  Recognition. pp. 7119--7128 (2021)

\bibitem{Yifan:IsoPoints:2021}
Yifan, W., Wu, S., {\"{O}}ztireli, C., Sorkine-Hornung, O.: Iso-points:
  Optimizing neural implicit surfaces with hybrid representations. In: CVPR
  (2021)

\bibitem{yu2021plenoctrees}
Yu, A., Li, R., Tancik, M., Li, H., Ng, R., Kanazawa, A.: {PlenOctrees} for
  real-time rendering of neural radiance fields. In: ICCV (2021)

\bibitem{yu2020pixelnerf}
Yu, A., Ye, V., Tancik, M., Kanazawa, A.: {pixelNeRF}: Neural radiance fields
  from one or few images. In: CVPR (2021)

\bibitem{zhang2021physg}
Zhang, K., Luan, F., Wang, Q., Bala, K., Snavely, N.: {PhySG: Inverse}
  rendering with spherical gaussians for physics-based material editing and
  relighting. In: Proceedings of the IEEE/CVF Conference on Computer Vision and
  Pattern Recognition. pp. 5453--5462 (2021)

\bibitem{zhang2018perceptual}
Zhang, R., Isola, P., Efros, A.A., Shechtman, E., Wang, O.: The unreasonable
  effectiveness of deep features as a perceptual metric. In: CVPR (2018)

\bibitem{zhou2018stereo}
Zhou, T., Tucker, R., Flynn, J., Fyffe, G., Snavely, N.: Stereo magnification:
  Learning view synthesis using multiplane images. arXiv preprint
  arXiv:1805.09817  (2018)

\bibitem{zollhofer2018state}
Zollh{\"o}fer, M., Stotko, P., G{\"o}rlitz, A., Theobalt, C., Nie{\ss}ner, M.,
  Klein, R., Kolb, A.: State of the art on {3D} reconstruction with {RGB-D}
  cameras. In: Computer graphics forum. vol.~37, pp. 625--652. Wiley Online
  Library (2018)

\end{thebibliography}
\newpage
\section{Pipeline and Algorithm}
In this section, we give a brief introduction for our overall pipleline of algorithms for training and rendering our hierarchical neural networks (\nascent). 

\paragraph{Hierarchical Neural Network Training}
Our method takes multiple viewpoint images $I_{gt}$ as input, and outputs the optimized octtree and contained neural networks $\model$. 
In \algref{alg:hnnt}, $S$ is the set of sampled points by using
sampling strategy in Sec. (3.3), and $I(r)$ is rendering RGB value for
ray direction $r$. The loss function $\mathcal{L}$ is the photometric
loss between rendered pixel values $\img$(r) and ground truth values
$\img_{gt}$(r). The loss is backpropated to each sub-network to update
the weights.
Every $T_{B}$ rounds of training, the octtree of scene will be updated
to adaptively reallocate computational resources to regions with high
density and high projected error, and the new sub-networks are
directly pre-trained by using stratified samples from the previous sub-networks (see Sec. (3.5) in main).

\begin{algorithm}
	\SetAlgoLined
	%\KwIn{image sequence $\{\mathcal{I}_n\}^{N}_{n=0}$, camera intrinsic $\{K_n\}^{N}_{n=0}$, camera extrinsic $\{[R_n|\mathbf{t}_n]\}^{N}_{n=0}$}
	\KwIn{viewpoint images $\img_{gt}$}
	\KwOut{$\model$}
	\KwResult{novel views $\img$}
	%\emph{special treatment of the first element of line $i$}\;
	\emph{Initialize $\mlp$, $\model$}\;
	\While{t $<$ T}
	{
		$S$ = \emph{Sampler}($\model$) \quad Sec. (3.3) in main\;
		$\img$(r) = \emph{Render}($S$) \quad \algref{alg:hnnr}\;
		loss = $\mathcal{L}$($\img$(r), $\img_{gt}$(r))\;
		\emph{BackPropagate}(loss)\;
		\emph{Step}($\mlp$)\;
		\If{$t \mod T_{B}=0$}
		{
			$\model$ = \emph{UpdateOctree}() \quad \eqnref{eqn:tree_opt}\;
			$\mlp$ = \emph{UpdateModel}($\mlp$, $\model$) \quad Sec. (3.5) in main\;
		}
		t = t + 1\;
	}
	\caption{Hierarchical Neural Network Training}\label{alg:hnnt}
\end{algorithm}

\paragraph{Hierachical Neural Network Rendering}
Rendering pipeline \algref{alg:hnnr} takes batches of samples $S$, hierarchical neural networks $\mlp$, and octtree models $\model$ as inputs. 
Sampling points $S$ are scheduled to corresponding sub-networks
$\mlp^{l}_{i}$ by their 3D sample location,. Samples are then
evaluates by the respective sub-network $\mlp^{l}_{i}$. To calculate
the ordered integral along the ray direction, samples are sorted by
Eqn. (4) in main and then composited by Eqn. (3) in main.

\begin{algorithm}
	\SetAlgoLined
	\KwIn{3D scene samples $S$, $\mlp$, $\model$}
	\KwOut{rendering value $\img$}
	\{$S^{l}_{i}$\} = \emph{Scheduler}($S$, $\model$)\; 
	$C = \{\}$, $D = \{\}$\;
	\For{$\mlp^{l}_{i}$ in $\mlp$}
	{
		$\rgb^{B^{l}_i}$, $\density^{B^{l}_i}$ = $\mlp^{l}_{i}(S^{l}_{i})$; \\
		C.add($\rgb^{B^{l}_i}$);\ D.add($\density^{B^{l}_i}$);\
	}
	$\{C, D\}$ = \emph{SortByZ}($\{C, D\}$); \quad Eqn. (4) in main\\ %\eqnref{eqn:sort_by_z}\\
	$\img(r)$ = \emph{Composite}(C, D); \quad Eqn. (3) in main \\%\eqnref{eq:composite}\\
	\caption{Hierarchical Neural Network Rendering}\label{alg:hnnr}
\end{algorithm}

\subsection{Importance Sampling}
\begin{figure}
	\centering
	\includegraphics[width=0.7\linewidth]{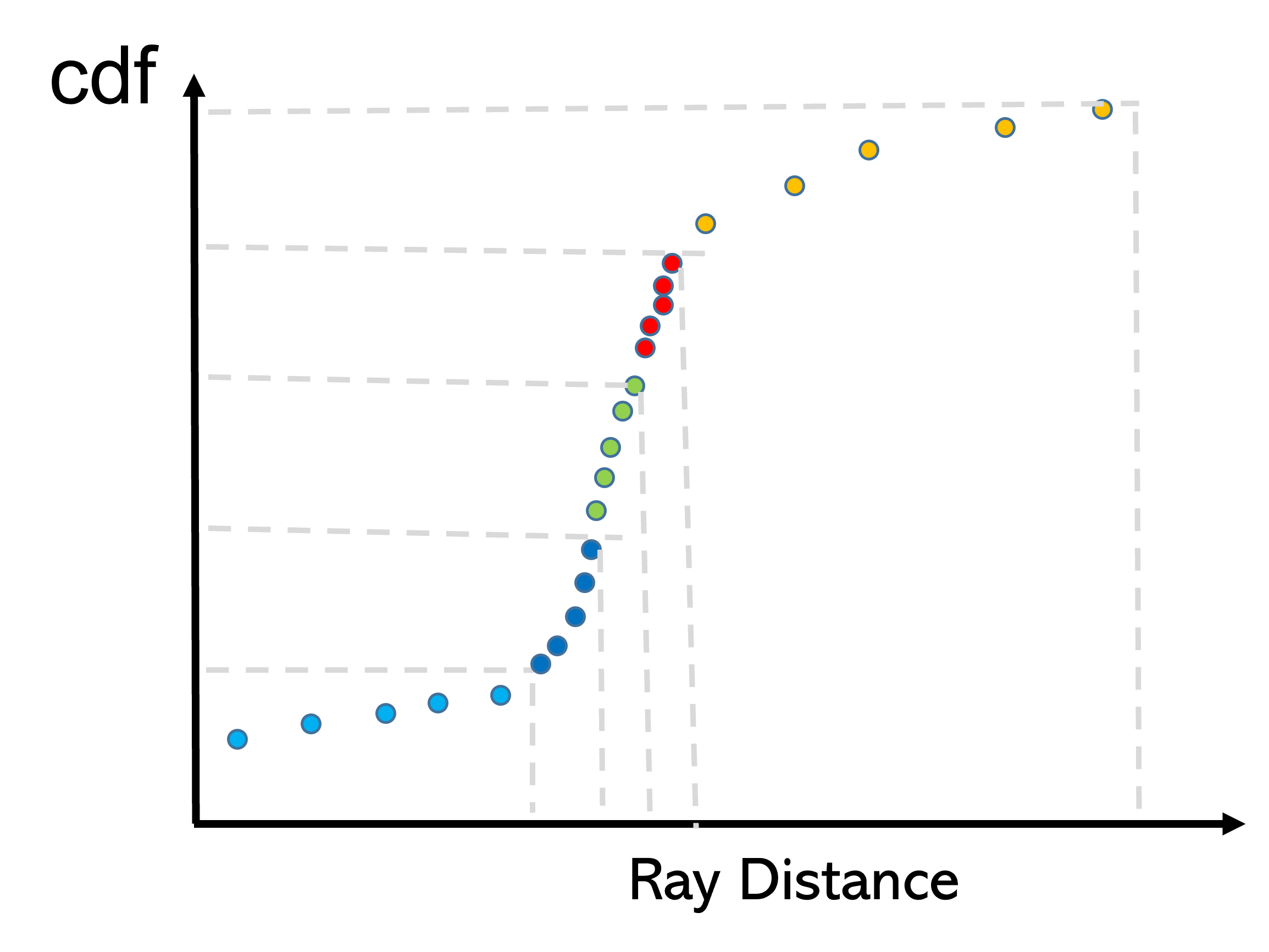}
	\caption{Illustration of sampling scheme that based on cumulative density field.}
	\label{fig:importancesampling}
\end{figure}

\figref{fig:importancesampling} gives a illustration of the
importance sampling in Sec. (3.3).

\section{OctTree Update Scheme}
In this section, we discuss the details of the octtree update scheme. 
The intuition of updating structural octtree are (1) avoid time-costly sampling and computation inside empty node, (2) reallocating representation (sub-networks) and computational (number of samples) resource to complex or poorly represented part of the scene.

Our objective mainly contains two parts, $\alpha_i$ is weighted
average alpha vector of node $i$ of sub-network, which indicates the
opaque of node, $\beta_i$ is projected rendering error vector of node
$i$, since flat or smooth surfaces may converge quickly and be
well-trained, while complex or poorly-represented scenes may still
need more epochs to obtain better quality. Therefore, the $\beta$ term will explore finer or coarser trees to encourage lower projected rendering error in octtree structure.  
Our objective is shown as, 

\begin{multline}\label{eqn:tree_opt}
\min \sum_{i} (1-\alpha^{\intercal}_i) I_i + \beta^{\intercal}_i I_i, 
\quad\text{s.t.}, 
\begin{cases}
I^{\uparrow}_i + I^{=}_i + I^{\downarrow}_i = 1, \\
\sum_i \frac{1}{N_c}I^{\uparrow}_i+I^{=}_i+N_c I^{\downarrow}_i \leq N_B, \\
\end{cases}
\end{multline}
where $I_i = [I^{\uparrow}_i, I^{=}_i, I^{\downarrow}_i]^{\intercal}$ are boolean flags of node operations, i.e., merge ($\uparrow$), split ($\downarrow$), and unchanged (=). $\alpha_i = [\alpha^{\uparrow}_i, \alpha^{=}_i, \alpha^{\downarrow}_i]^{\intercal}$ is the weighted average alpha in octree node i for three possible operations, if $\alpha_i$ . $\beta_i = [\beta^{\uparrow}_i, \beta^{=}_i, \beta^{\downarrow}_i]^{\intercal}$ is the weighted average projected rendering error respectively. $N_B$ is user-defined maximal block in system. 

%We formulate octree updating problem as Mixed Integer Problem (MIP) where each block has its 4 strategies to update its status, i.e., for any block we can use 4 binary variables to encode tree updating procedure, (1) merge them to form upper block ($I^{\uparrow}$), (2) stays unchanged ($I^{=}$), (3) split it to children block ($I^{\downarrow}$), (4) set to active or inactive ($I^{+}$). We define $S_i$ is the sample point set of $B_i$.
%
%The overall Mixed Integer Problem can be formulated as,
%\begin{multline}\label{eqn:tree_opt}
%\min \sum_{i} (1-\alpha^{\intercal}_i) I_i + \beta^{\intercal}_i I_i, \\
%\text{subject to}, 
%\begin{cases}
%I^{\uparrow}_i + I^{=}_i + I^{\downarrow}_i = 1, \\
%\sum_i \frac{1}{N_c}I^{\uparrow}_i+I^{=}_i+N_c I^{\downarrow}_i \leq N_B, \\
%\end{cases}
%\end{multline}
%where $I_i = [I^{\uparrow}_i, I^{=}_i, I^{\downarrow}_i]$. $\alpha_i = [\alpha^{\uparrow}_i, \alpha^{=}_i, \alpha^{\downarrow}_i]$ is the weighted average alpha in each octree block. $N_B$ is user-defined maximal block in system. 
%$w_i = [w^{\uparrow}_i, w^{=}_i, w^{\downarrow}_i]$ weighted average rendering error respectively. 
%
To calculate value $\alpha_i$, we first perform stratified sampling
from top to bottom in the octree hierarchy and predict the density
value for each sample by running the forward rendering network $\Phi(\mathbf{x}, \mathbf{d}) = (\rgb, \density)$, then the $\alpha_i^{=}$ for each block by,
\begin{multline}\label{eqn:alpha_eq_i}
\begin{cases}
\alpha^{=}_i = \frac{1}{|S_i|}\sum_{\mathbf{x}\in S_i} \delta(\mathbf{x}),\\
\alpha^{\uparrow}_i = \frac{1}{N_c}\alpha^{=}_{\mathcal{P}(i)},\\
\alpha^{\downarrow}_i = \sum_{j \in \mathcal{C}(i)}\alpha^{=}_j, \\
\end{cases}
\end{multline}
where $\mathcal{P}$ and $\mathcal{C}$ are query functions for octree
parent and child nodes. $S_i$ denotes the samples inside an active block.

To calculate $w_i$ for different cases, we first evaluate rendering error for each ray $E(\mathbf{r}) = \mathcal{L}(\img(\ray), \img_{gt}(\ray))$, and $\mathcal{L}$ is simple function for mean square error.

\begin{multline}\label{eqn:alpha_eq_i2}
\begin{cases}
\beta^{=}_i = \frac{1}{|S_i|}\sum_{\mathbf{x}\in S_i} \frac{w(x)}{W} E(\mathbf{r}),\\
\beta^{\uparrow}_i = \frac{1}{N_c}\beta^{=}_{\mathcal{P}(i)},\\
\beta^{\downarrow}_i = \sum_{j \in \mathcal{C}(i)}\beta^{=}_j, \\
\end{cases}
\end{multline}
where $w(x)$ is the weight in rendering function Eqn. (3) in the main
paper (i.e., $T_i$) for samples $x \in \mathbf{r}$. $W = \sum_{x \in
  \mathbf{r}} w(x)$ is the total sum of weights along the ray
direction. To optimize \eqnref{eqn:tree_opt}, we use {\tt or-tools} to solve MIP problems. 

%\paragraph*{Solve MIP problems} use ortools + extra update step for $I^{+}$.
%\todo{full sentence would be nice, or probably delete entirely}

\section{Additional Comparison and Results}

\begin{figure*}[th]
	\def \scale {0.18}
	\def \scaleB {0.99}
	\centering
	\subfigure[GT]{
		\begin{minipage}[t]{\scale\linewidth}
			\includegraphics[width=\scaleB\linewidth]{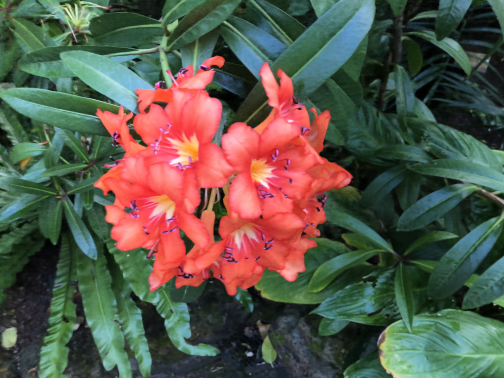}
			\color{white}
			\includegraphics[draft, width=\scaleB\linewidth]{figure/real_scene/flower/gt/gt_000}
			\includegraphics[width=\scaleB\linewidth]{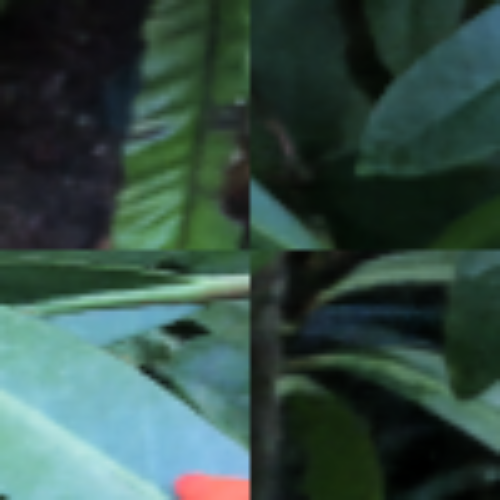}
			\color{white}
			\includegraphics[draft, width=\scaleB\linewidth]{figure/real_scene/flower/gt/gt_details}
			
			%			\color{white}
			%			\includegraphics[draft, width=\scaleB\linewidth]{figure/extreme_view_llff/nerf/001_b}
			%			\color{white}
			%			\includegraphics[draft, width=\scaleB\linewidth]{figure/extreme_view_llff/nerf/001_01}
			%			\color{white}
			%			\includegraphics[draft, width=\scaleB\linewidth]{figure/extreme_view_llff/nerf/001_b}
			%			\color{white}
			%			\includegraphics[draft, width=\scaleB\linewidth]{figure/extreme_view_llff/nerf/001_01}
			
	\end{minipage}}
	\subfigure[NeRF\cite{mildenhall2020nerf}]{
		\begin{minipage}[t]{\scale\linewidth}
			\includegraphics[width=\scaleB\linewidth]{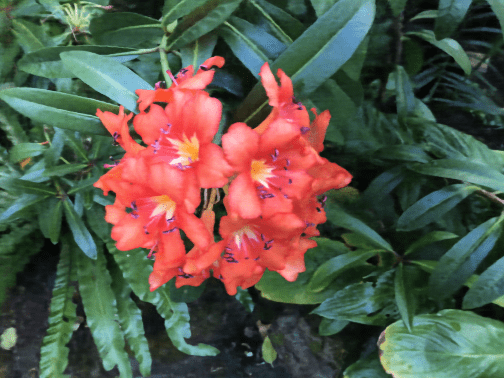}
			\includegraphics[width=\scaleB\linewidth]{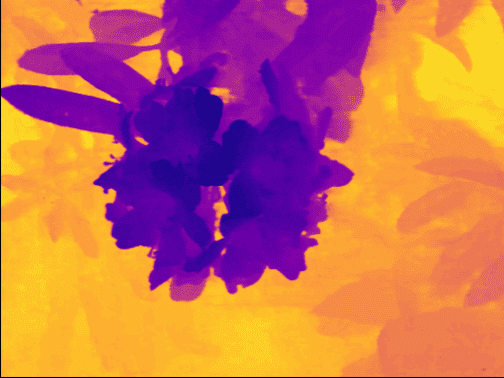}
			\includegraphics[width=\scaleB\linewidth]{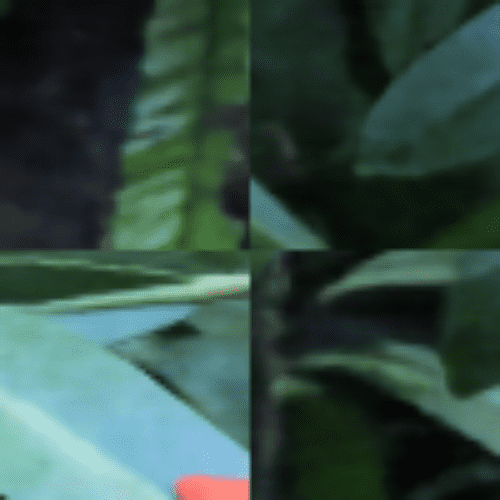}
			\includegraphics[width=\scaleB\linewidth]{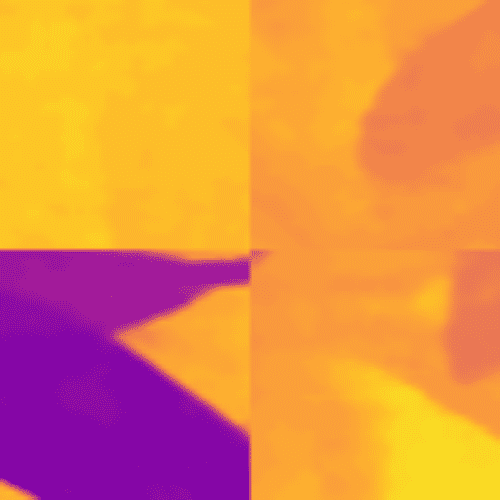}
	\end{minipage}}
	\subfigure[KiloNeRF\cite{Reiser2021ICCV}]{
		\begin{minipage}[t]{\scale\linewidth}
			\includegraphics[width=\scaleB\linewidth]{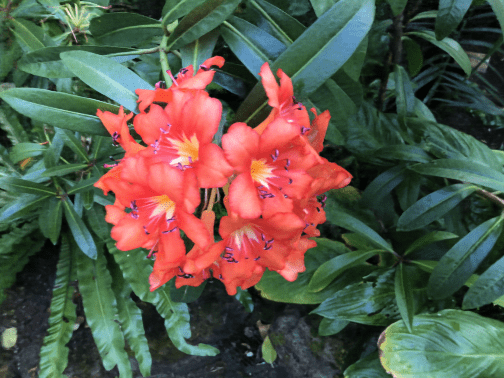}
			\includegraphics[width=\scaleB\linewidth]{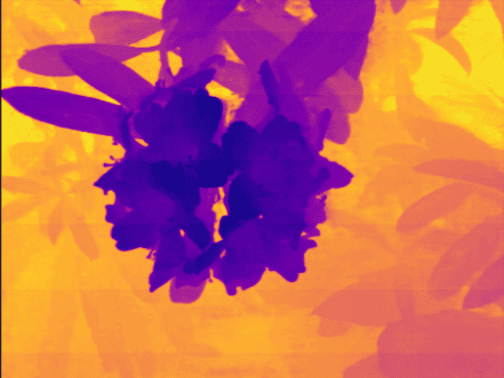}
			\includegraphics[width=\scaleB\linewidth]{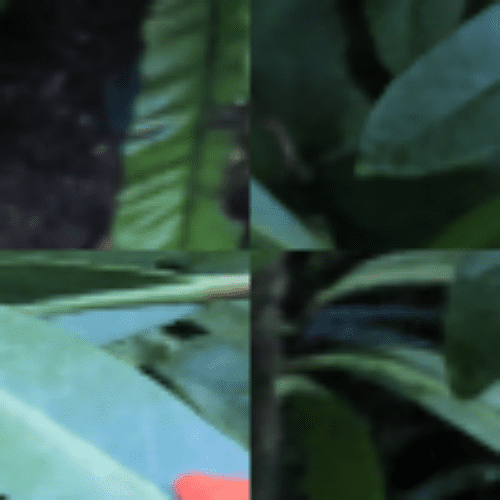}
			\includegraphics[width=\scaleB\linewidth]{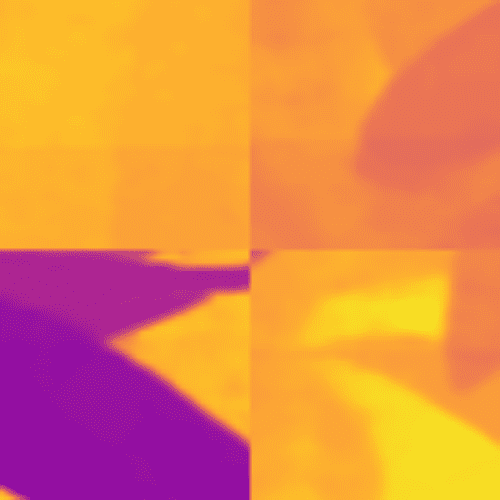}
	\end{minipage}}
	%	\subfigure[NSVF]{
	%		\begin{minipage}[t]{\scale\linewidth}
	%			\includegraphics[width=\scaleB\linewidth]{figure/real_scene/flower/nsvf/000}
	%			\includegraphics[width=\scaleB\linewidth]{figure/real_scene/flower/nsvf/000}
	%			%\includegraphics[width=\scaleB\linewidth]{figure/real_scene/horns/nsvf/000}
	%			%\includegraphics[width=\scaleB\linewidth]{figure/real_scene/leaves/nsvf/000}
	%\end{minipage}}
	\subfigure[Our]{
		\begin{minipage}[t]{\scale\linewidth}
			\includegraphics[width=\scaleB\linewidth]{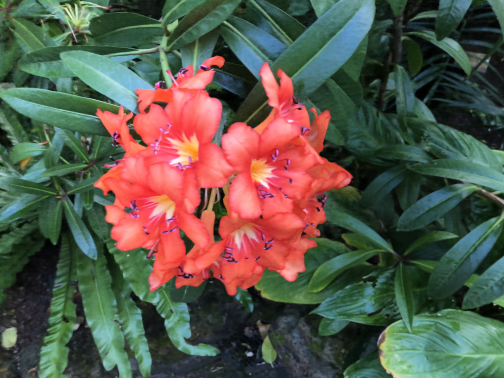}
			\includegraphics[width=\scaleB\linewidth]{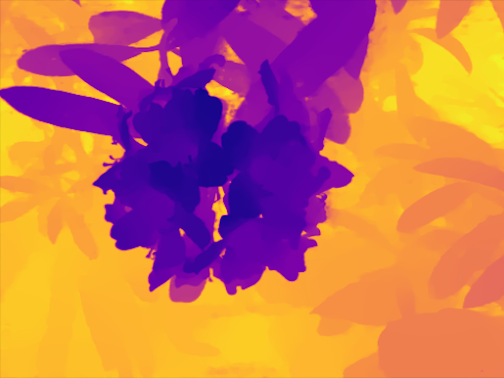}
			\includegraphics[width=\scaleB\linewidth]{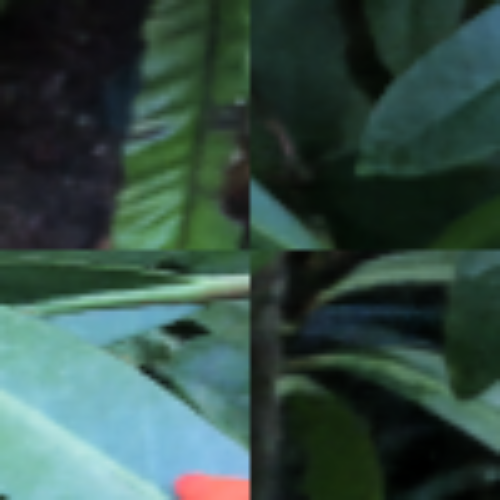}
			\includegraphics[width=\scaleB\linewidth]{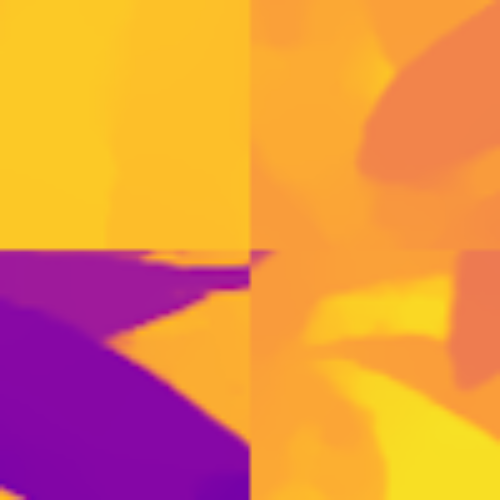}
	\end{minipage}}
	\subfigure[Octree]{
		\begin{minipage}[t]{\scale\linewidth}
			\includegraphics[width=\scaleB\linewidth]{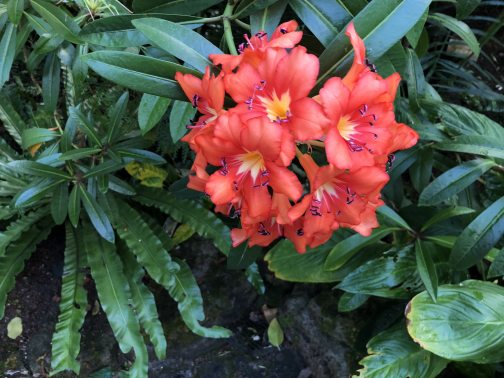}
			\includegraphics[width=0.91\linewidth]{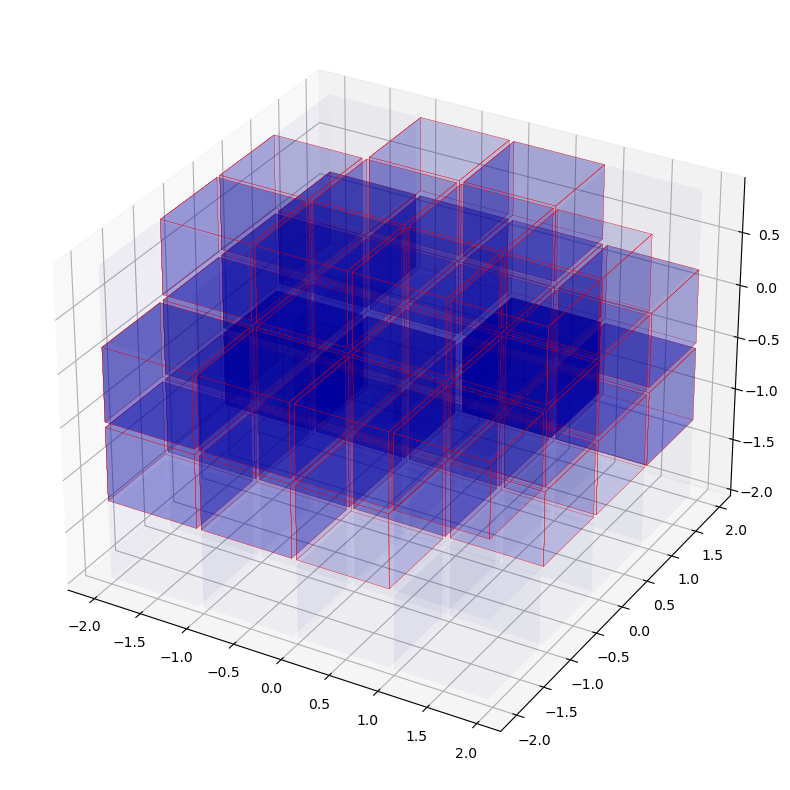}
			\includegraphics[width=0.91\linewidth]{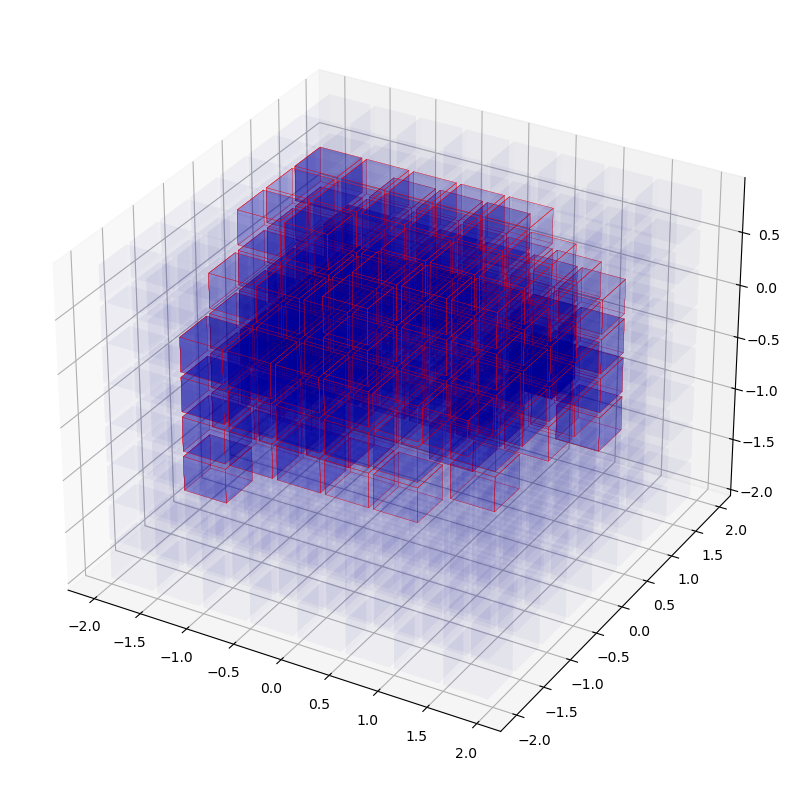}
			\includegraphics[width=0.91\linewidth]{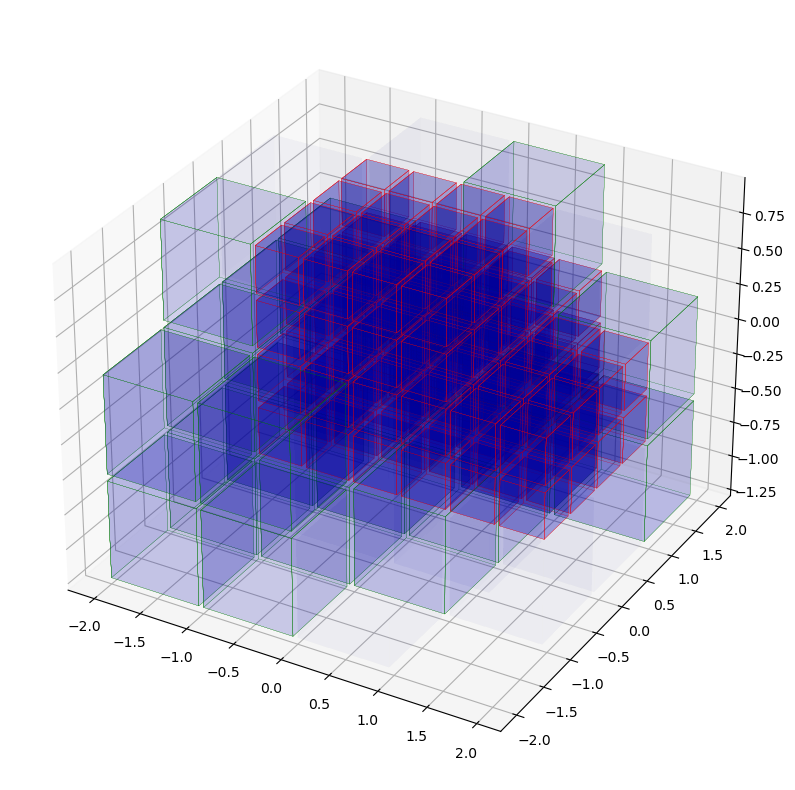}
	\end{minipage}}
	\caption{Visual Comparison on LLFF-NeRF dataset~\cite{mildenhall2019llff}.(a) is ground truth view of flower scene with highlighting details. (b)-(d) are the novel view of NeRF~\cite{mildenhall2020nerf}, KiloNeRF~\cite{Reiser2021ICCV} and our methods with highlight details. (e) the visualization of example view and octree optimization process from initial level 2 to level 3, and merge to simple structure to save computational and sampling resource.}
	\label{fig:compare_real_scene}
\end{figure*}

In this section, we show extensive comparison in details and results.
As discussed in the main text, for
views close to the training views all methods produce visually very
similar results; differences only become apparent at close inspection
and when analyzing depth structure. However, as the results in the
main paper show, the differences in the depth estimation amplify the
visual quality differences for extrapolated views far from the
training data.

\figref{fig:compare_real_scene} shows visual comparisons of novel view
synthesis on real scenes from the LLFF-NeRF
dataset~\cite{mildenhall2019llff}.  As we can see in the figure,
NeRF~\cite{mildenhall2020nerf} can miss surfaces with its sampling
process so that back surfaces can ``shine through''. This is due
NeRF's sampling scheme that applies stratified search for a
coarse-level density distribution estimation and then sampling
according to coarse density distribution along the ray to give more
samples near the object surface. Thus, for a thin structure, the
coarse level search may missing the important part of scene, leads to
leaking light effect in rendering novel view. KiloNeRF only applies
dense stratified sampling inside each block, and collects output
samples along the ray. Thus, blocking-effect are again visible just
like in the synthetic data. Our method achieves sharper novel depth
map, and the light-weight sub-network enables a dense coarse level
surface search, alleviate blocking effect as well as light leakage.

\begin{figure*}[htb]
	\def \scale {0.18}
	\def \scaleB {0.99}
	\centering
	\subfigure[GT]{
		\begin{minipage}[t]{\scale\linewidth}
			\includegraphics[width=\scaleB\linewidth]{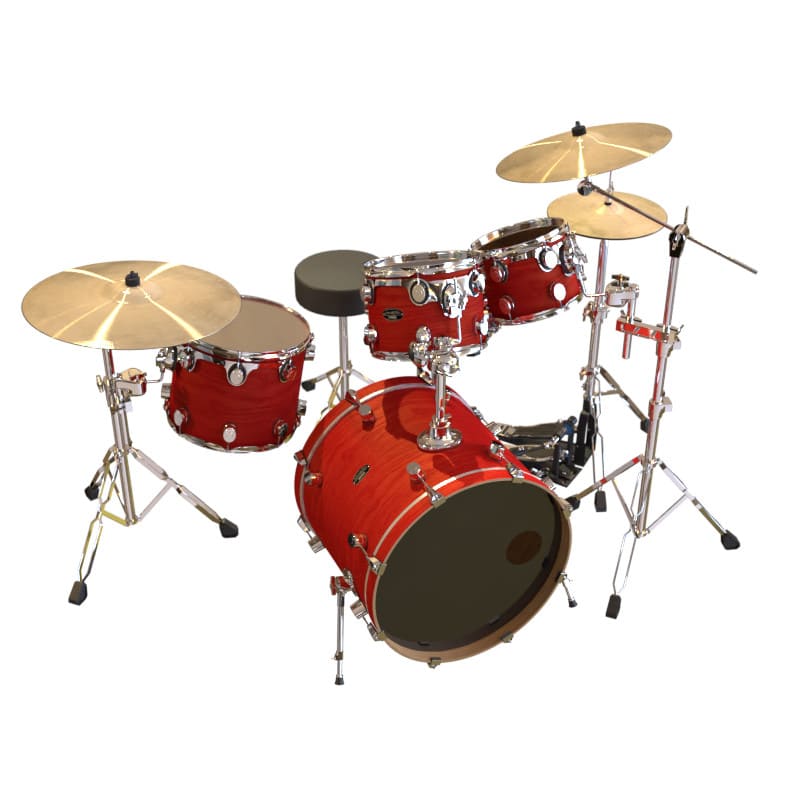}
			\color{white}
			\includegraphics[draft, width=\scaleB\linewidth]{figure/synthetic/drums/gt/r_43}
			\includegraphics[ width=\scaleB\linewidth]{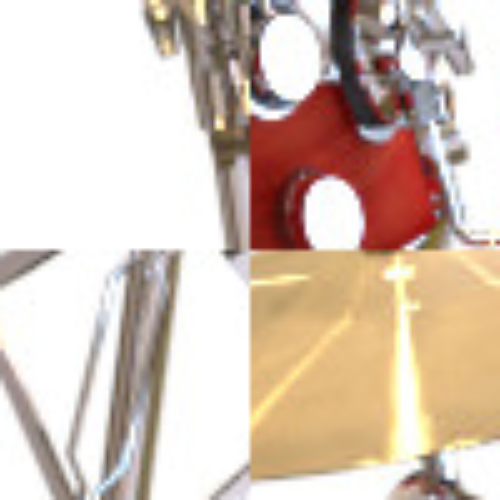}
			\color{white}
			\includegraphics[draft, width=\scaleB\linewidth]{figure/synthetic/drums/gt/gt_details}
	\end{minipage}}
	\subfigure[NeRF\cite{mildenhall2020nerf}]{
		\begin{minipage}[t]{\scale\linewidth}
			\includegraphics[width=\scaleB\linewidth]{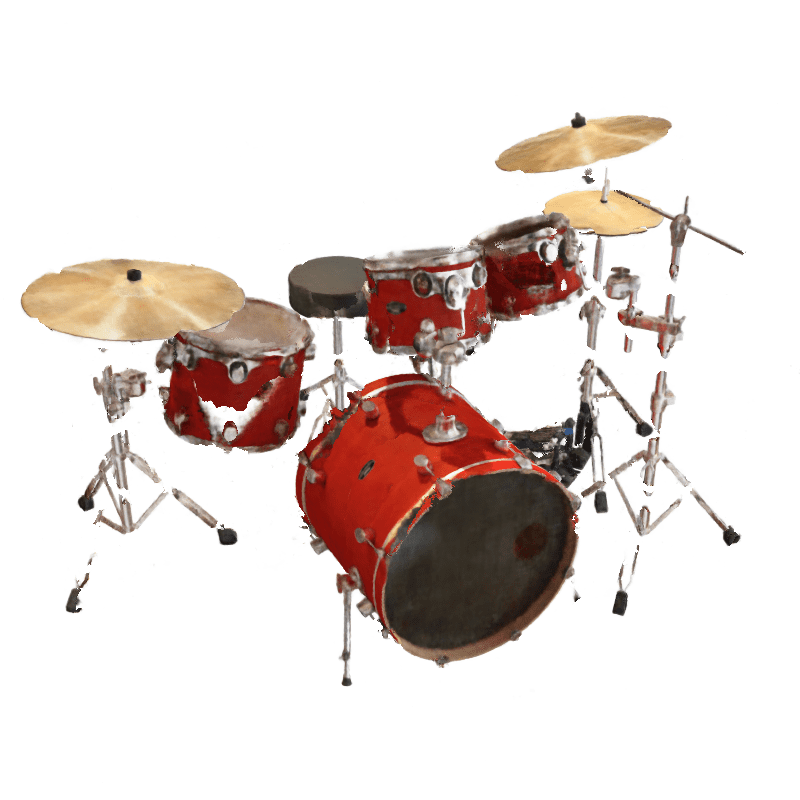}
			\includegraphics[width=\scaleB\linewidth]{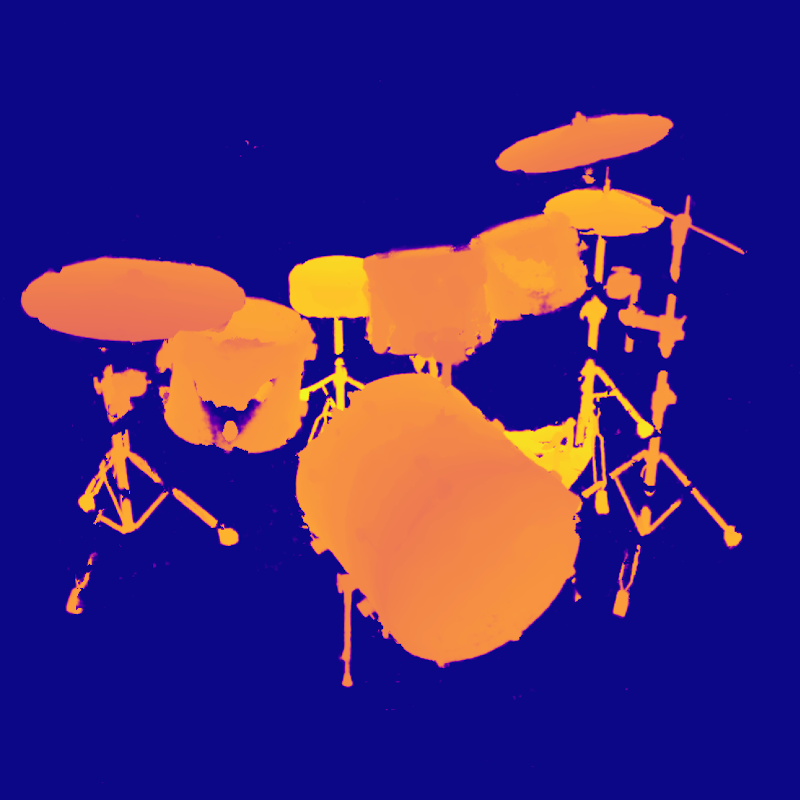}
			\includegraphics[width=\scaleB\linewidth]{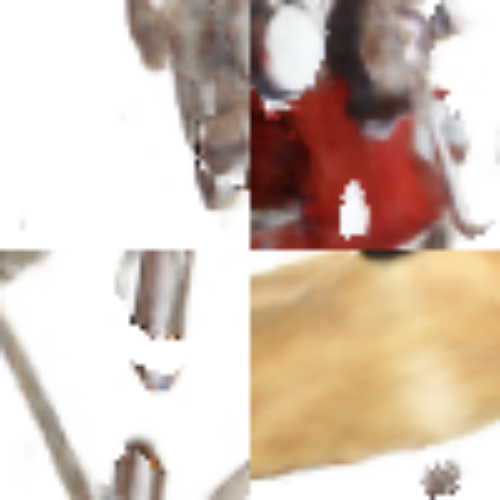}
			\includegraphics[width=\scaleB\linewidth]{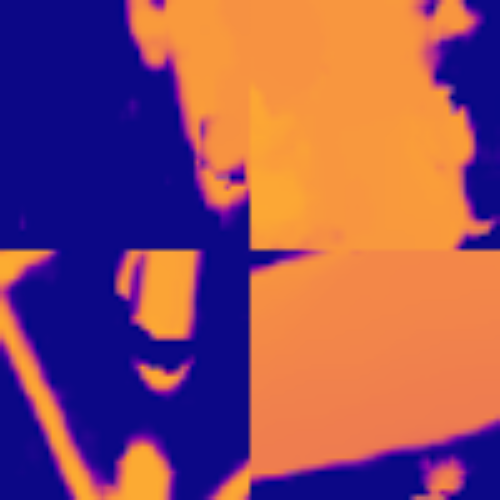}
	\end{minipage}}
	\subfigure[KiloNeRF\cite{Reiser2021ICCV}]{
		\begin{minipage}[t]{\scale\linewidth}
			\includegraphics[width=\scaleB\linewidth]{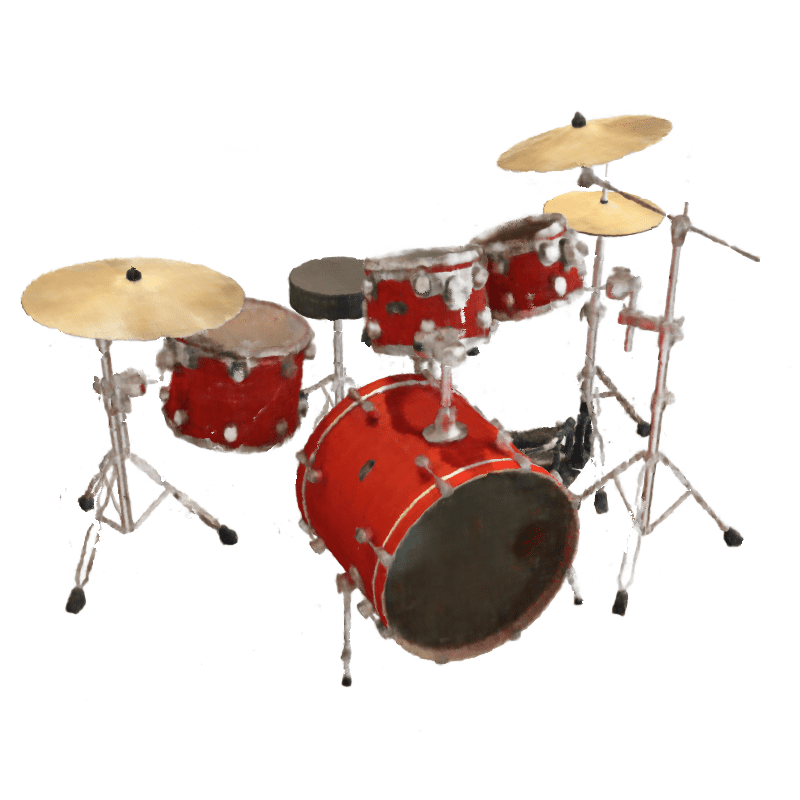}
			\includegraphics[width=\scaleB\linewidth]{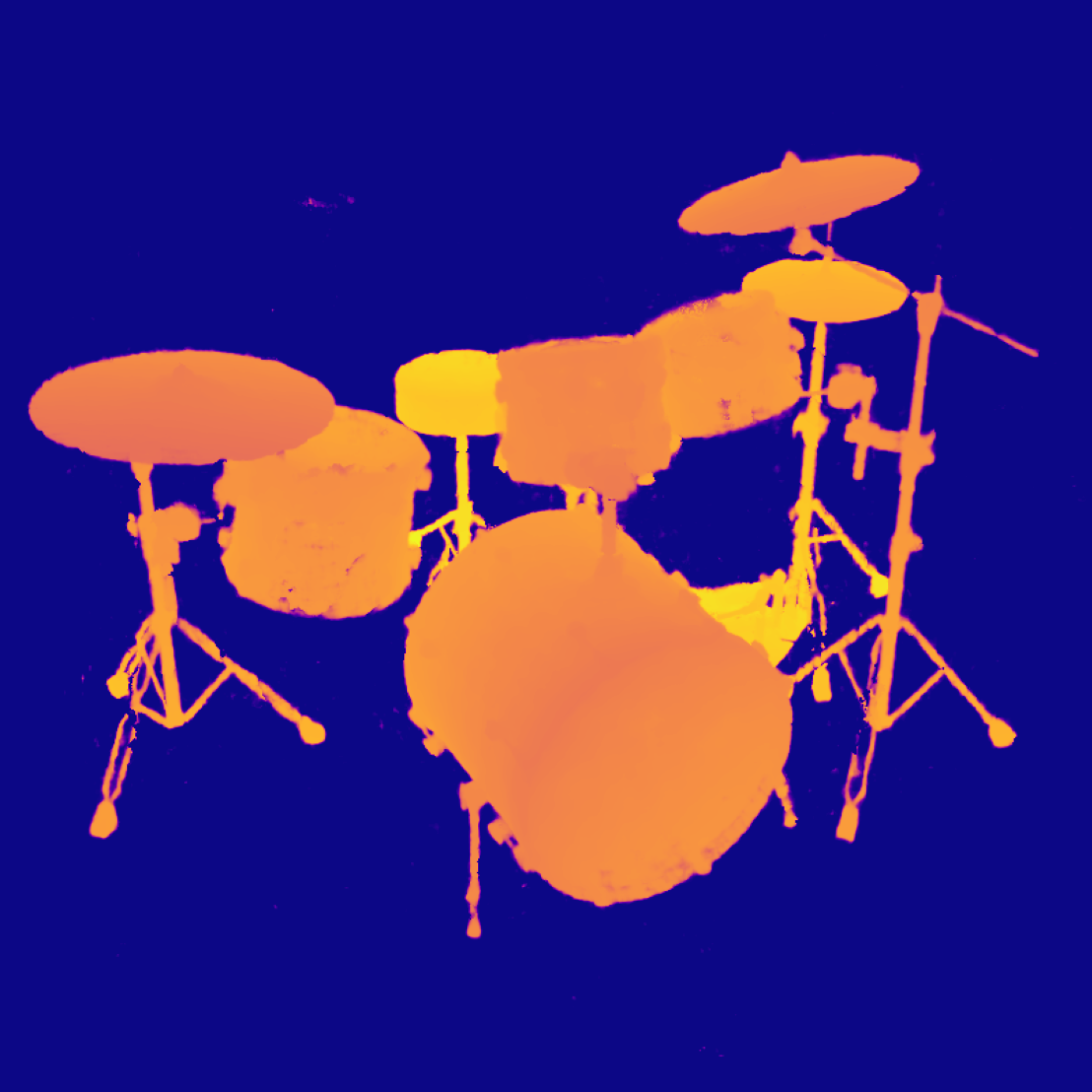}
			\includegraphics[width=\scaleB\linewidth]{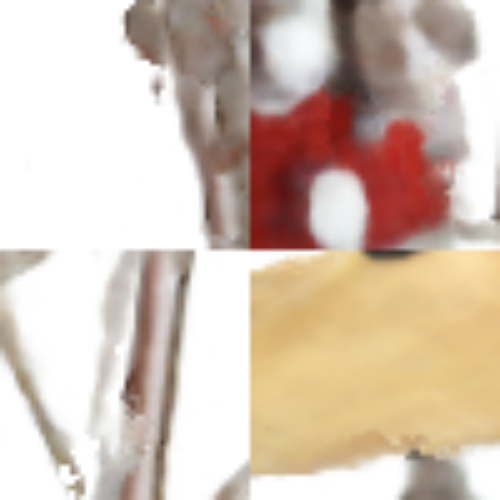}
			\includegraphics[width=\scaleB\linewidth]{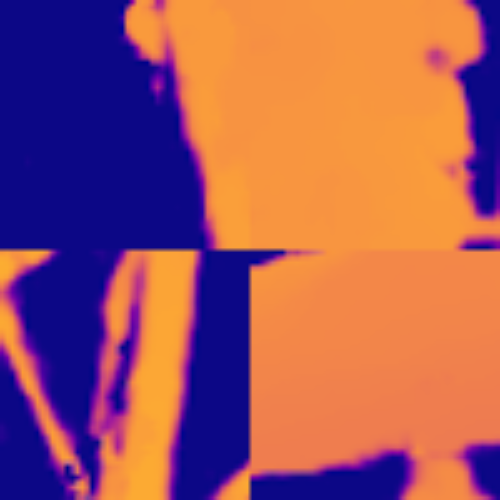}
	\end{minipage}}
	\subfigure[Our]{
		\begin{minipage}[t]{\scale\linewidth}
			\includegraphics[width=\scaleB\linewidth]{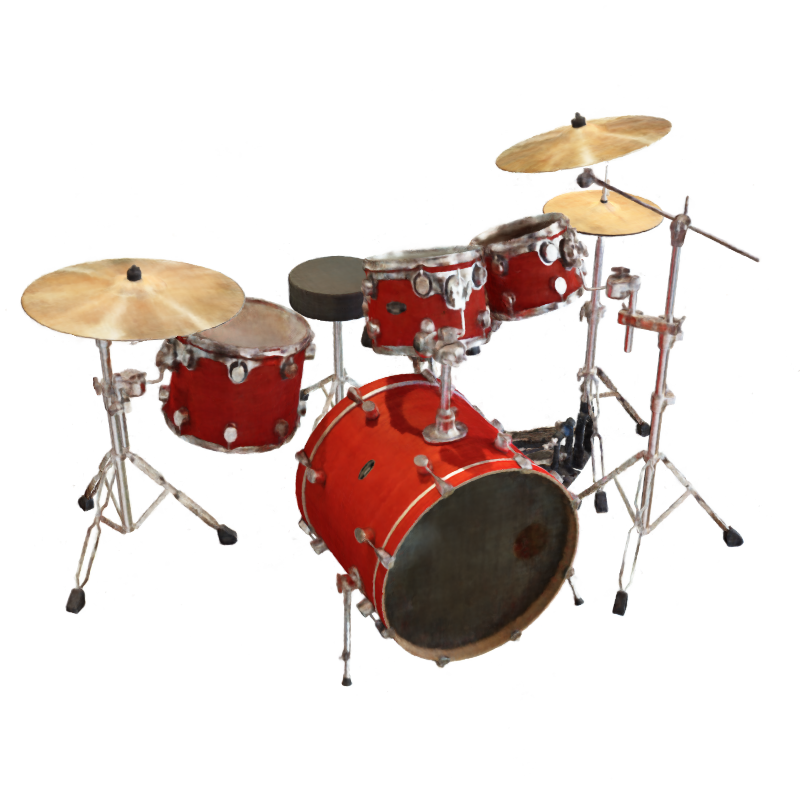}
			\includegraphics[width=\scaleB\linewidth]{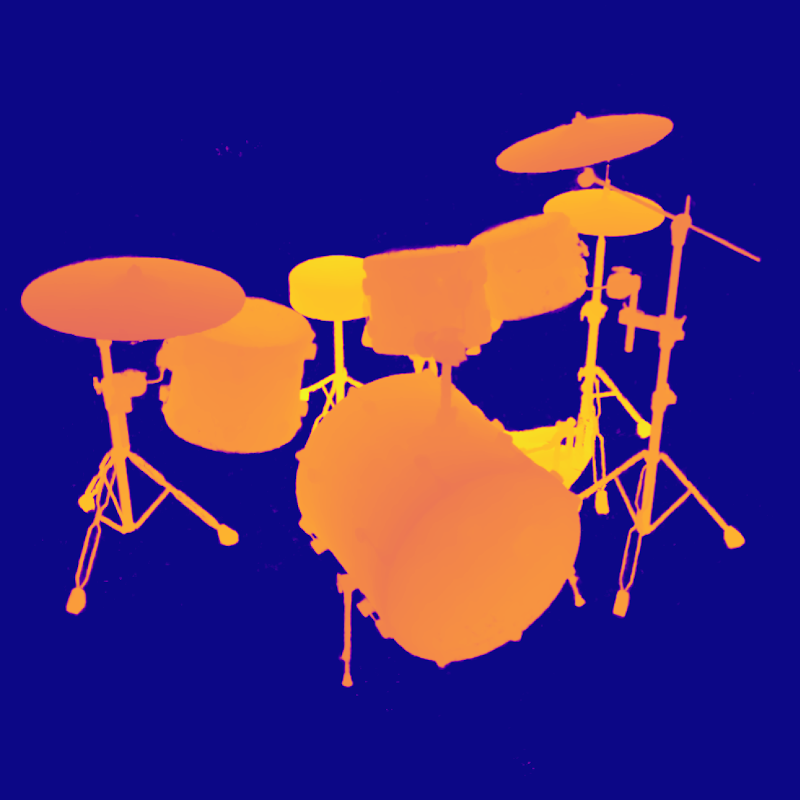}
			\includegraphics[width=\scaleB\linewidth]{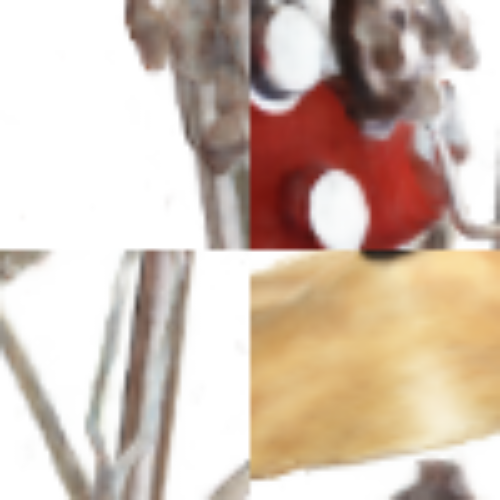}
			\includegraphics[width=\scaleB\linewidth]{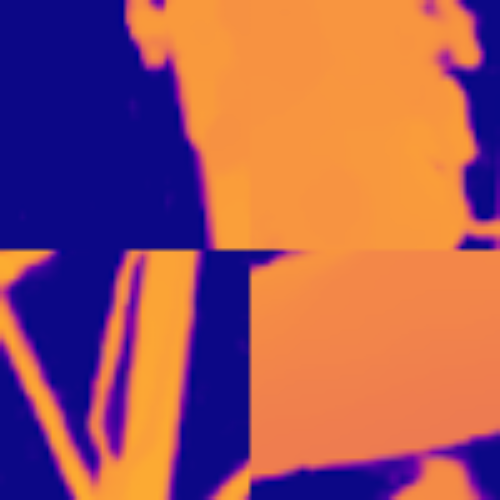}
	\end{minipage}}
	\subfigure[Octree]{
		\begin{minipage}[t]{\scale\linewidth}
			\includegraphics[width=\scaleB\linewidth]{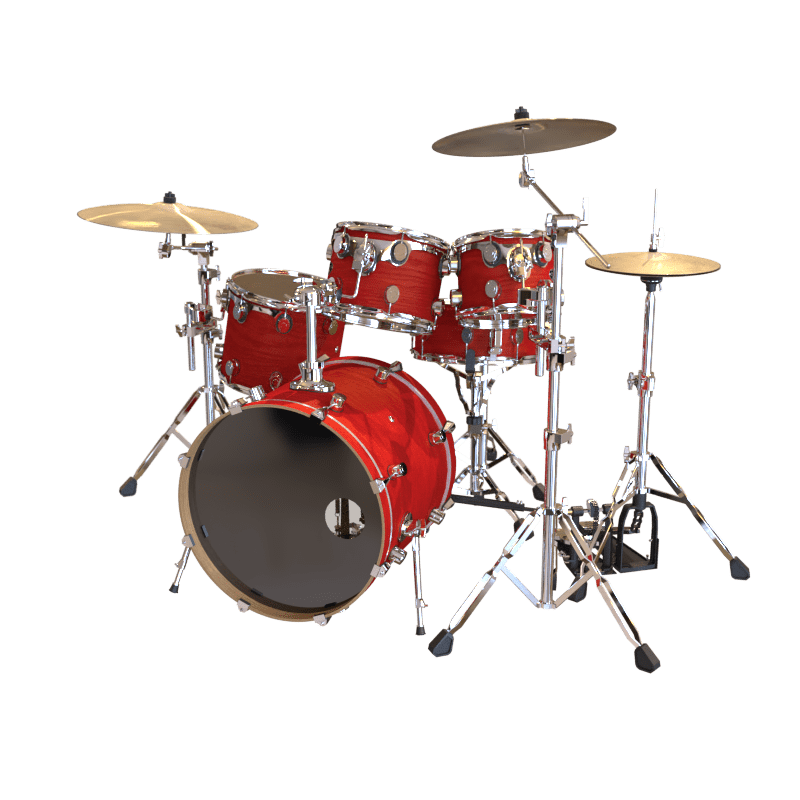}
			\includegraphics[width=\scaleB\linewidth]{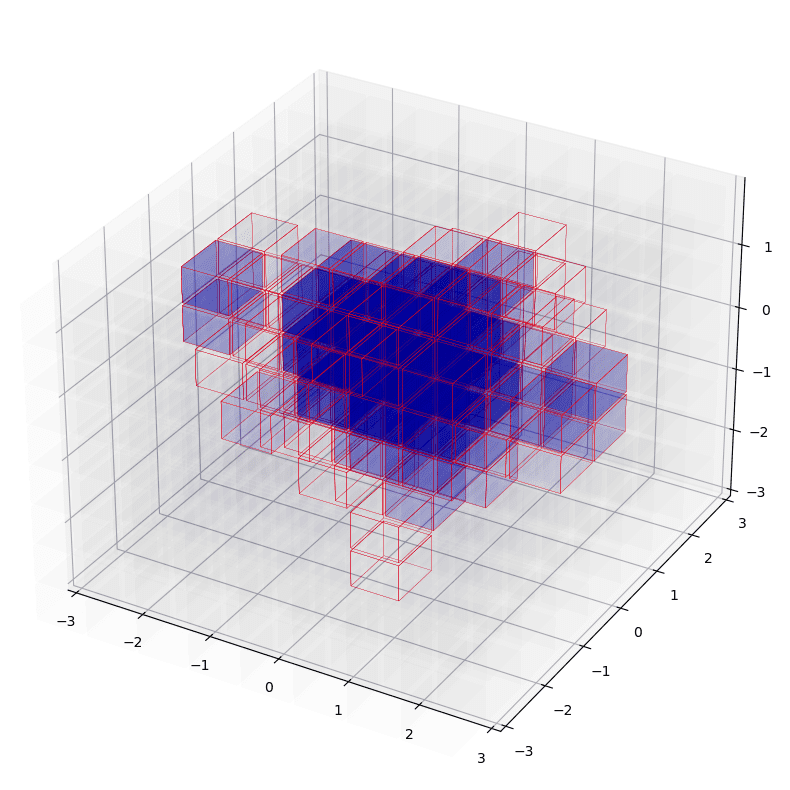}
			\includegraphics[width=\scaleB\linewidth]{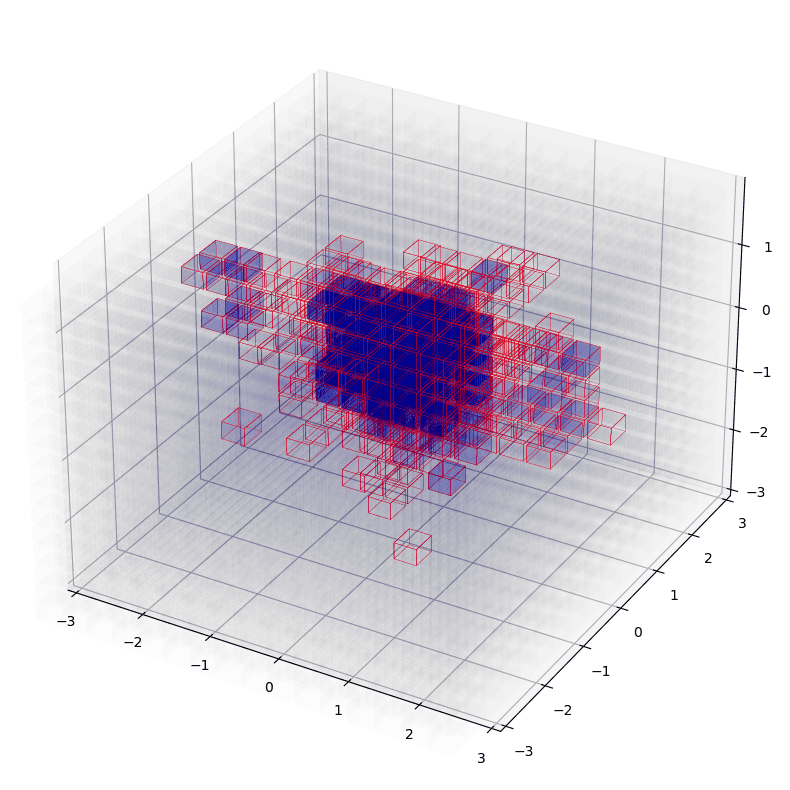}
			\includegraphics[width=\scaleB\linewidth]{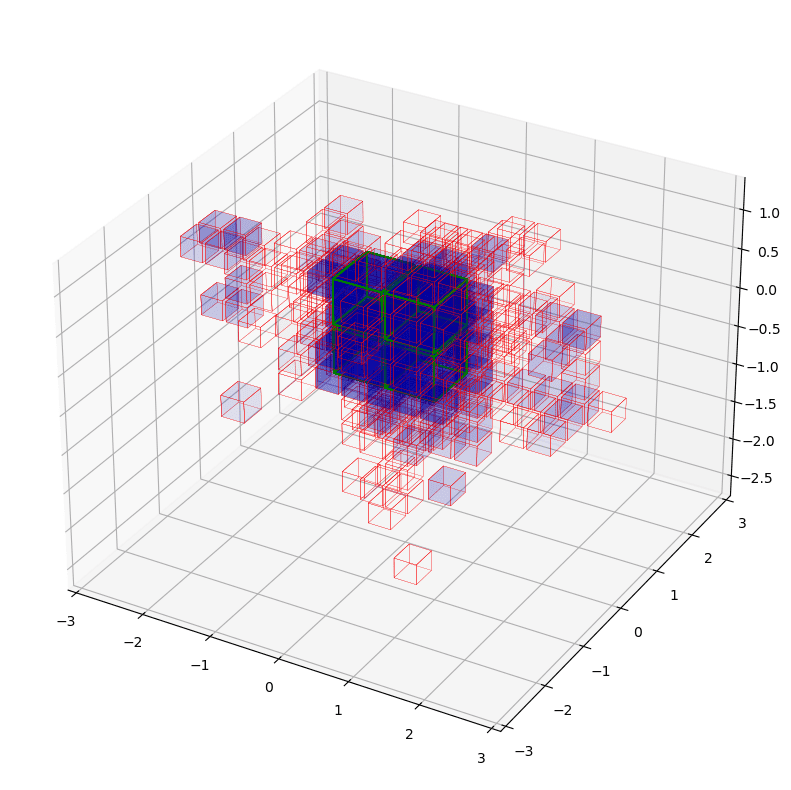}
	\end{minipage}}
	\caption{Visual Comparison on synthetic NeRF
		dataset~\cite{mildenhall2020nerf}. (a) is ground truth view
		with zoom in details. (b), (c), (d) are state-of-the-arts
		methods of NeRF~\cite{mildenhall2020nerf},
		KiloNeRF~\cite{Reiser2021ICCV}, and our proposed method with
		highlighted detail regions. (e) illustrate example view and
		octree optimization process from coarse to fine (Blocks with
		green line indicates block merging, and red line indicates
		blocks splitting).}
	\label{fig:compare_synthetic_scene}
\end{figure*}

\figref{fig:compare_synthetic_scene} shows visual comparisons on the
synthetic NeRF dataset. All the state-of-the-art methods achieve
reasonable performance in rendering novel views from camera positions
close to the training views. However, in visualizations of the depth
map scene, NeRF~\cite{mildenhall2020nerf} shows blurring and
topological artifacts in the depth, indicating that the actual 3D
structure is less accurate. As we show below, this has an impact on
the quality of extrapolated views far from the training data.
KiloNeRF~\cite{Reiser2021ICCV} shows high quality results in the RGB
view, but exhibits a slight blocking effect when visualizing the depth
view, since KiloNeRF's sub-network are pre-trained by a global
network, i.e., NeRF, each sub-network is independent in the
pre-training stage and mix the query results in the fine-tuning stage,
which introduces a discontinuity. Our method takes advantage of the
tree structure for flexible and scalable representation with an
adaptive training scheme for computational resource allocation, all
the sub-networks are trained from coarse to fine. Therefore, no
pre-training is required, and back propagation will update all
sub-network along the integral ray direction, which shows consistent
and smooth rendering results in both RGB view and depth.
To better illustrate our octree-based representation, we also show a
progression of the octree update last column (ground truth rendering
on top).

%\begin{figure}[h]
%	\includegraphics[width=0.9\linewidth]{figure/uav/uav_camera/Island_Mosque_Flight_01_00524}\\
%	%\includegraphics[width=0.45\linewidth]{figure/uav/uav_camera/Island_Mosque_Flight_01_00524}\\
%	%\includegraphics[width=0.45\linewidth]{figure/uav/uav_camera/Island_Mosque_Flight_01_00524}
%	\includegraphics[width=0.9\linewidth]{figure/uav/uav_camera/kaust_s01}
%	\caption{Illustration of UAV view camera poses and an example
%		image from the dataset. Due to expensive ground scanning and
%		storage limitation of UAV, such datasets are usually sparse
%		with respect to the number of views.}
%	\label{fig:uav_camera_pose_kaust}
%\end{figure}

\subsection{Extreme Novel View Comparison}
\begin{figure*}
	\def \scale {0.22}
	\def \scaleB {1.0}
	\centering
	\subfigure[NeRF\cite{mildenhall2020nerf}]{
		\begin{minipage}[t]{\scale\linewidth}
			\includegraphics[width=\scaleB\linewidth]{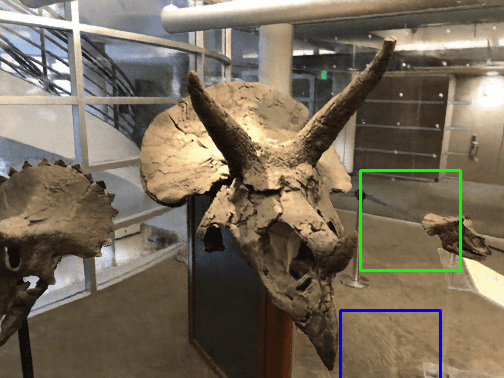}
			\includegraphics[width=\scaleB\linewidth]{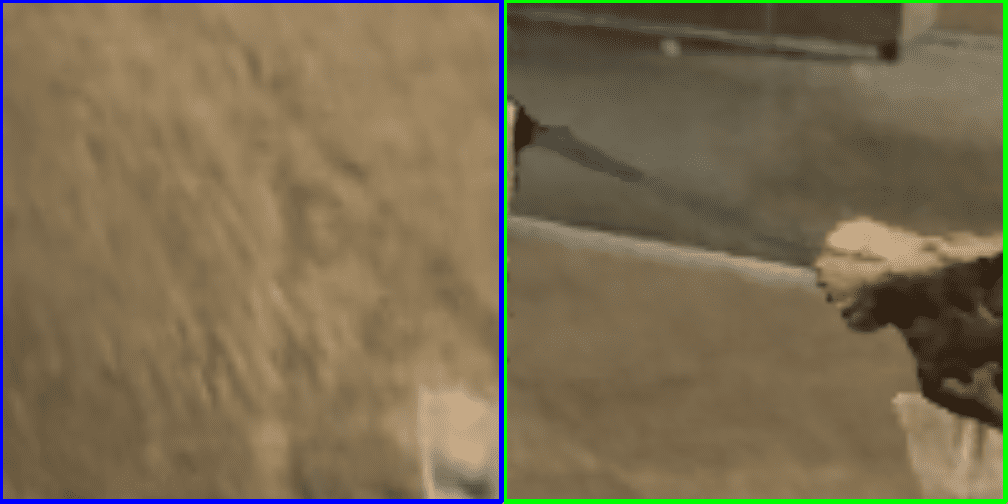}
			\includegraphics[width=\scaleB\linewidth]{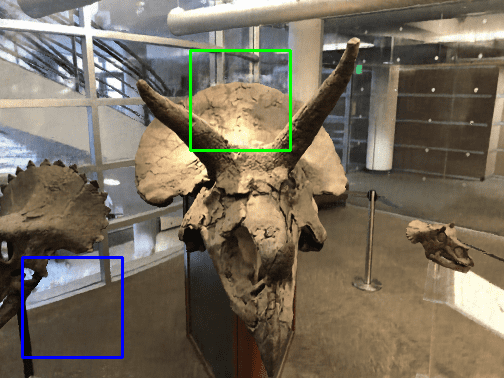}
			\includegraphics[width=\scaleB\linewidth]{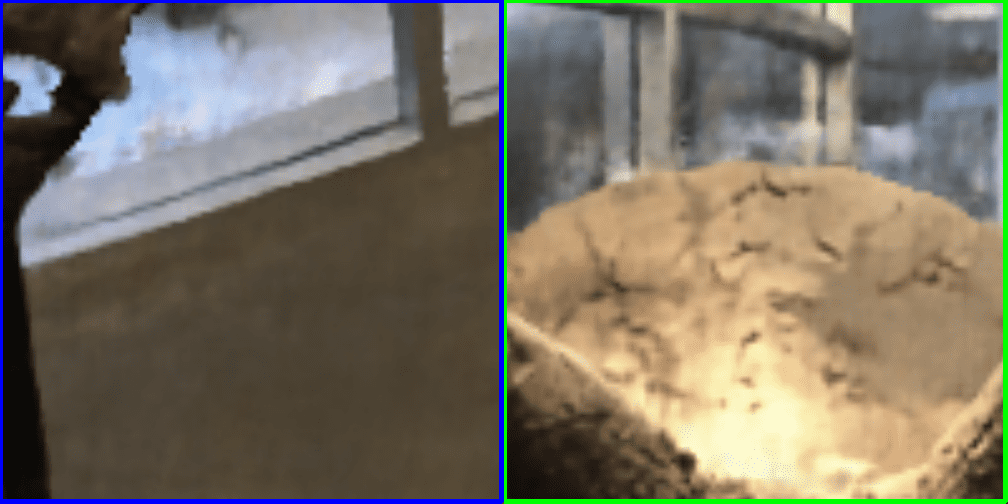}
			\includegraphics[width=\scaleB\linewidth]{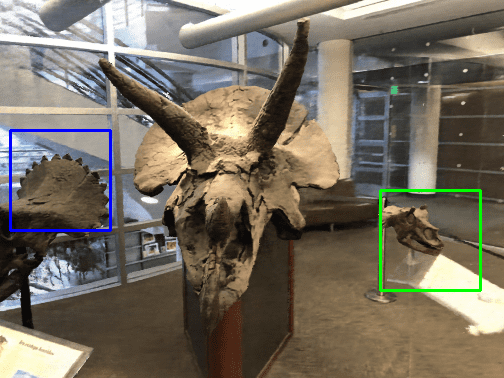}
			\includegraphics[width=\scaleB\linewidth]{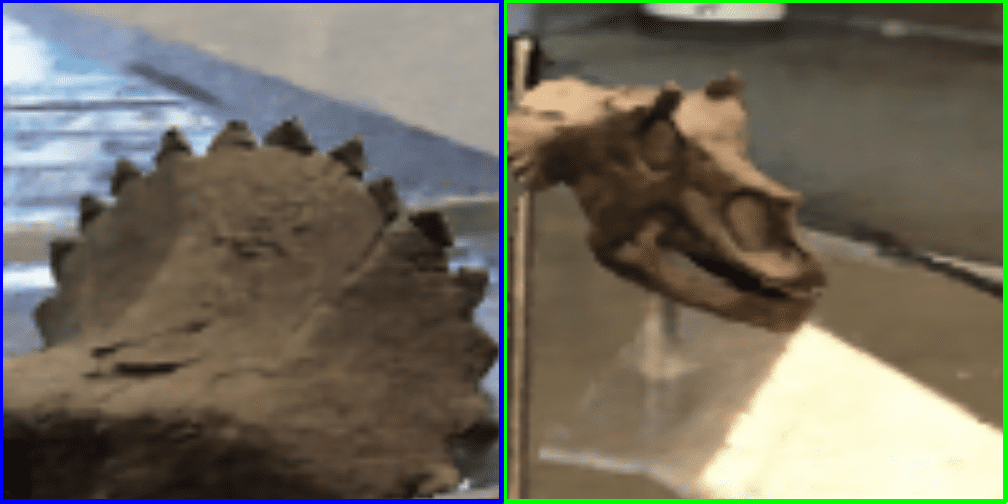}
			\includegraphics[width=\scaleB\linewidth]{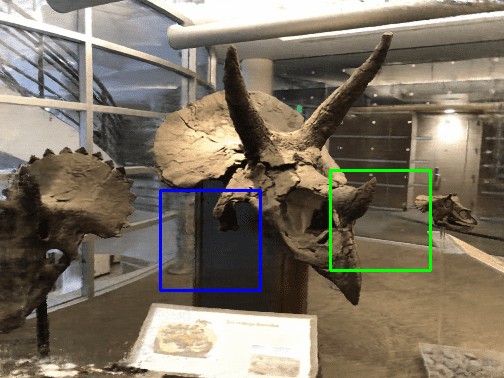}
			\includegraphics[width=\scaleB\linewidth]{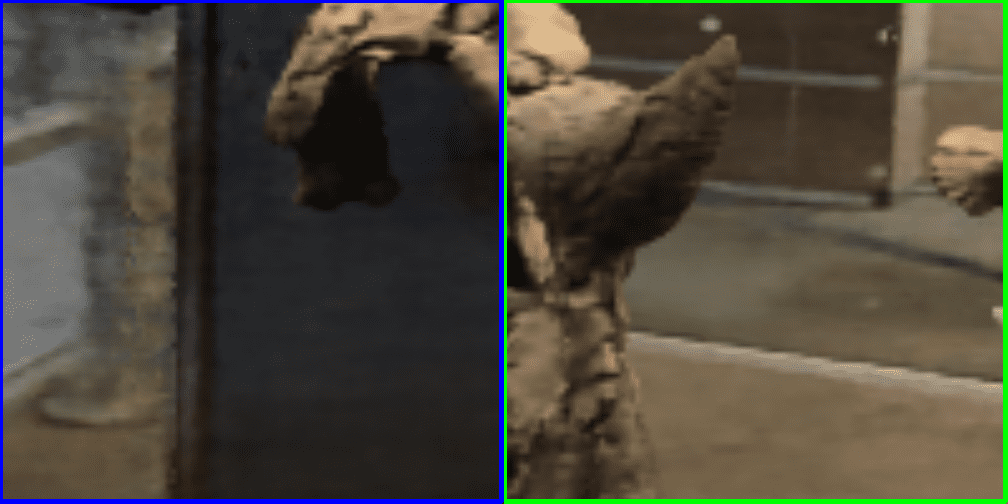}
	\end{minipage}}
	\subfigure[KiloNeRF]{
		\begin{minipage}[t]{\scale\linewidth}
			\includegraphics[width=\scaleB\linewidth]{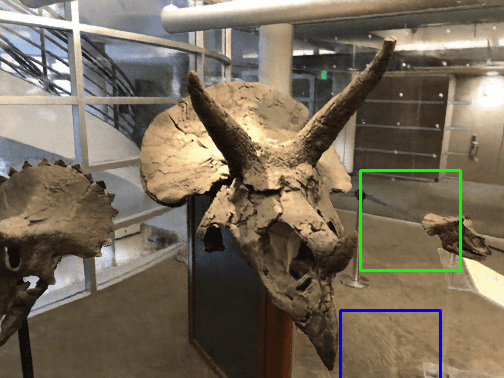}
			\includegraphics[width=\scaleB\linewidth]{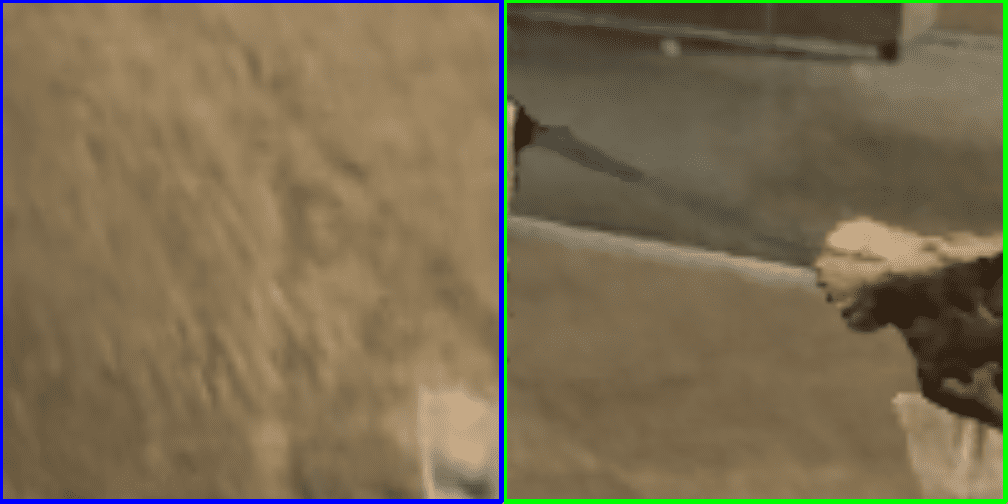}
			\includegraphics[width=\scaleB\linewidth]{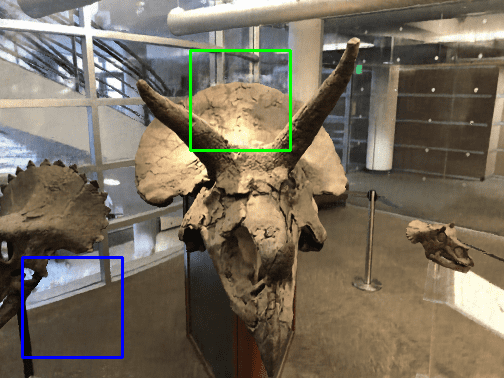}
			\includegraphics[width=\scaleB\linewidth]{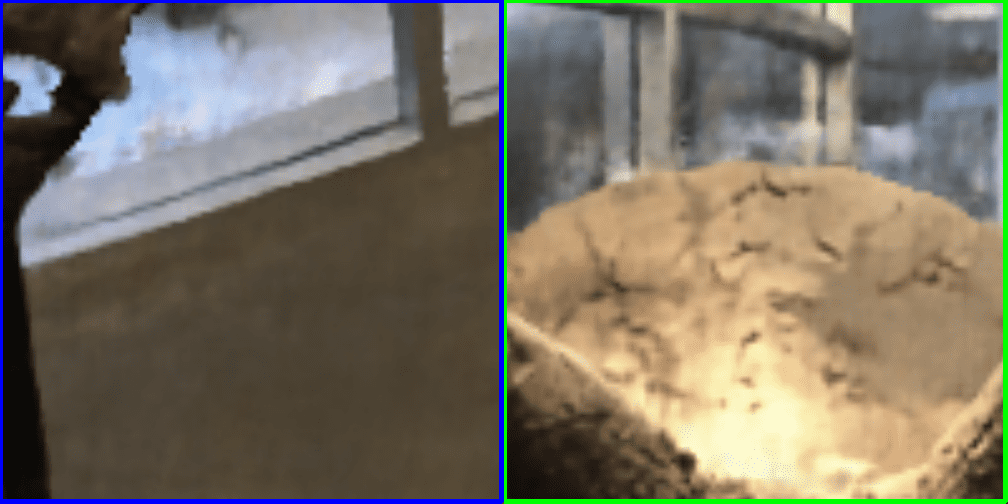}
			\includegraphics[width=\scaleB\linewidth]{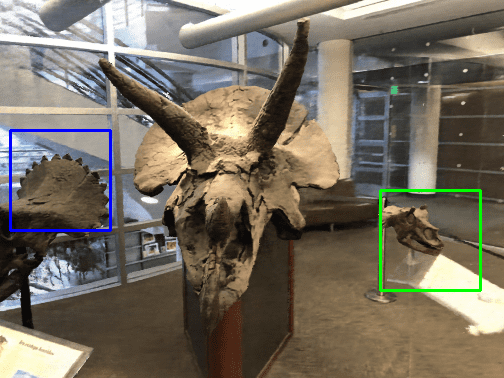}
			\includegraphics[width=\scaleB\linewidth]{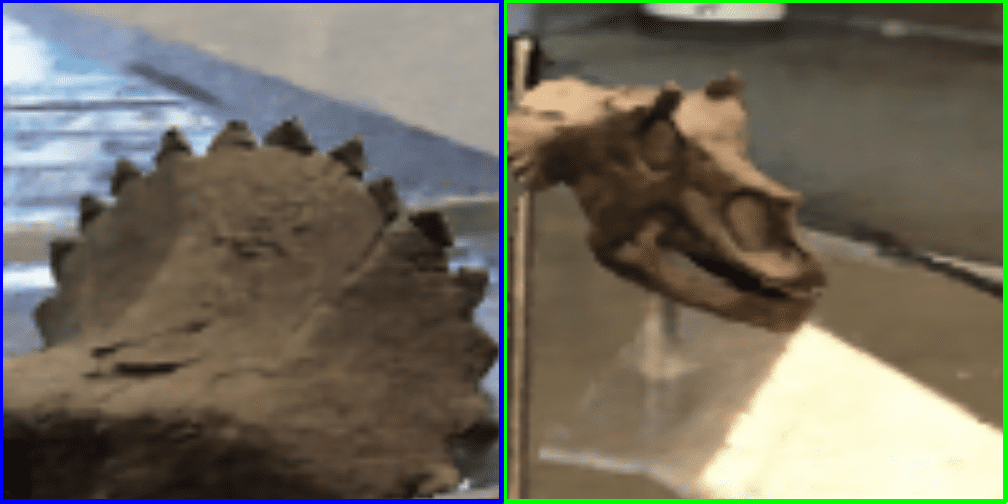}
			\includegraphics[width=\scaleB\linewidth]{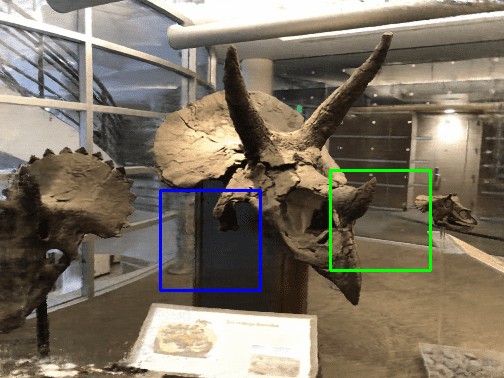}
			\includegraphics[width=\scaleB\linewidth]{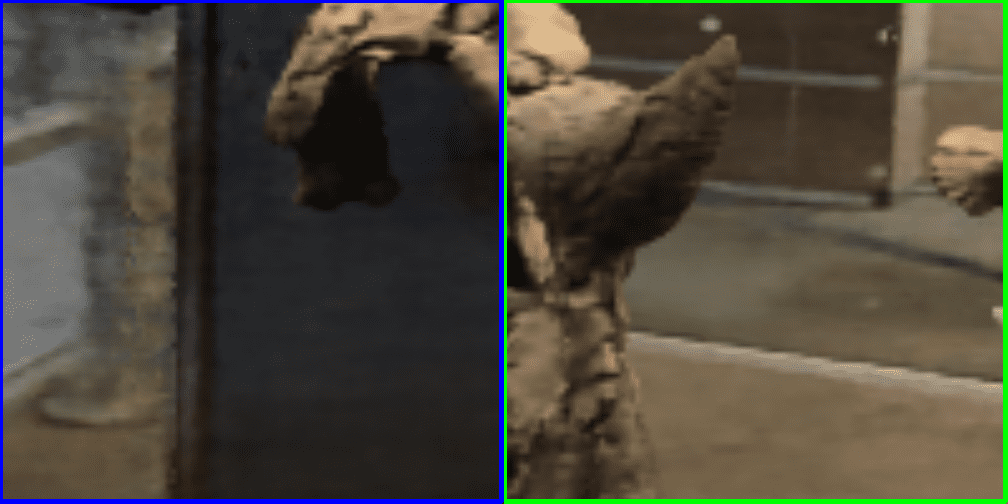}
	\end{minipage}}
	\subfigure[MipNeRF]{
		\begin{minipage}[t]{\scale\linewidth}
			\includegraphics[width=\scaleB\linewidth]{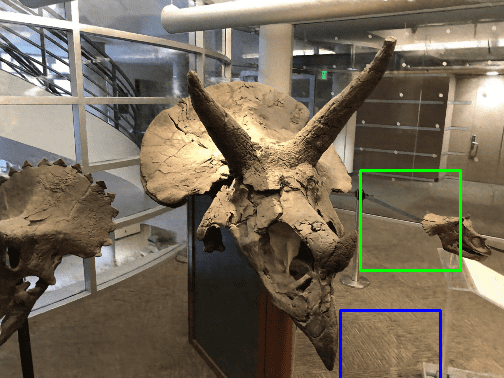}
			\includegraphics[width=\scaleB\linewidth]{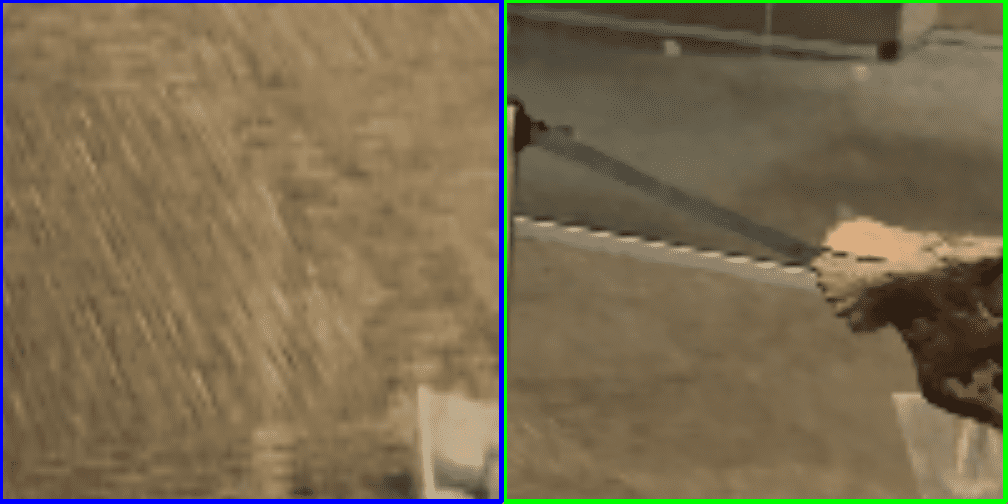}
			\includegraphics[width=\scaleB\linewidth]{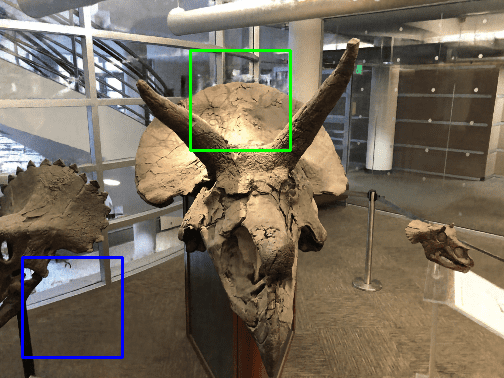}
			\includegraphics[width=\scaleB\linewidth]{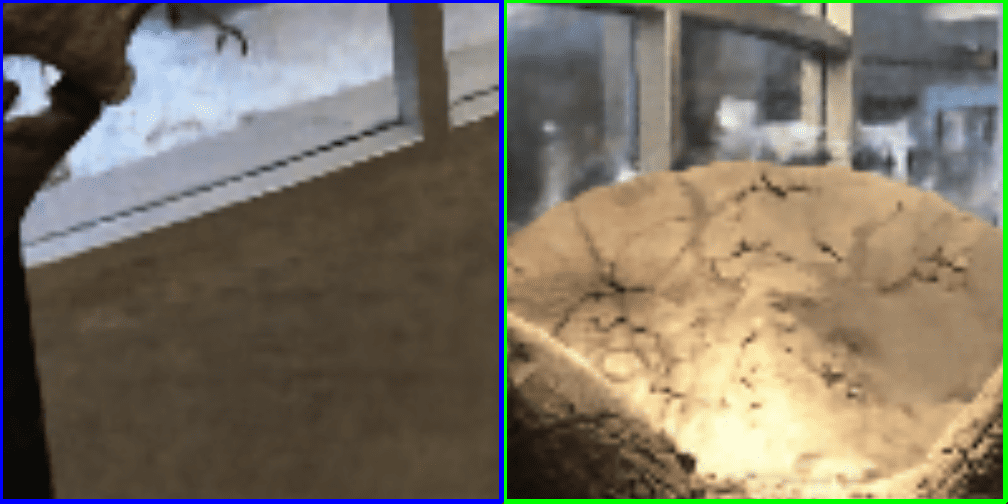}
			\includegraphics[width=\scaleB\linewidth]{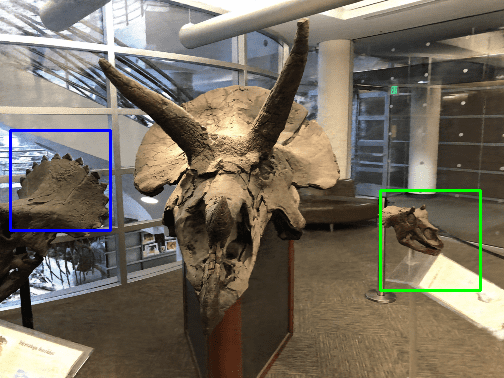}
			\includegraphics[width=\scaleB\linewidth]{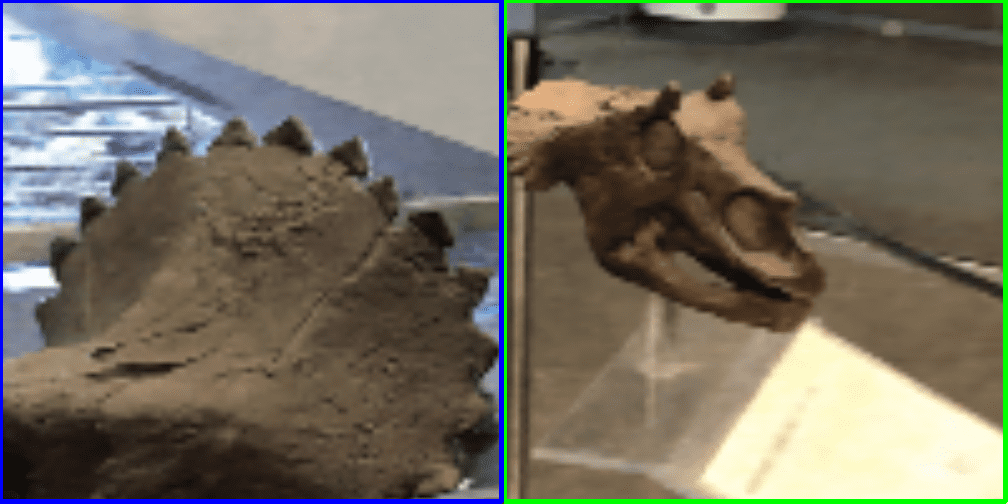}
			\includegraphics[width=\scaleB\linewidth]{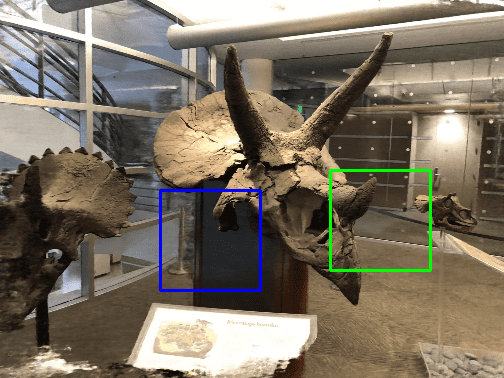}
			\includegraphics[width=\scaleB\linewidth]{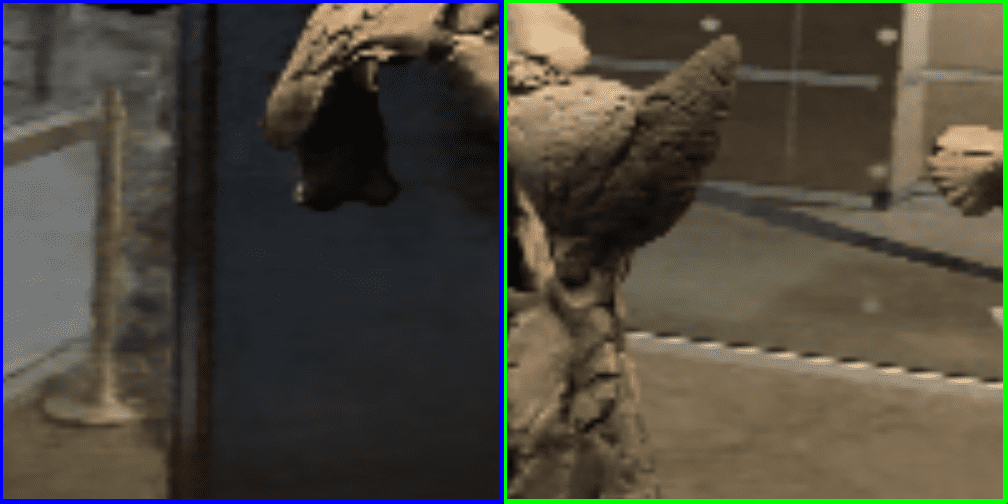}
	\end{minipage}}
	\subfigure[Our]{
		\begin{minipage}[t]{\scale\linewidth}
			\includegraphics[width=\scaleB\linewidth]{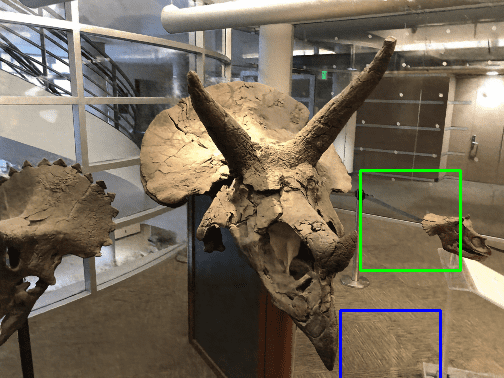}
			\includegraphics[width=\scaleB\linewidth]{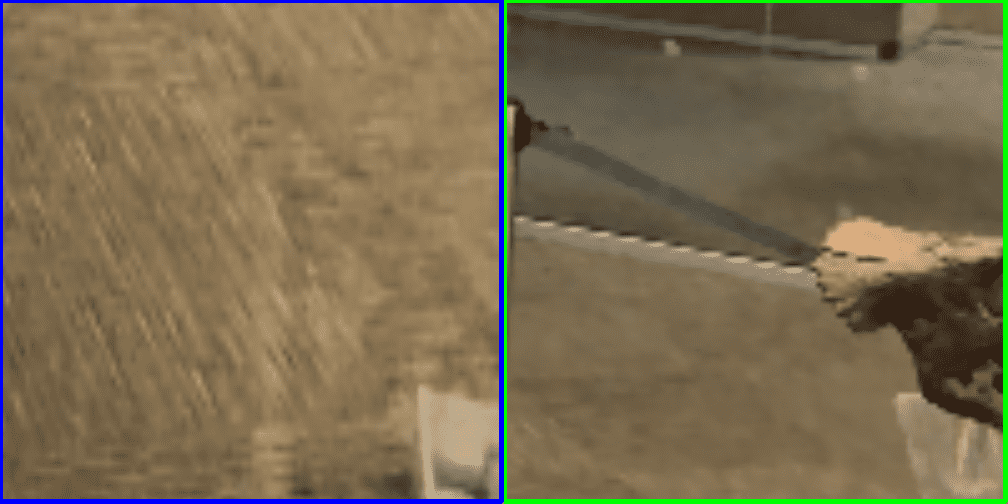}
			\includegraphics[width=\scaleB\linewidth]{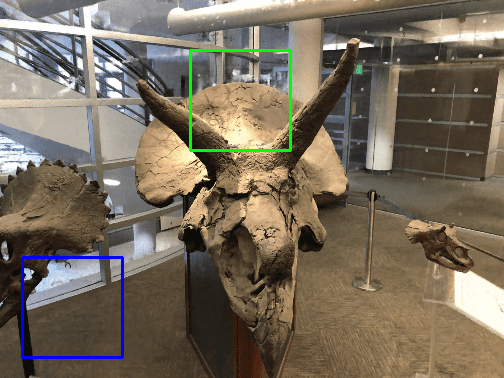}
			\includegraphics[width=\scaleB\linewidth]{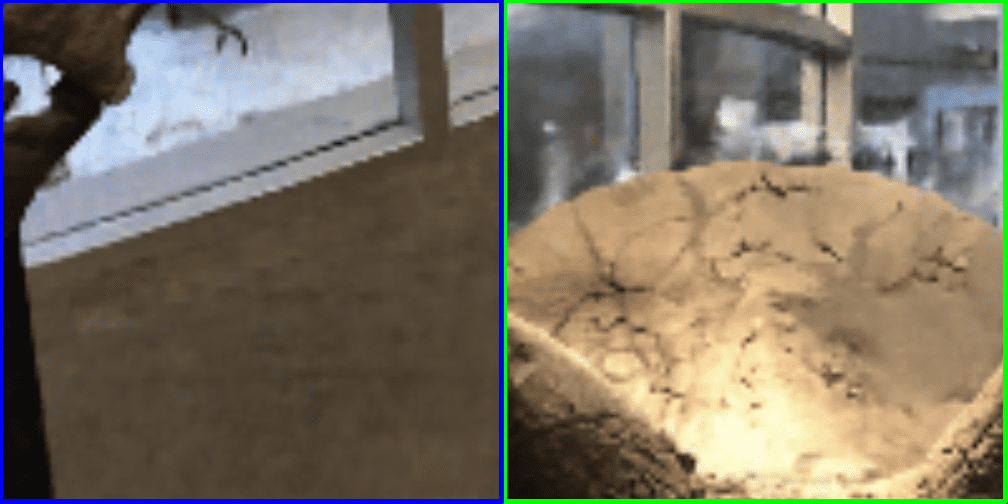}
			\includegraphics[width=\scaleB\linewidth]{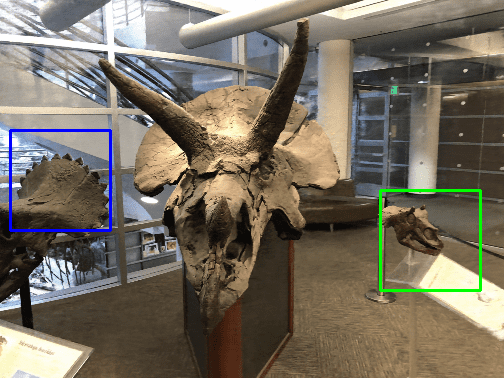}
			\includegraphics[width=\scaleB\linewidth]{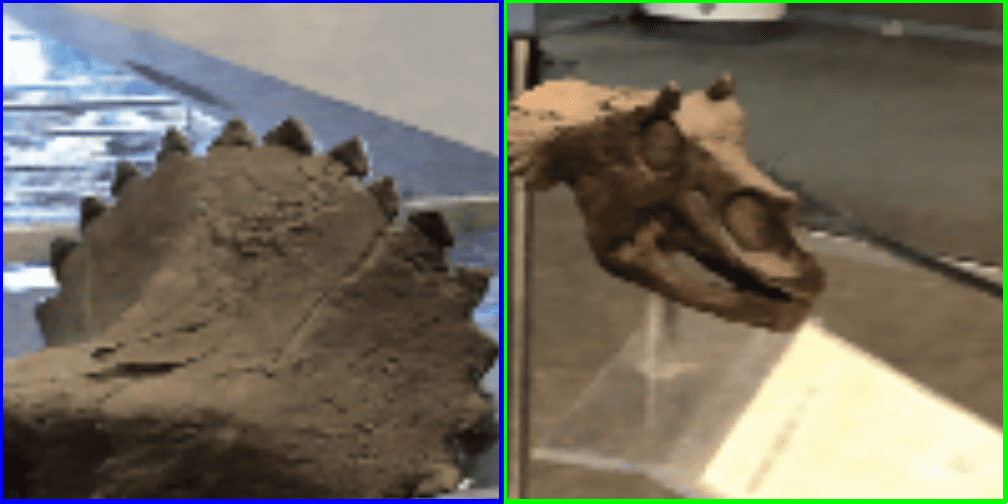}
			\includegraphics[width=\scaleB\linewidth]{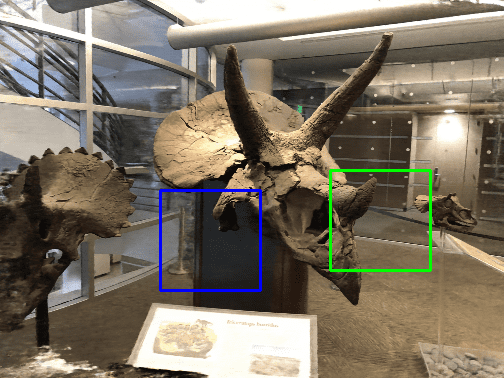}
			\includegraphics[width=\scaleB\linewidth]{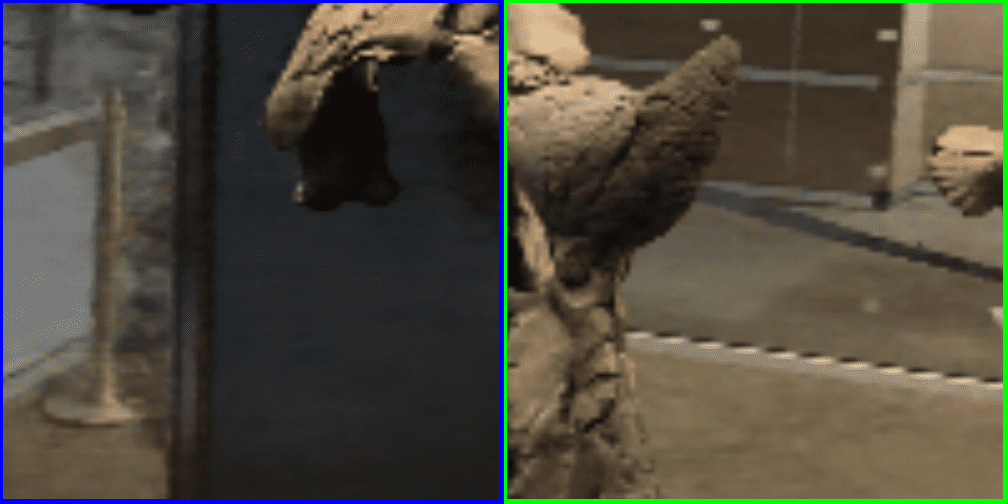}
	\end{minipage}}
	\caption{Extreme novel view synthesis for HORNS dataset with view rotation $R=1.5$. We compare our method against NeRF~\cite{mildenhall2020nerf}, KiloNeRF~\cite{Reiser2021ICCV}, MipNeRF~\cite{barron2021mipnerf}.}
	\label{fig:horns1_5}
\end{figure*}

\begin{figure*}
	\def \scale {0.22}
	\def \scaleB {1.0}
	\centering
	\subfigure[NeRF\cite{mildenhall2020nerf}]{
		\begin{minipage}[t]{\scale\linewidth}
			\includegraphics[width=\scaleB\linewidth]{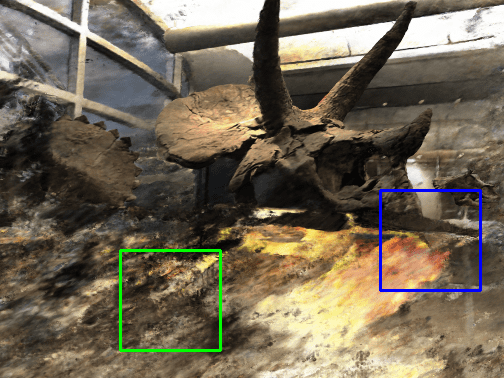}
			\includegraphics[width=\scaleB\linewidth]{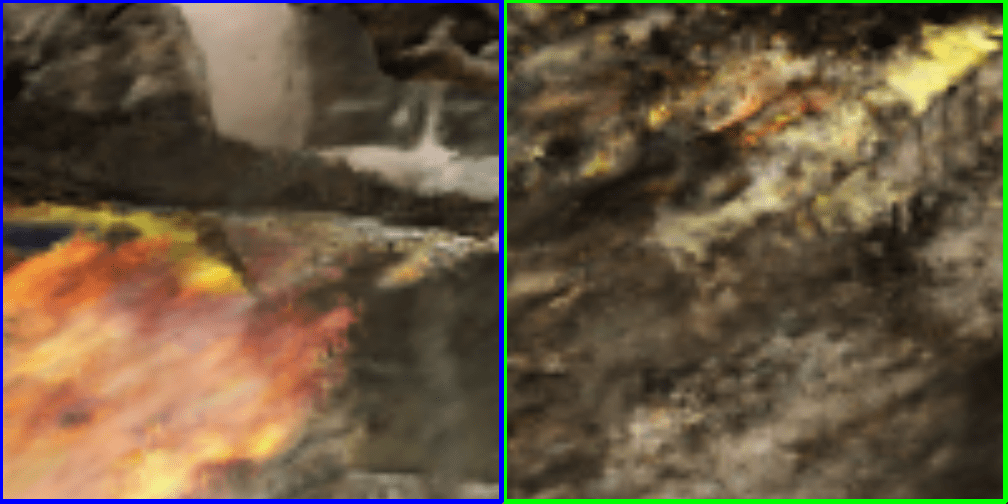}
			\includegraphics[width=\scaleB\linewidth]{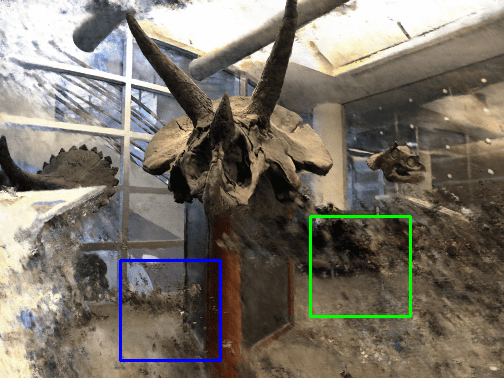}
			\includegraphics[width=\scaleB\linewidth]{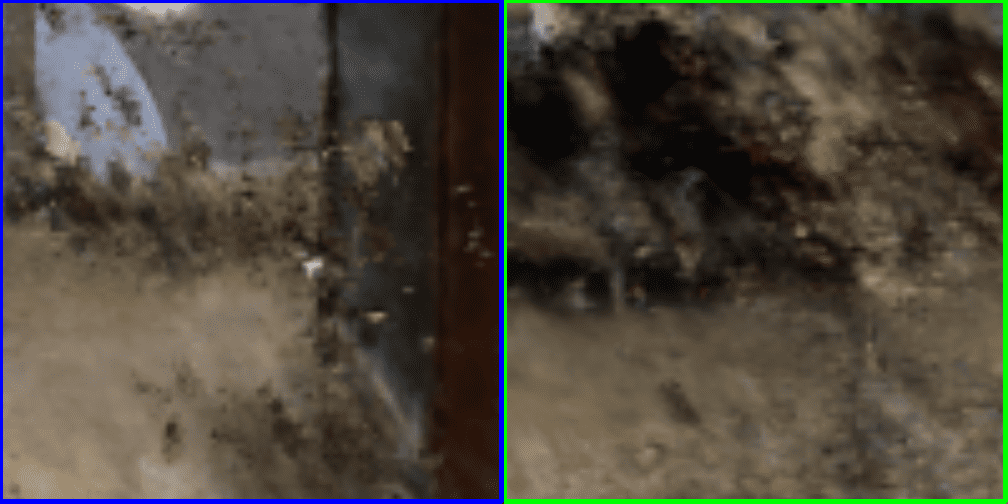}
			\includegraphics[width=\scaleB\linewidth]{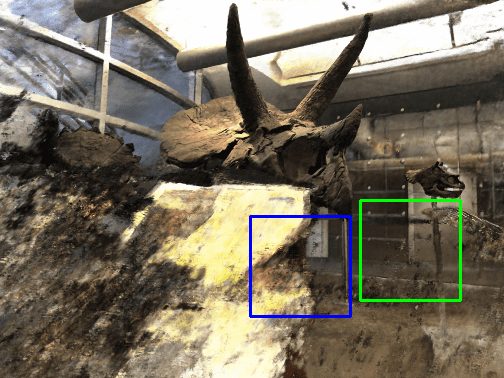}
			\includegraphics[width=\scaleB\linewidth]{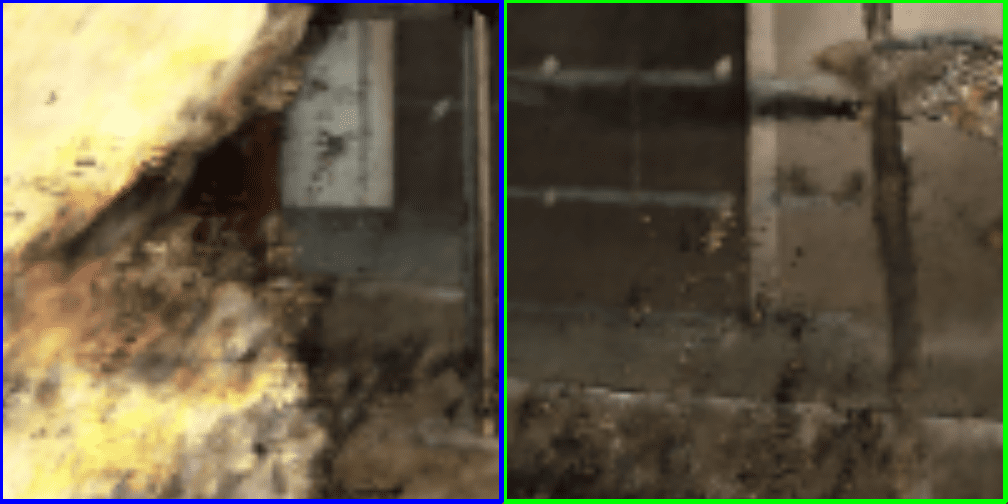}
			\includegraphics[width=\scaleB\linewidth]{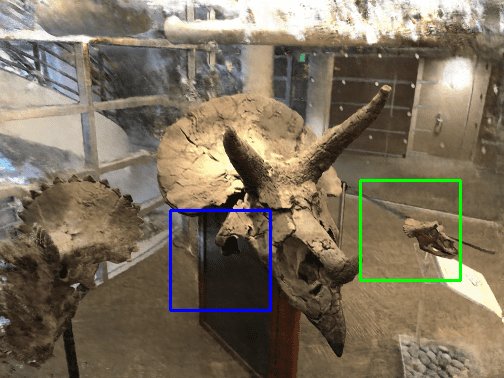}
			\includegraphics[width=\scaleB\linewidth]{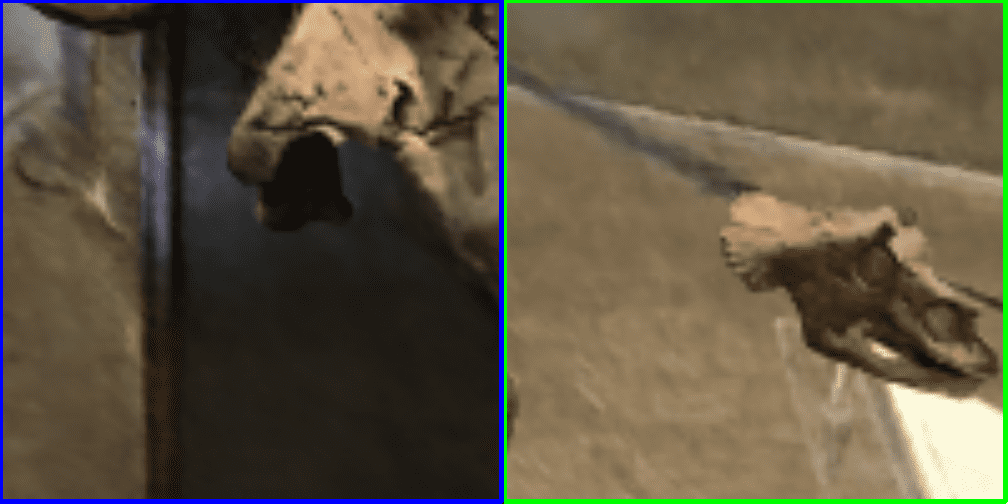}
	\end{minipage}}
	\subfigure[KiloNeRF\cite{Reiser2021ICCV}]{
		\begin{minipage}[t]{\scale\linewidth}
			\includegraphics[width=\scaleB\linewidth]{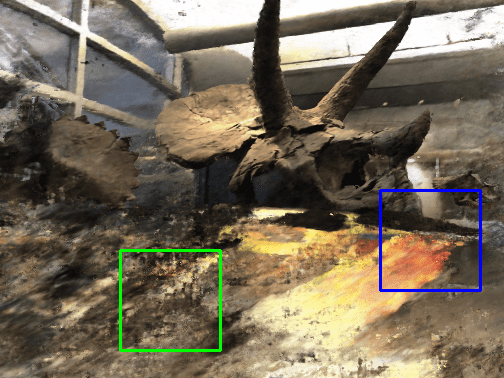}
			\includegraphics[width=\scaleB\linewidth]{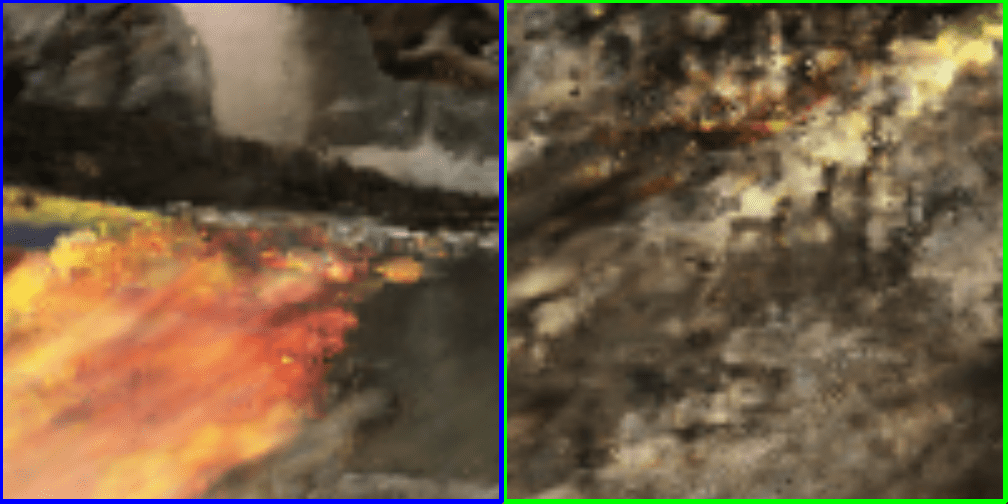}
			\includegraphics[width=\scaleB\linewidth]{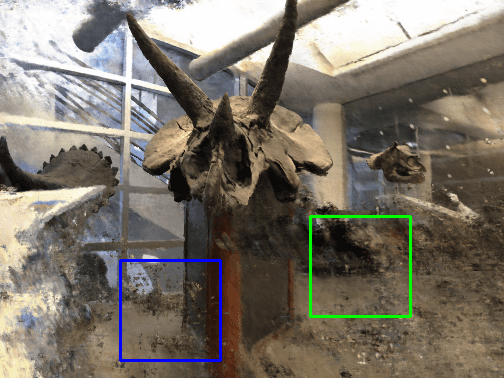}
			\includegraphics[width=\scaleB\linewidth]{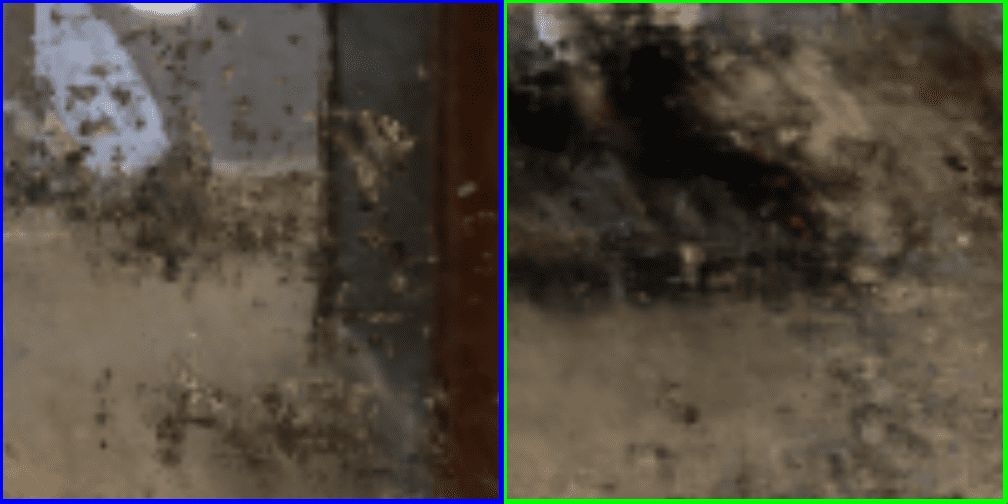}
			\includegraphics[width=\scaleB\linewidth]{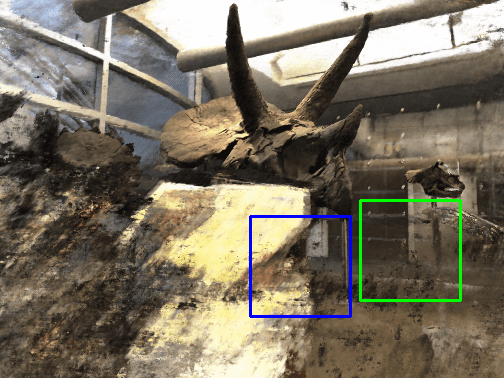}
			\includegraphics[width=\scaleB\linewidth]{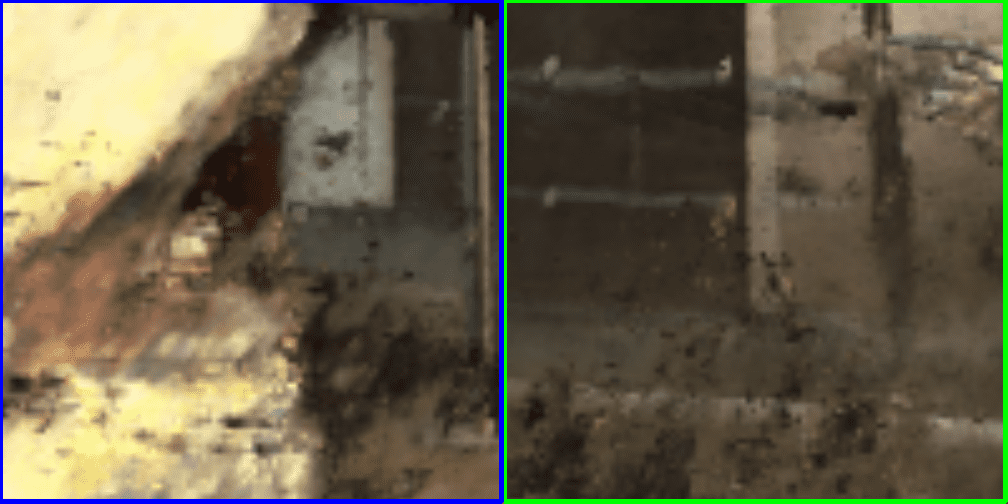}
			\includegraphics[width=\scaleB\linewidth]{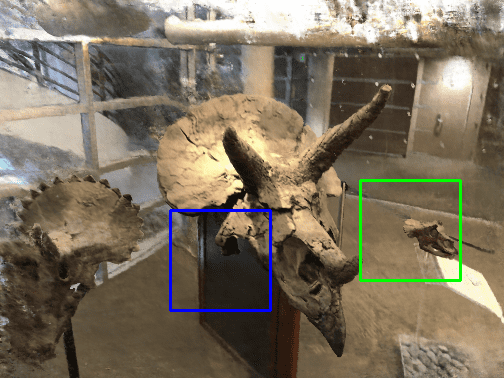}
			\includegraphics[width=\scaleB\linewidth]{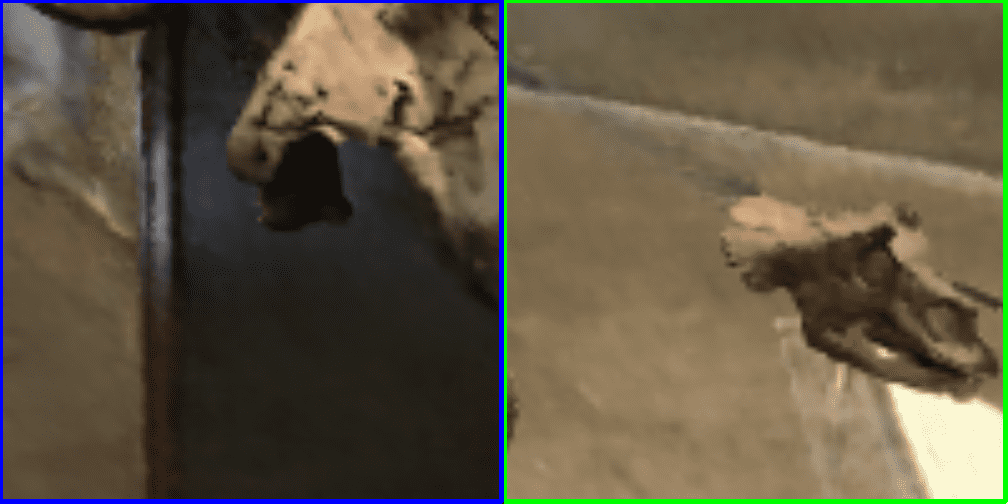}
	\end{minipage}}
	\subfigure[MipNeRF\cite{barron2021mipnerf}]{
		\begin{minipage}[t]{\scale\linewidth}
			\includegraphics[width=\scaleB\linewidth]{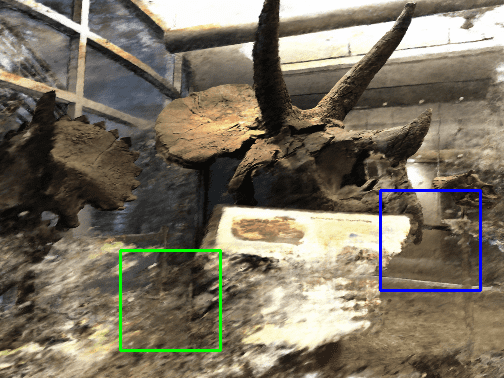}
			\includegraphics[width=\scaleB\linewidth]{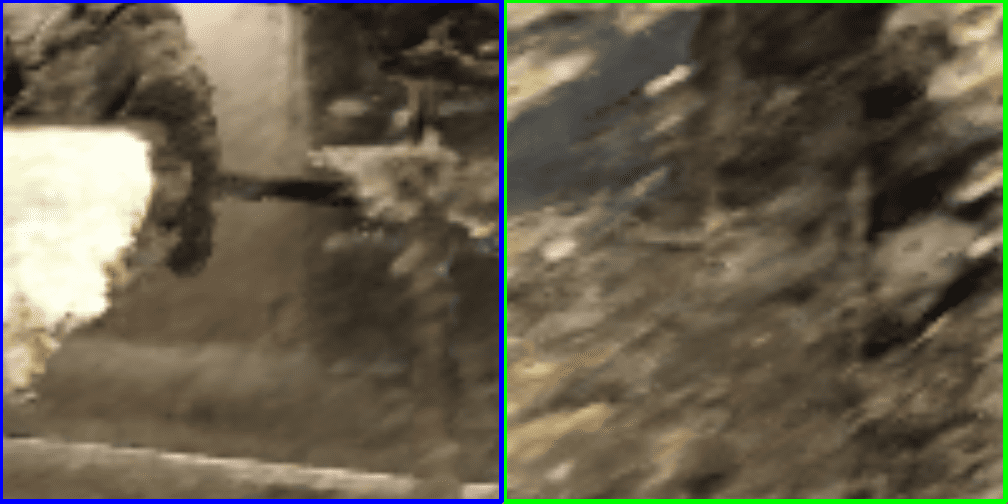}
			\includegraphics[width=\scaleB\linewidth]{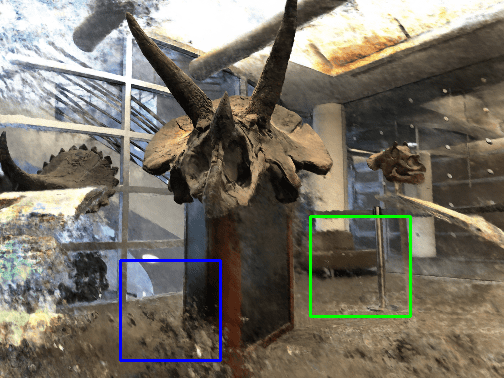}
			\includegraphics[width=\scaleB\linewidth]{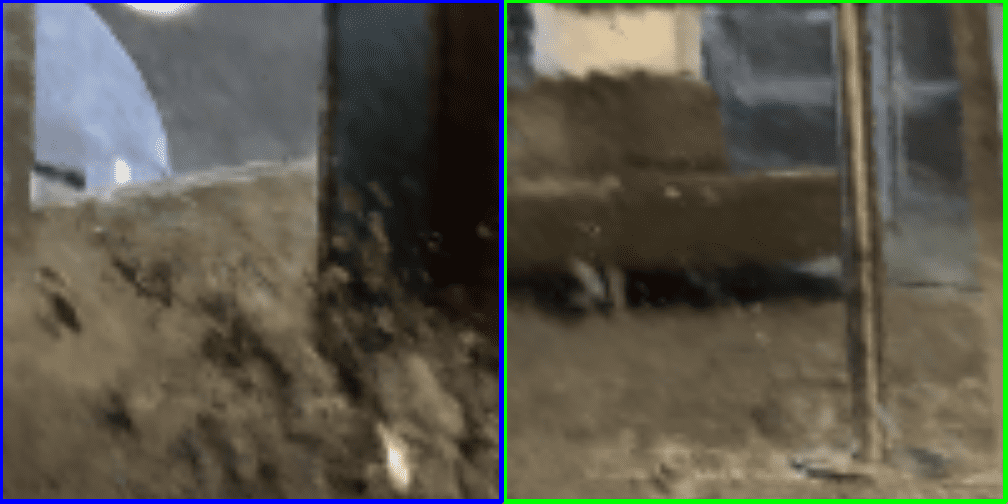}
			\includegraphics[width=\scaleB\linewidth]{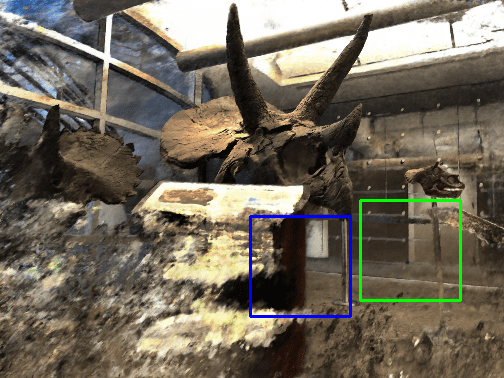}
			\includegraphics[width=\scaleB\linewidth]{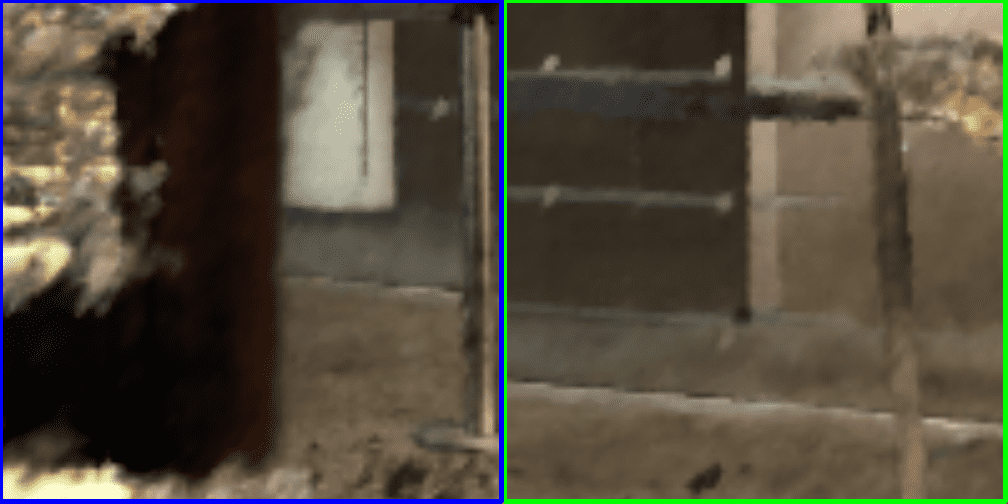}
			\includegraphics[width=\scaleB\linewidth]{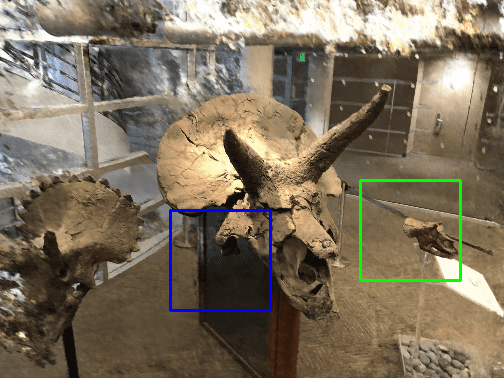}
			\includegraphics[width=\scaleB\linewidth]{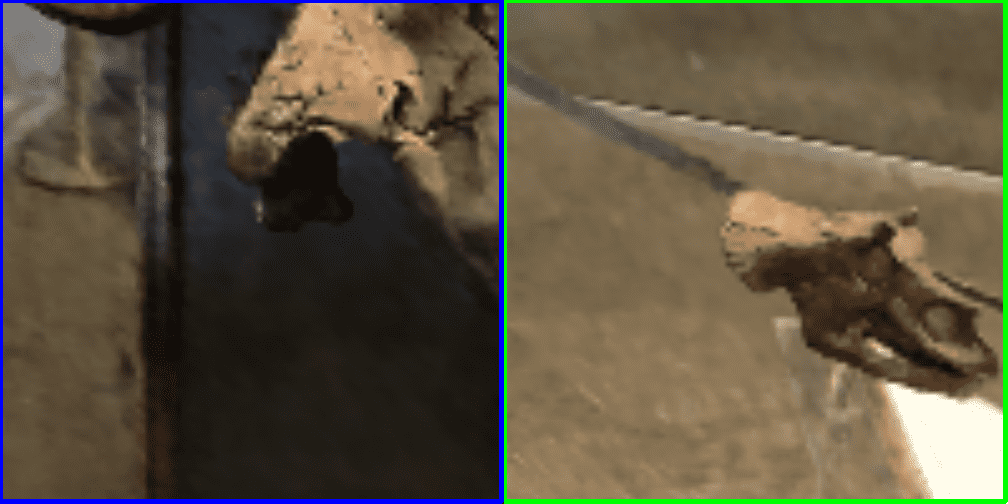}
	\end{minipage}}
	\subfigure[Our]{
		\begin{minipage}[t]{\scale\linewidth}
			\includegraphics[width=\scaleB\linewidth]{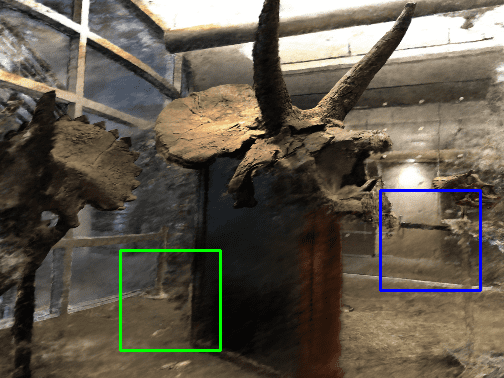}
			\includegraphics[width=\scaleB\linewidth]{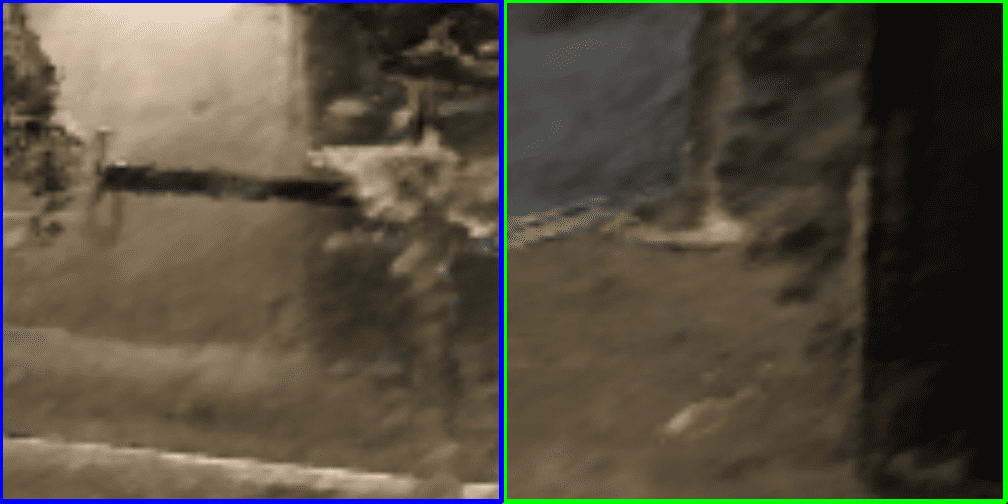}
			\includegraphics[width=\scaleB\linewidth]{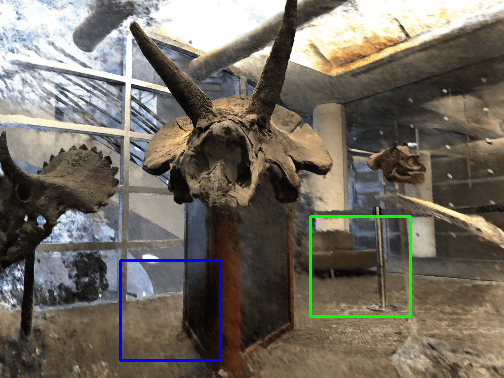}
			\includegraphics[width=\scaleB\linewidth]{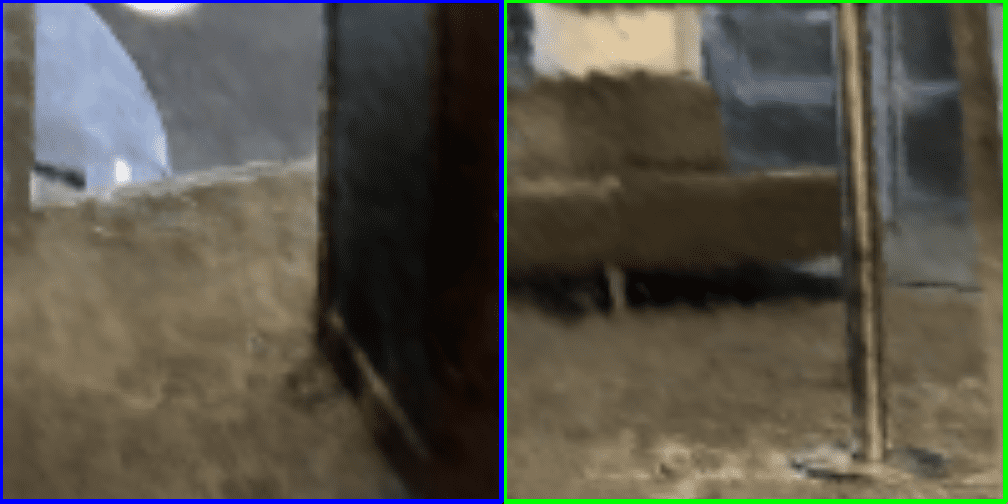}
			\includegraphics[width=\scaleB\linewidth]{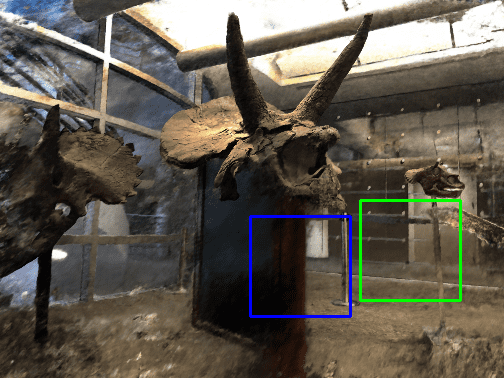}
			\includegraphics[width=\scaleB\linewidth]{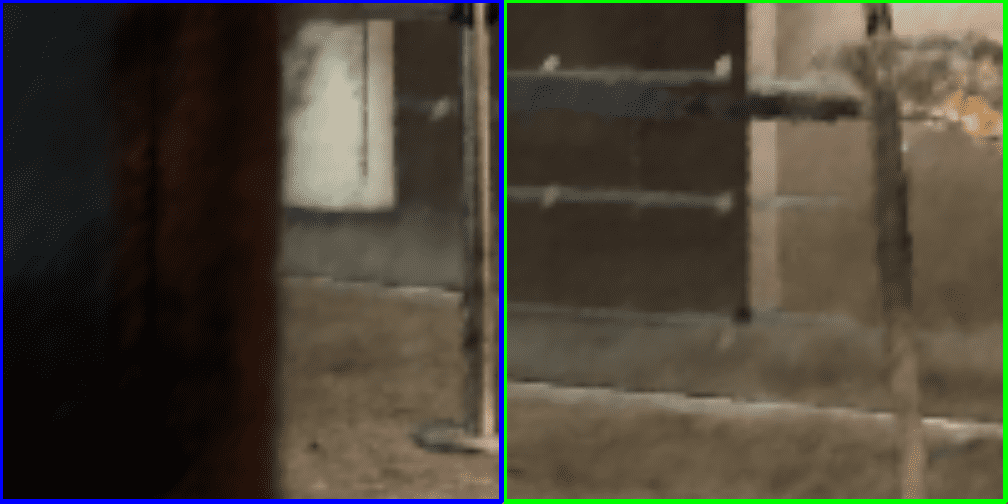}
			\includegraphics[width=\scaleB\linewidth]{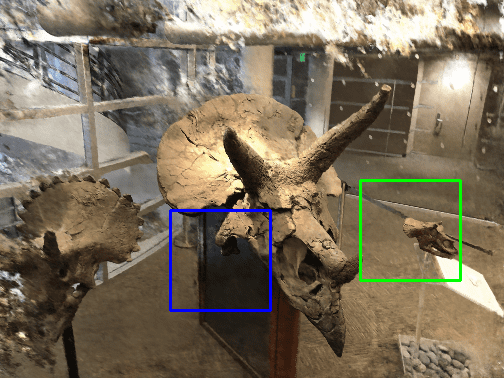}
			\includegraphics[width=\scaleB\linewidth]{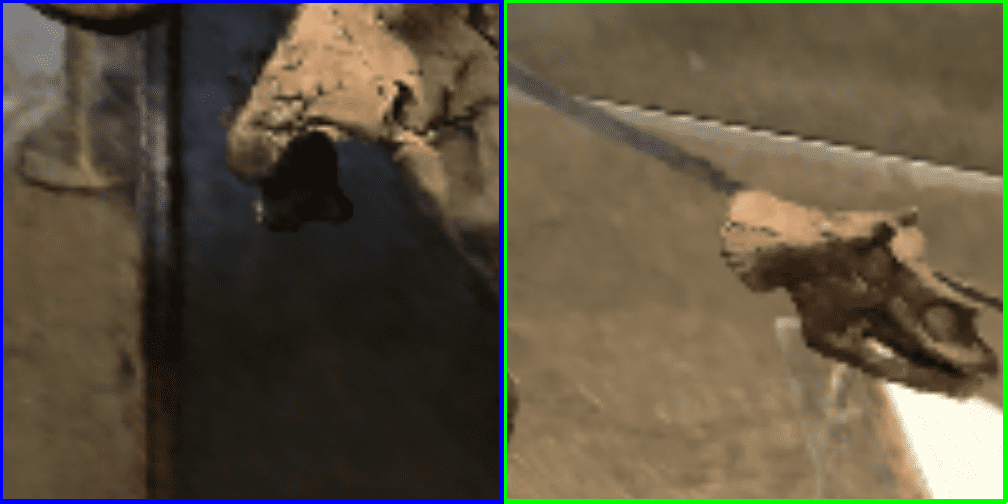}
	\end{minipage}}
	\caption{Extreme novel view synthesis for HORNS dataset with view rotation $R=3.0$. We compare our method against NeRF~\cite{mildenhall2020nerf}, KiloNeRF~\cite{Reiser2021ICCV}, MipNeRF~\cite{barron2021mipnerf}.}
	\label{fig:horns3}
\end{figure*}

\begin{figure*}
	\def \scale {0.22}
	\def \scaleB {1.0}
	\centering
	\subfigure[NeRF\cite{mildenhall2020nerf}]{
		\begin{minipage}[t]{\scale\linewidth}
			\includegraphics[width=\scaleB\linewidth]{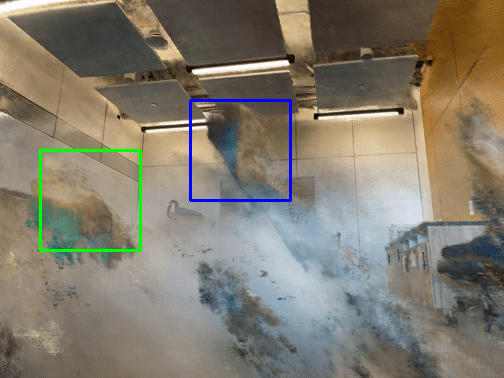}
			\includegraphics[width=\scaleB\linewidth]{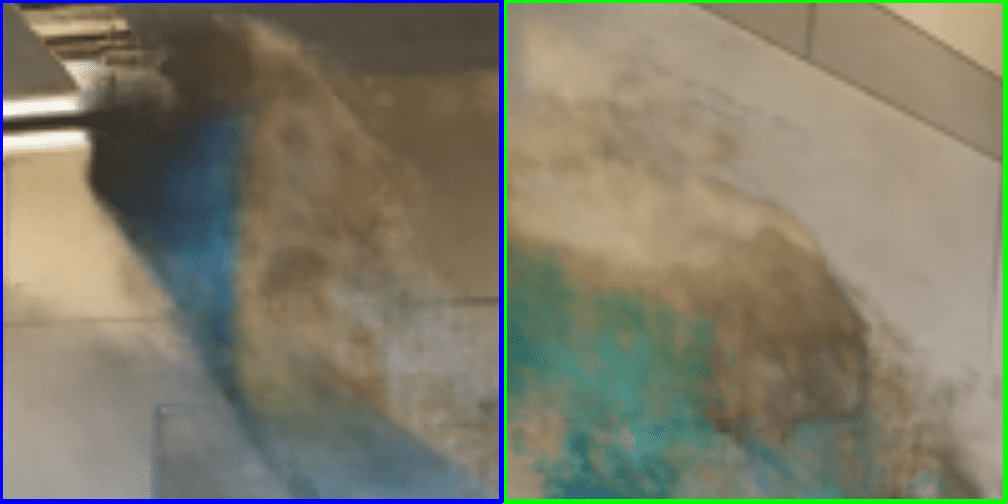}
			\includegraphics[width=\scaleB\linewidth]{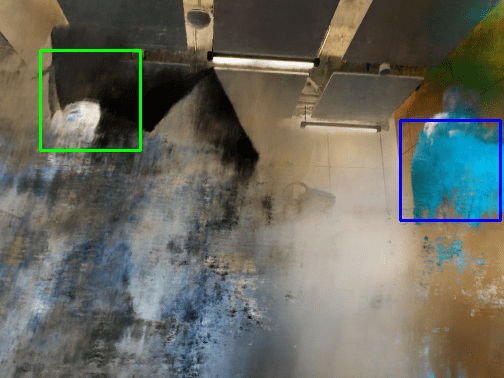}
			\includegraphics[width=\scaleB\linewidth]{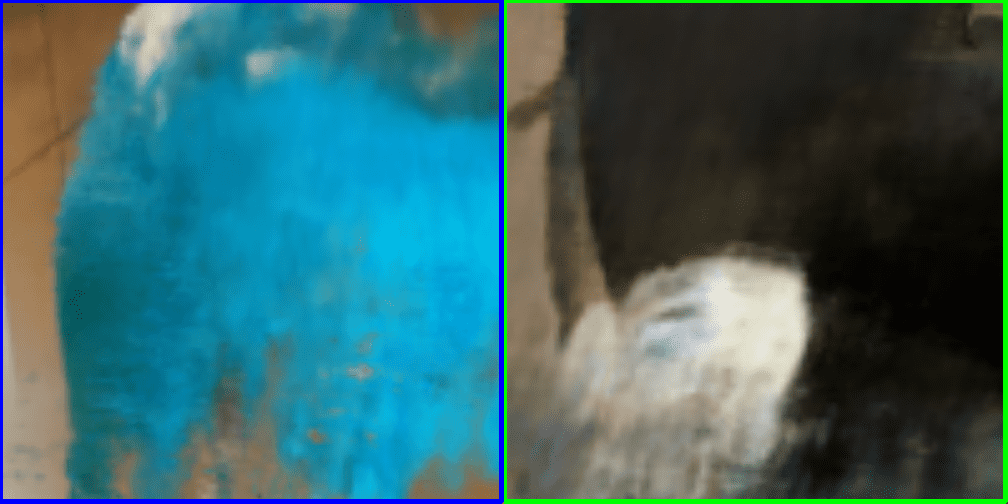}
			\includegraphics[width=\scaleB\linewidth]{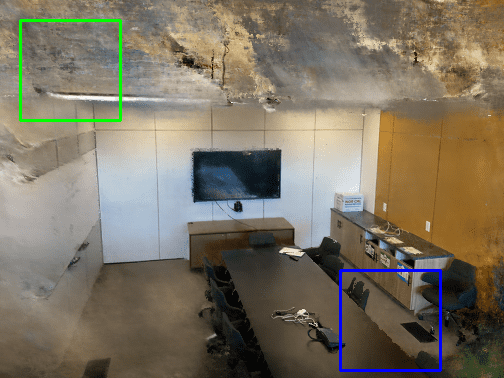}
			\includegraphics[width=\scaleB\linewidth]{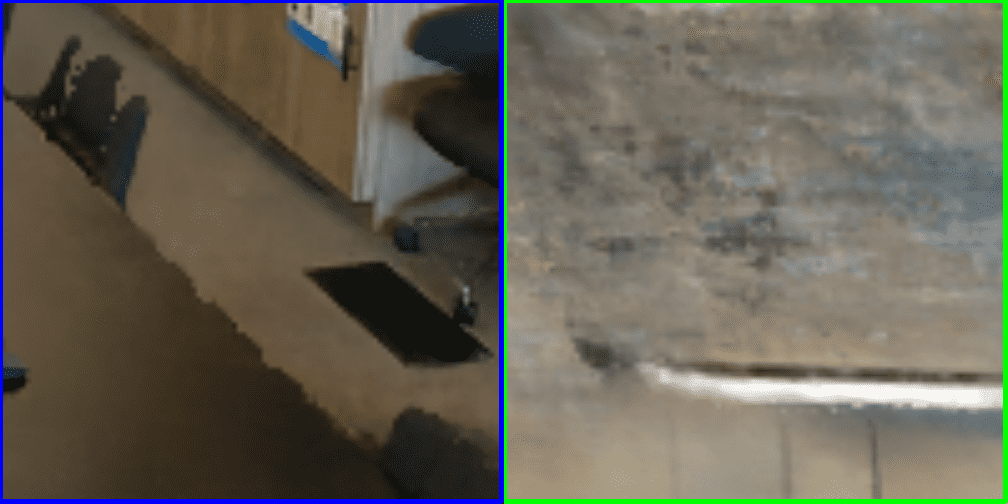}
			\includegraphics[width=\scaleB\linewidth]{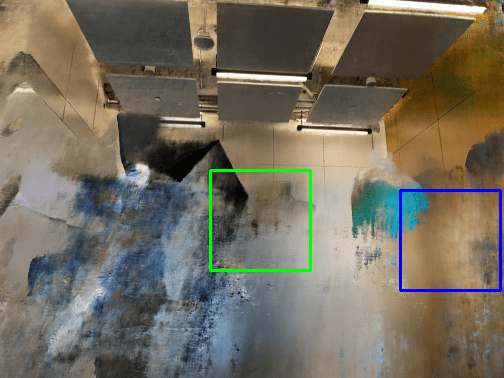}
			\includegraphics[width=\scaleB\linewidth]{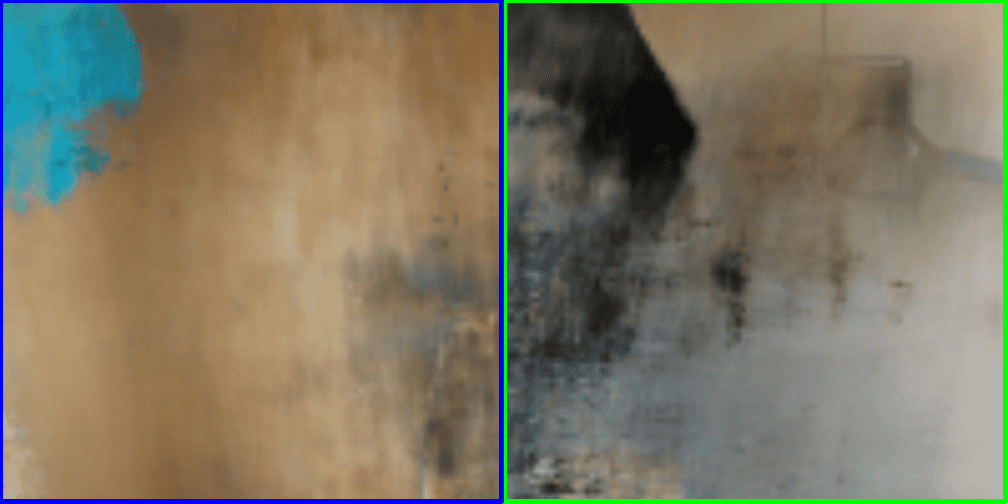}
	\end{minipage}}
	\subfigure[KiloNeRF\cite{Reiser2021ICCV}]{
		\begin{minipage}[t]{\scale\linewidth}
			\includegraphics[width=\scaleB\linewidth]{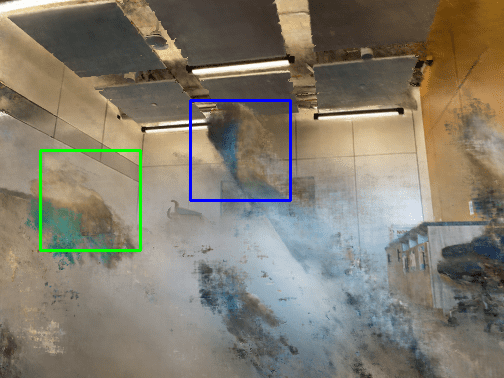}
			\includegraphics[width=\scaleB\linewidth]{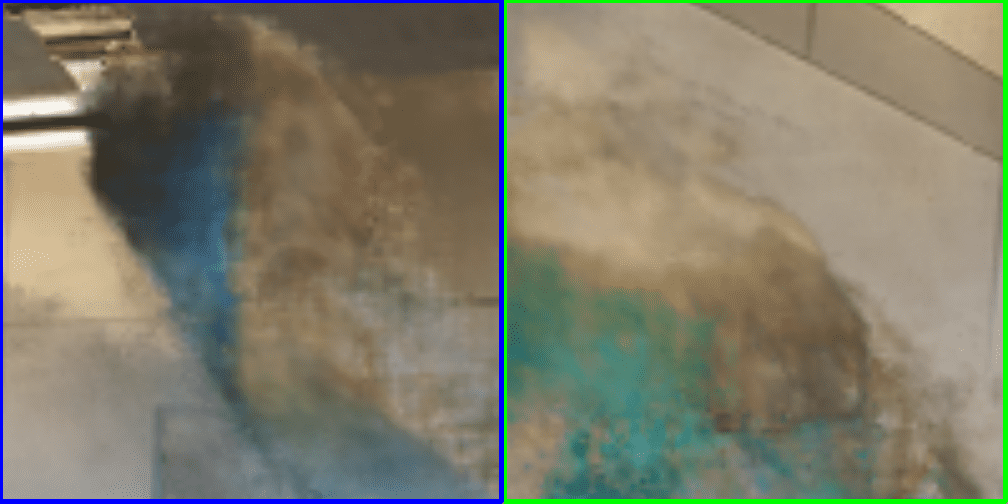}
			\includegraphics[width=\scaleB\linewidth]{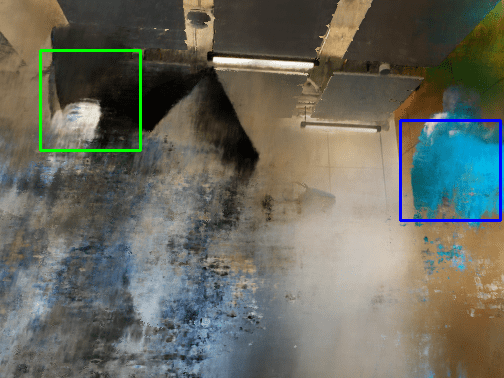}
			\includegraphics[width=\scaleB\linewidth]{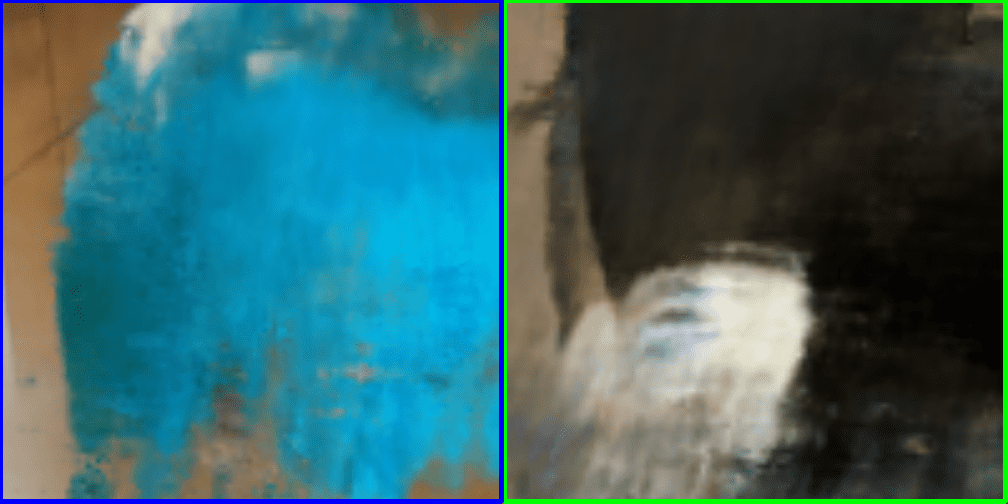}
			\includegraphics[width=\scaleB\linewidth]{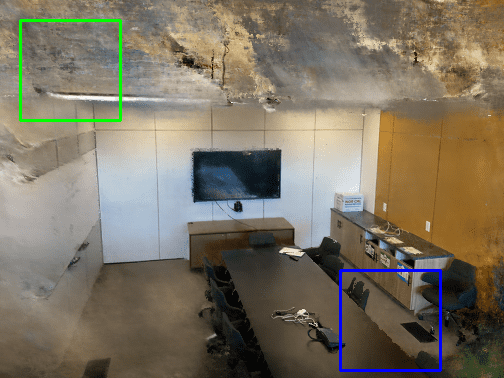}
			\includegraphics[width=\scaleB\linewidth]{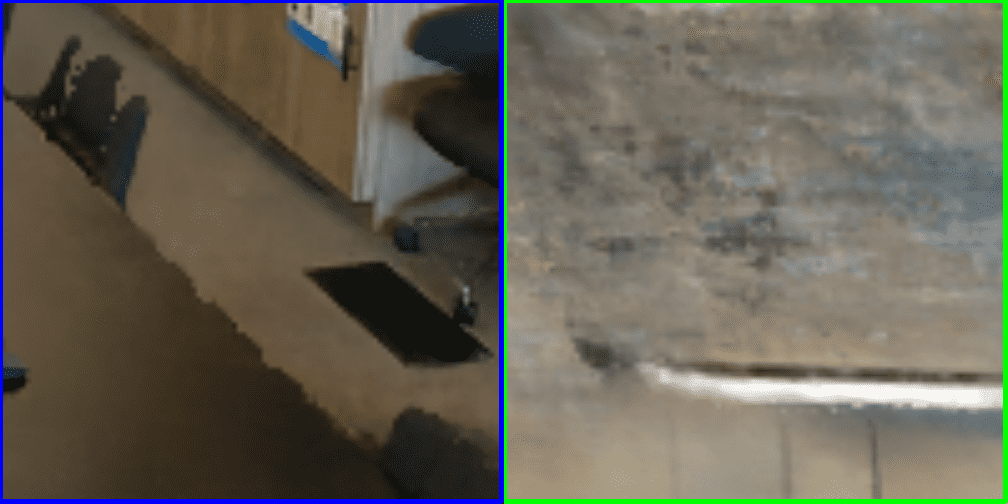}
			\includegraphics[width=\scaleB\linewidth]{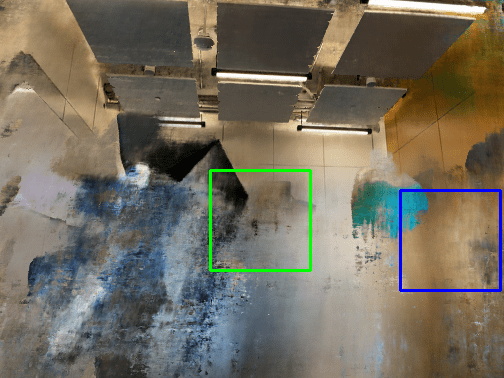}
			\includegraphics[width=\scaleB\linewidth]{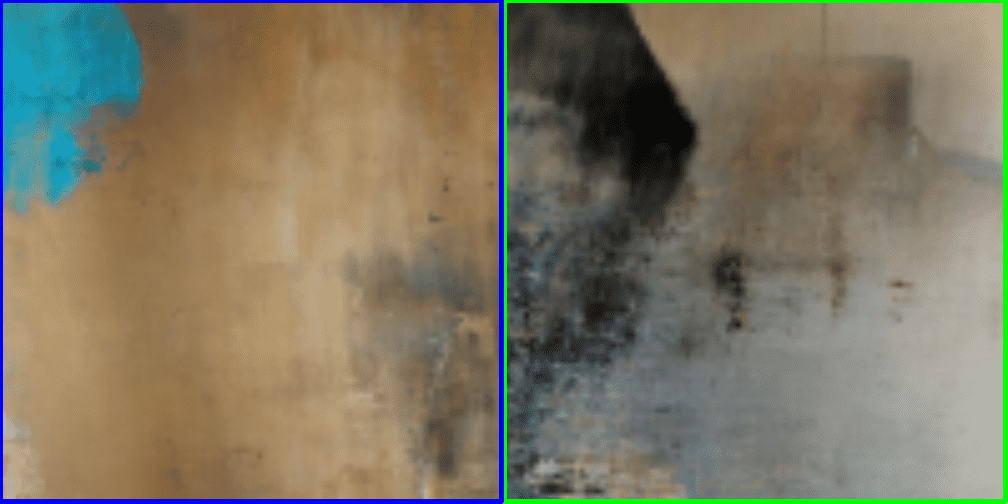}
	\end{minipage}}
	\subfigure[MipNeRF\cite{barron2021mipnerf}]{
		\begin{minipage}[t]{\scale\linewidth}
			\includegraphics[width=\scaleB\linewidth]{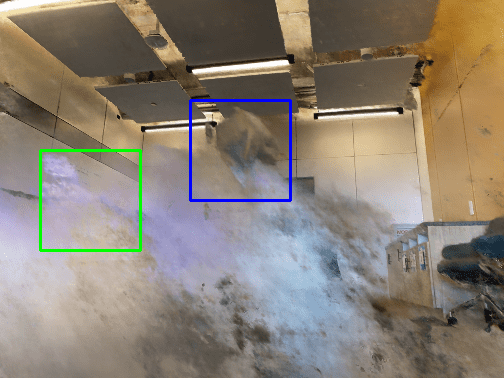}
			\includegraphics[width=\scaleB\linewidth]{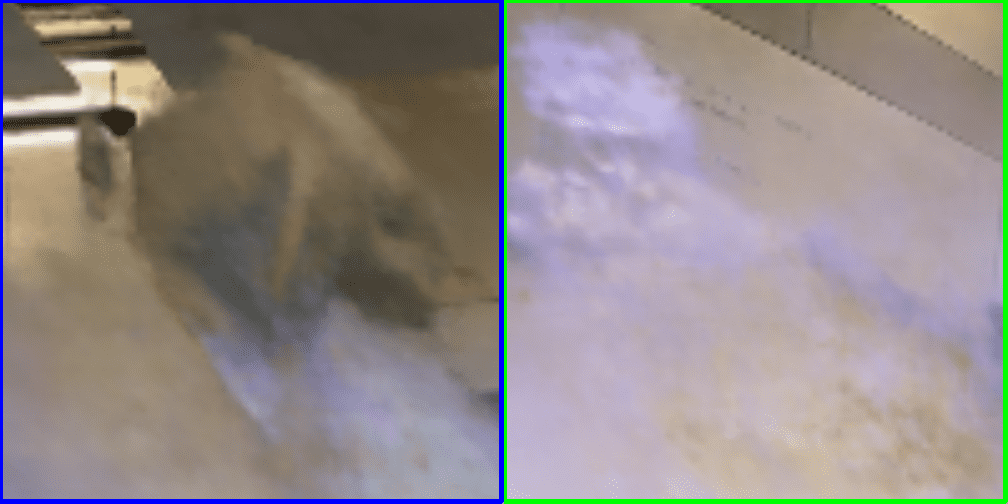}
			\includegraphics[width=\scaleB\linewidth]{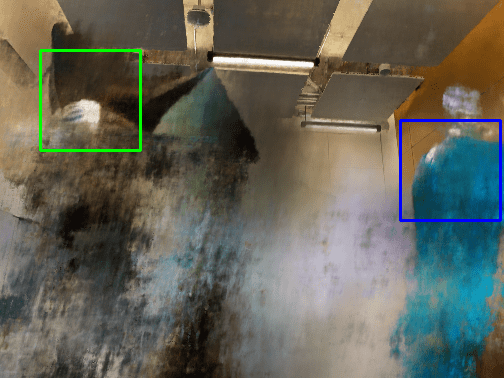}
			\includegraphics[width=\scaleB\linewidth]{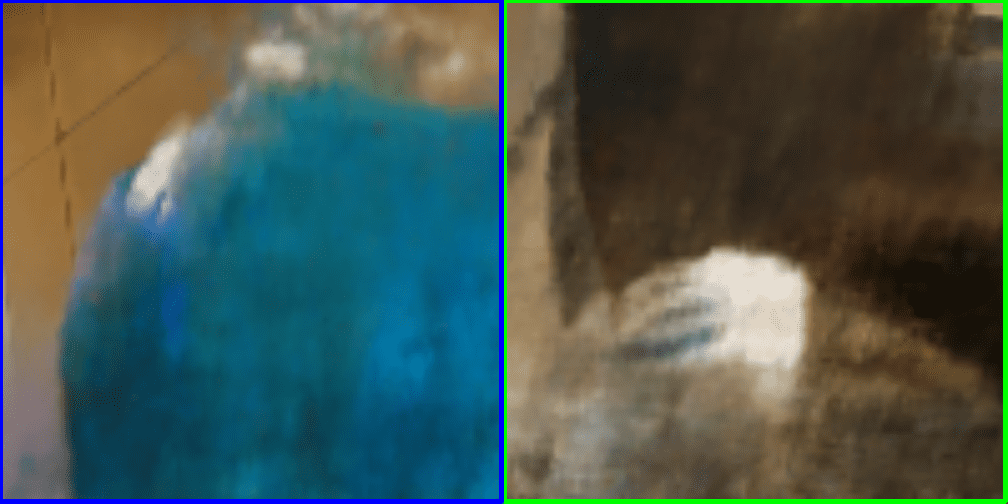}
			\includegraphics[width=\scaleB\linewidth]{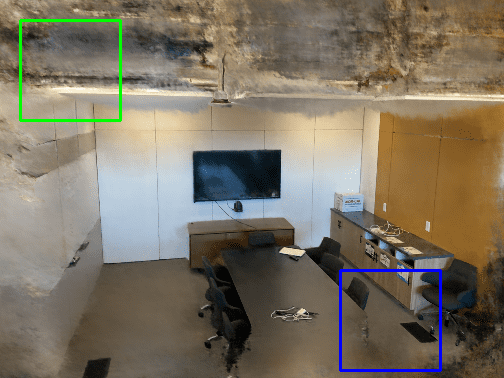}
			\includegraphics[width=\scaleB\linewidth]{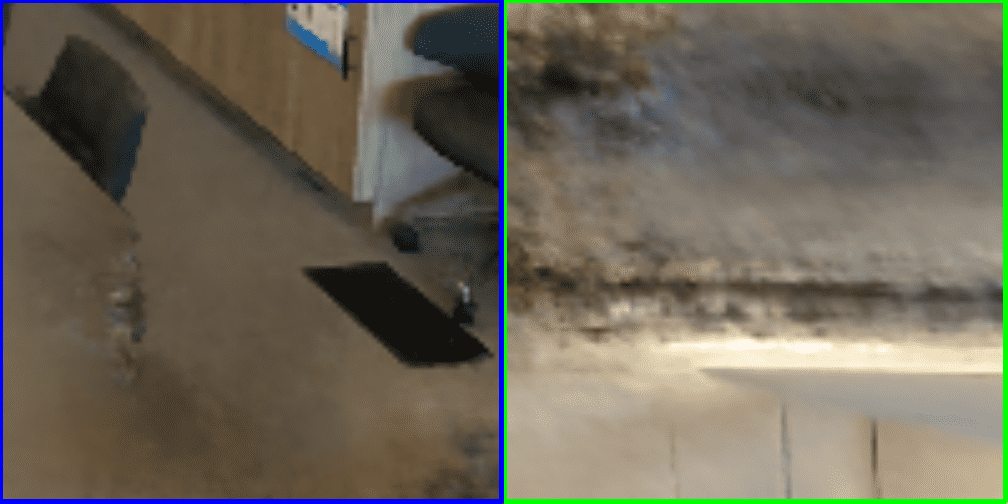}
			\includegraphics[width=\scaleB\linewidth]{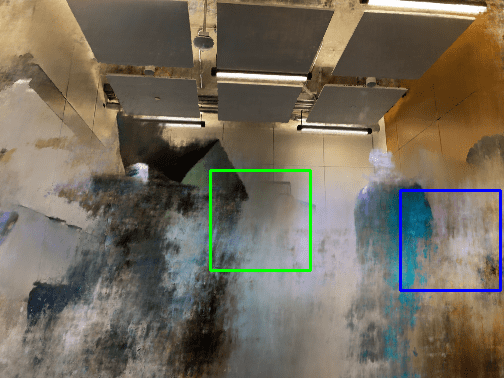}
			\includegraphics[width=\scaleB\linewidth]{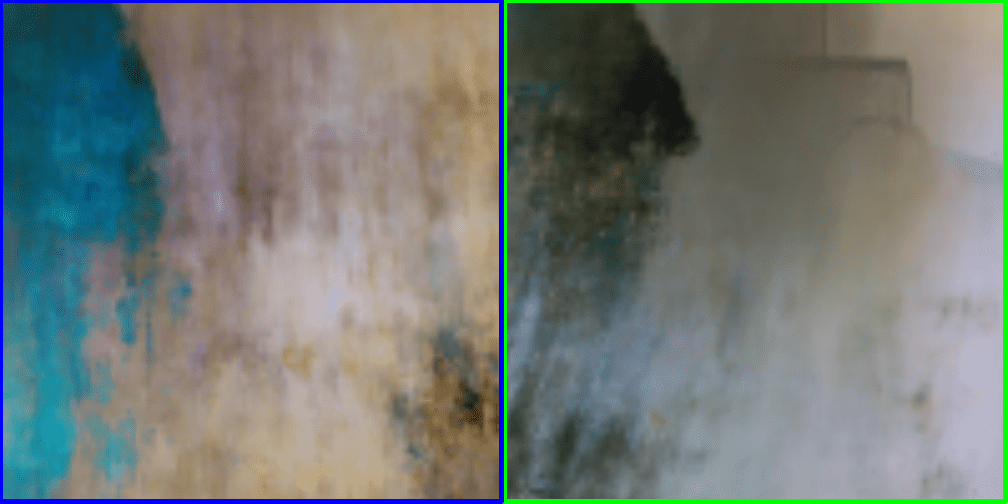}
	\end{minipage}}
	\subfigure[Our]{
		\begin{minipage}[t]{\scale\linewidth}
			\includegraphics[width=\scaleB\linewidth]{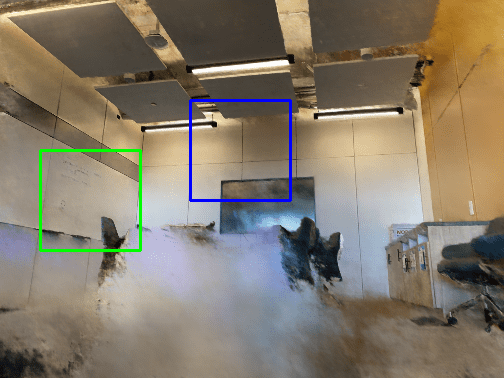}
			\includegraphics[width=\scaleB\linewidth]{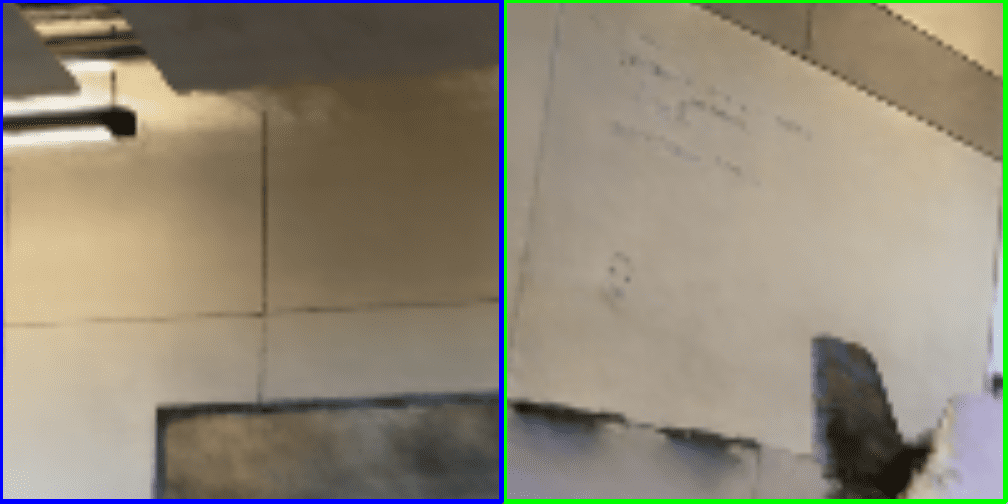}
			\includegraphics[width=\scaleB\linewidth]{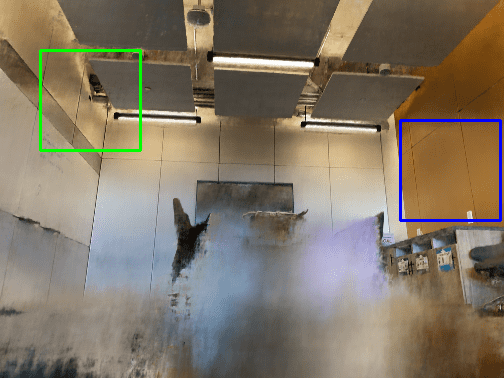}
			\includegraphics[width=\scaleB\linewidth]{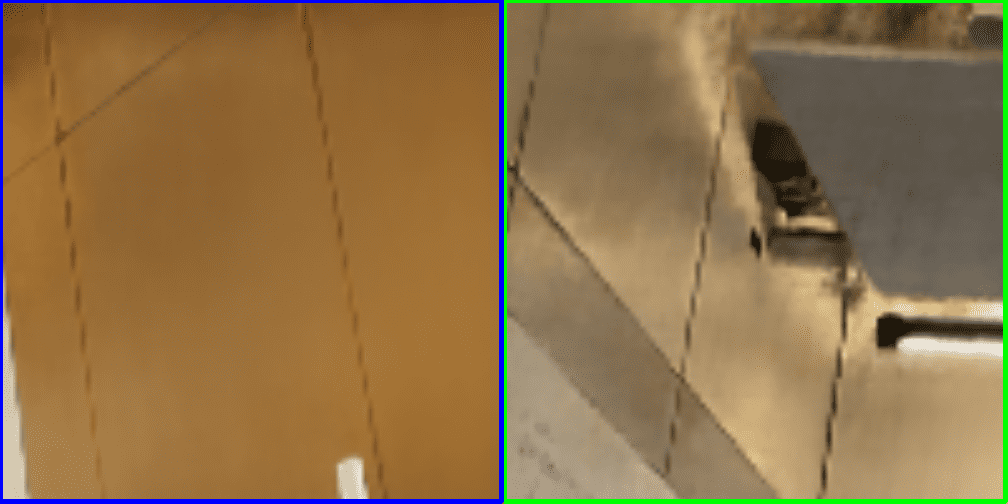}
			\includegraphics[width=\scaleB\linewidth]{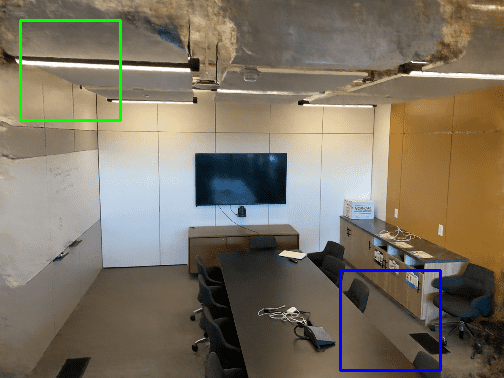}
			\includegraphics[width=\scaleB\linewidth]{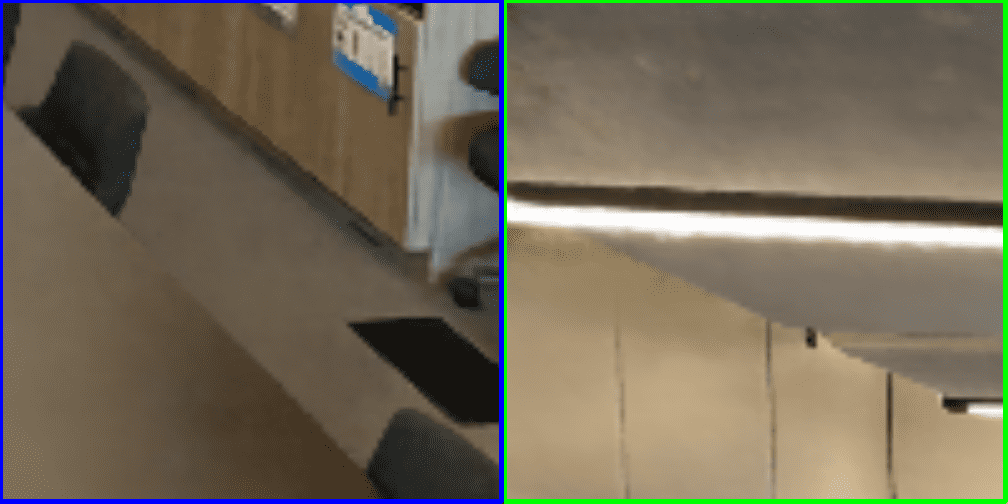}
			\includegraphics[width=\scaleB\linewidth]{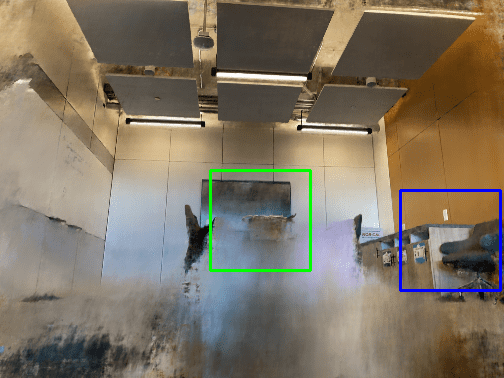}
			\includegraphics[width=\scaleB\linewidth]{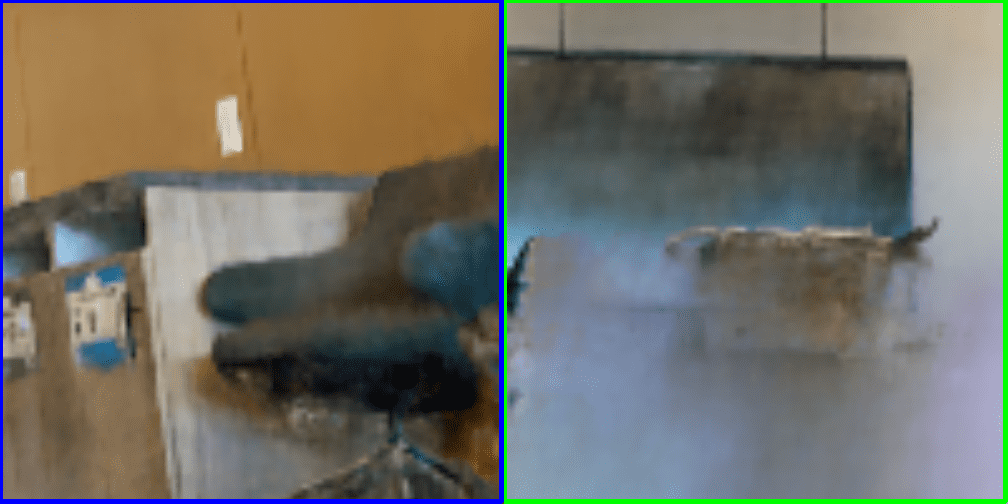}
	\end{minipage}}
	\caption{Extreme novel view synthesis for ROOMS dataset with view rotation $R=3.0$. We compare our method against NeRF~\cite{mildenhall2020nerf}, KiloNeRF~\cite{Reiser2021ICCV}, MipNeRF~\cite{barron2021mipnerf}.}
	\label{fig:room3}
\end{figure*}

{
In this section, we show the extensive comparison in details for extreme view of real scene dataset. \figref{fig:horns1_5} shows the novel view synthesis with view rotation radius $R=1.5$, our method shows similar performance with MipNeRF~\cite{barron2021mipnerf}, but outperforms NeRF and KiloNeRF in details texture recovering. \figref{fig:horns3} increase rotation radius for extreme view rendering, NeRF, KiloNeRF and MipNeRF show significant false trails in extrapolated view due to neural network tends to output unknown density value in extrapolated part of scene. Octtree could explicitly define rendering space, and significantly allevate rendering trails. Alought KiloNeRF can also use pre-trained octtree to accelerate rendering process, it still require train a extra single network for model distilling, thus reserve same trails effect in the fine-tune stage. \figref{fig:room3} also shows more comparison for extrapolated novel view synthesis, our method outperform other alternatives, see details for better visual comparison.
}

\subsection{UAV Scene Reconstruction}
\begin{figure*}
	\def \scale {0.31}
	\def \scaleB {1.0}
	\centering
	\subfigure[NeRF\cite{mildenhall2020nerf}]{
		\begin{minipage}[t]{\scale\linewidth}
			\includegraphics[width=\scaleB\linewidth]{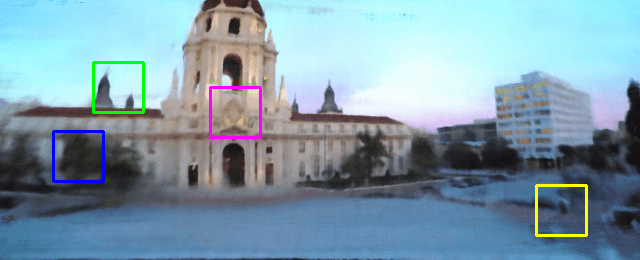}
			\includegraphics[width=\scaleB\linewidth]{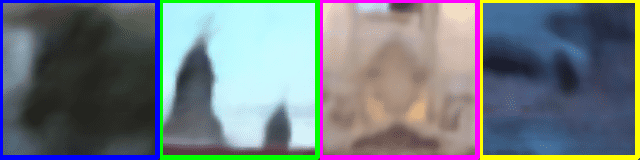}
			\includegraphics[width=\scaleB\linewidth]{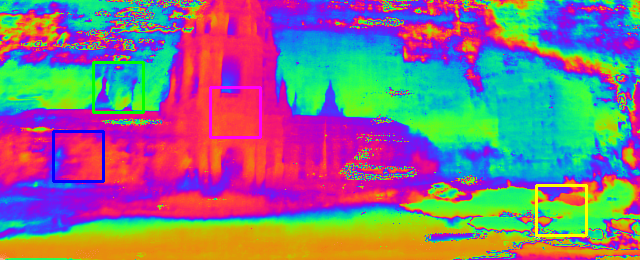}
			\includegraphics[width=\scaleB\linewidth]{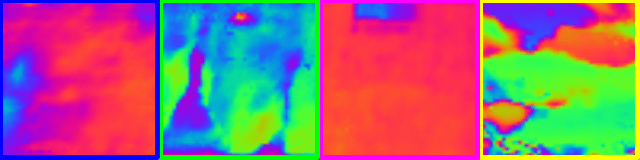}
			\includegraphics[width=\scaleB\linewidth]{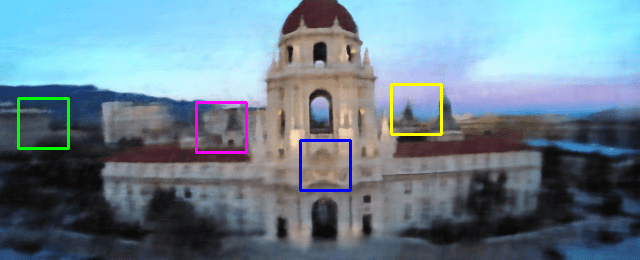}
			\includegraphics[width=\scaleB\linewidth]{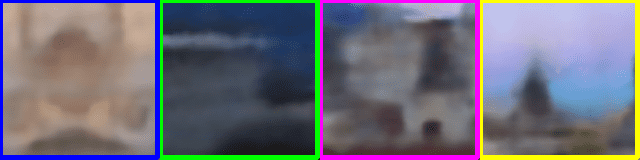}
			\includegraphics[width=\scaleB\linewidth]{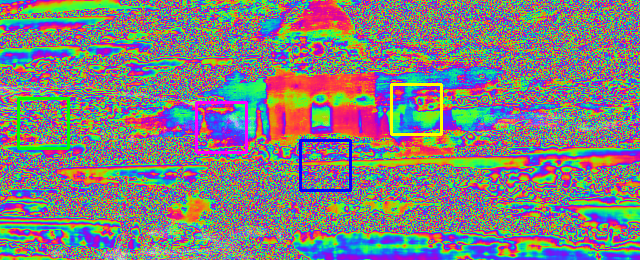}
			\includegraphics[width=\scaleB\linewidth]{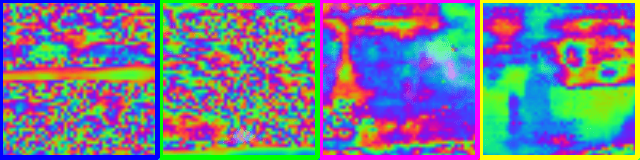}
			\includegraphics[width=\scaleB\linewidth]{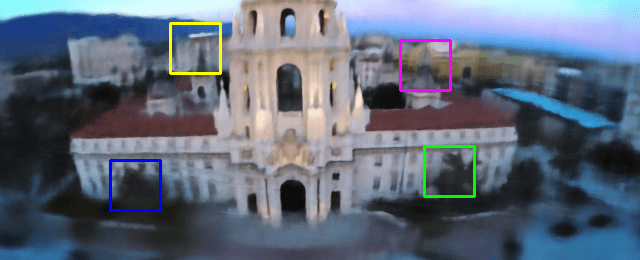}
			\includegraphics[width=\scaleB\linewidth]{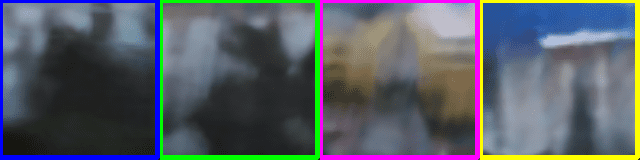}
			\includegraphics[width=\scaleB\linewidth]{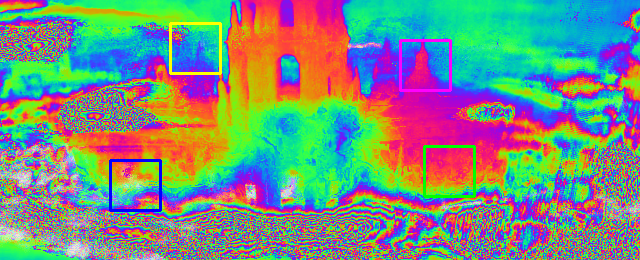}
			\includegraphics[width=\scaleB\linewidth]{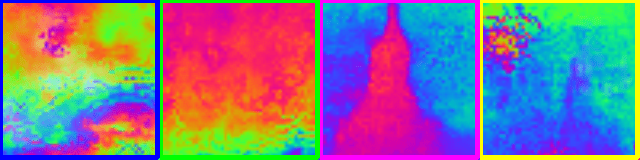}

	\end{minipage}}
	\subfigure[MipNeRF\cite{barron2021mipnerf}]{
		\begin{minipage}[t]{\scale\linewidth}
			\includegraphics[width=\scaleB\linewidth]{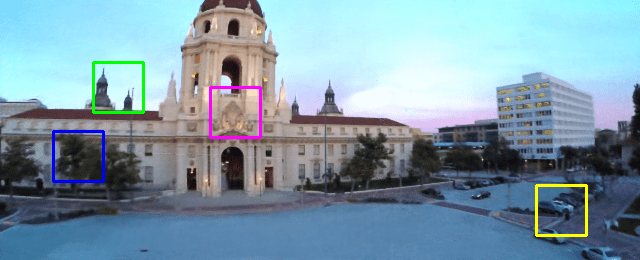}
			\includegraphics[width=\scaleB\linewidth]{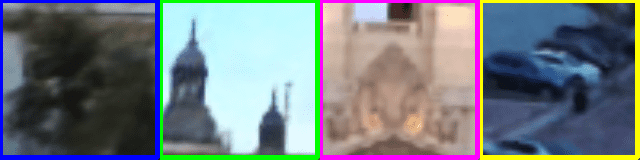}
			\includegraphics[width=\scaleB\linewidth]{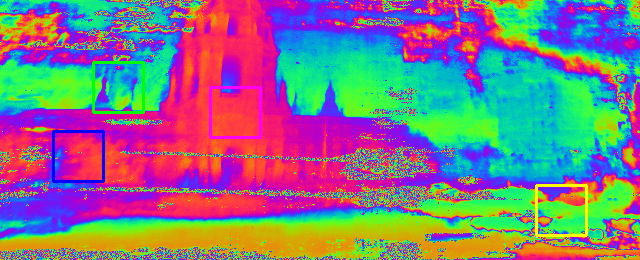}
			\includegraphics[width=\scaleB\linewidth]{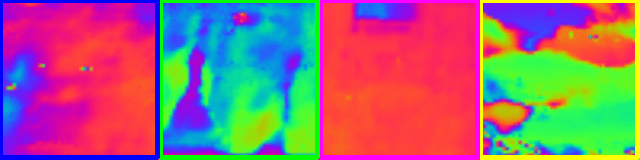}
			\includegraphics[width=\scaleB\linewidth]{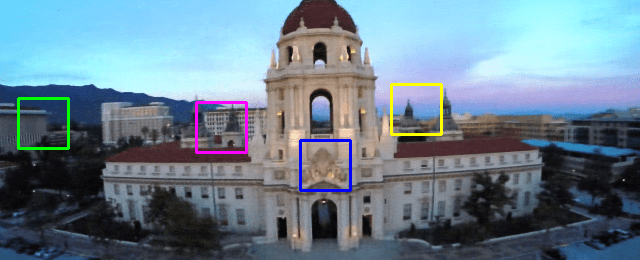}
			\includegraphics[width=\scaleB\linewidth]{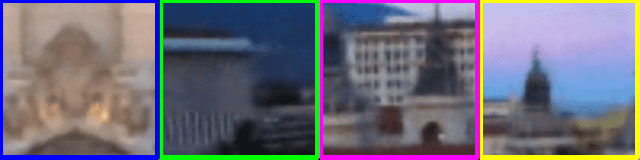}v
			\includegraphics[width=\scaleB\linewidth]{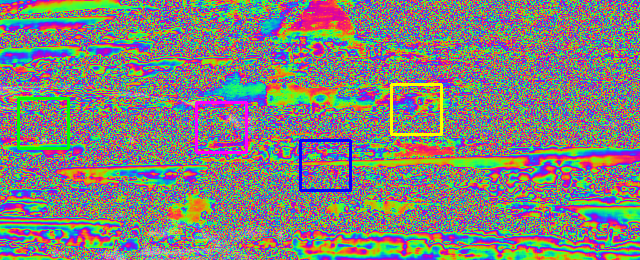}
			\includegraphics[width=\scaleB\linewidth]{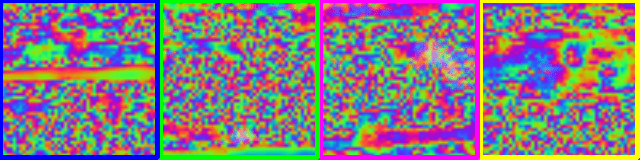}
			\includegraphics[width=\scaleB\linewidth]{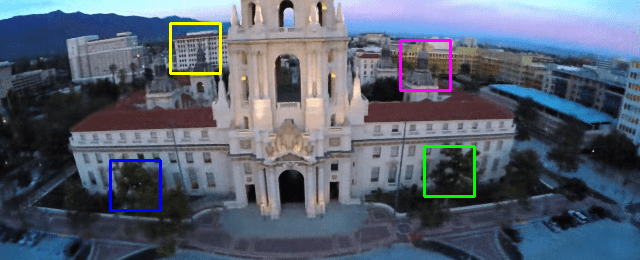}
			\includegraphics[width=\scaleB\linewidth]{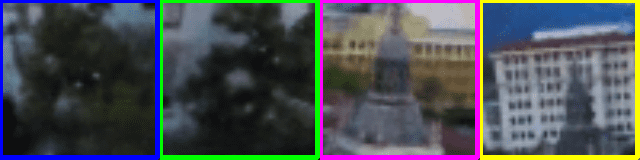}
			\includegraphics[width=\scaleB\linewidth]{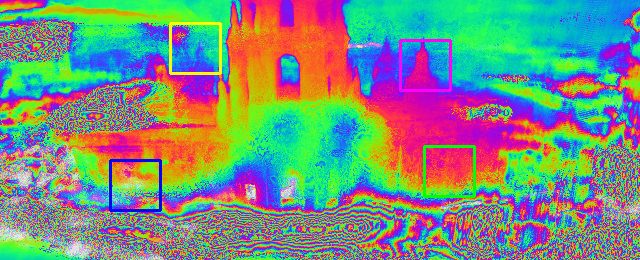}
			\includegraphics[width=\scaleB\linewidth]{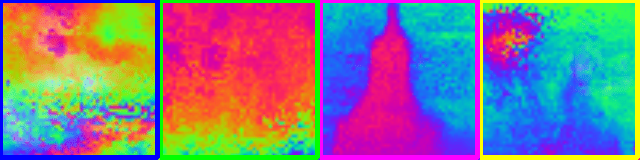}
	\end{minipage}}
	\subfigure[Our]{
		\begin{minipage}[t]{\scale\linewidth}
			\includegraphics[width=\scaleB\linewidth]{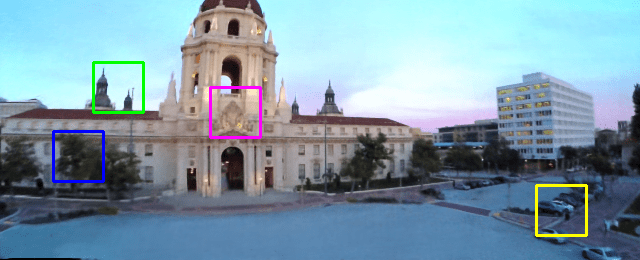}
			\includegraphics[width=\scaleB\linewidth]{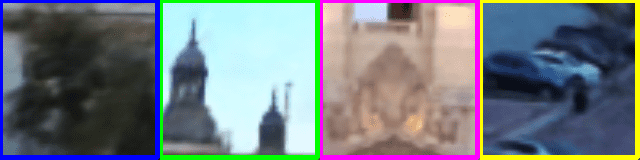}
			\includegraphics[width=\scaleB\linewidth]{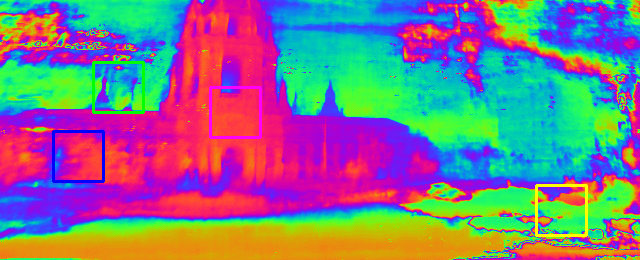}
			\includegraphics[width=\scaleB\linewidth]{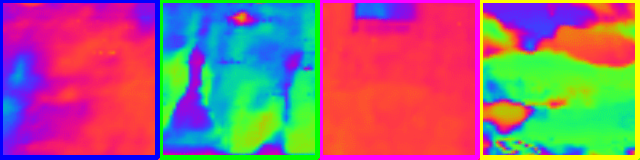}
			\includegraphics[width=\scaleB\linewidth]{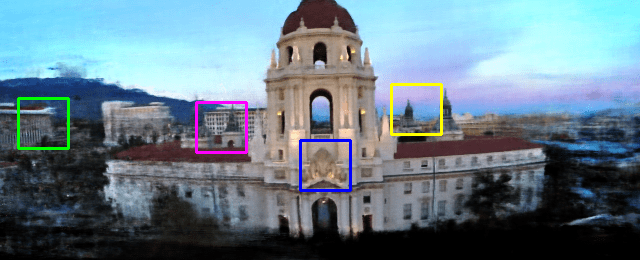}
			\includegraphics[width=\scaleB\linewidth]{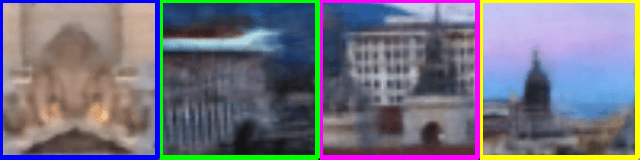}
			\includegraphics[width=\scaleB\linewidth]{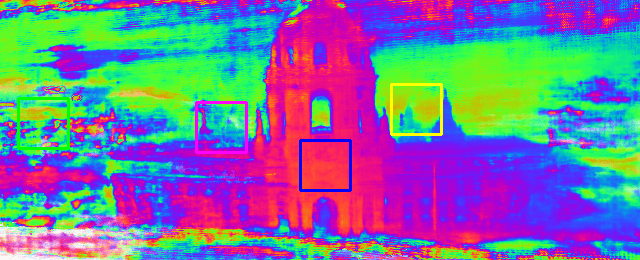}
			\includegraphics[width=\scaleB\linewidth]{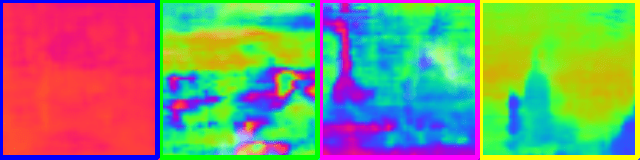}
			\includegraphics[width=\scaleB\linewidth]{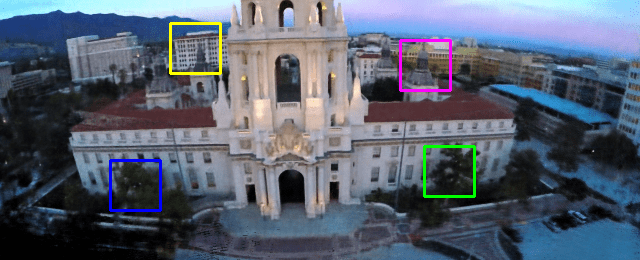}
			\includegraphics[width=\scaleB\linewidth]{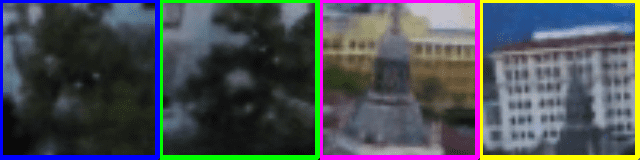}
			\includegraphics[width=\scaleB\linewidth]{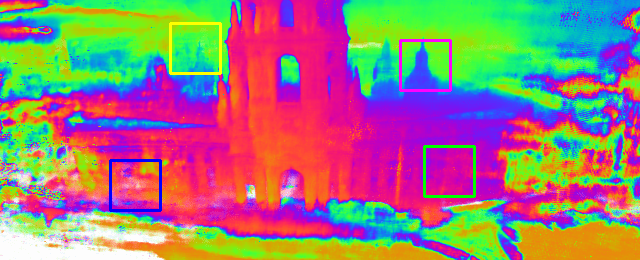}
			\includegraphics[width=\scaleB\linewidth]{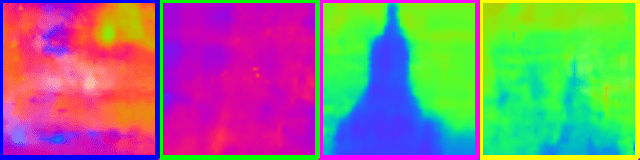}
	\end{minipage}}

	\caption{UAV scene reconstruction. We compare our method against NeRF~\cite{mildenhall2020nerf}, KiloNeRF~\cite{Reiser2021ICCV}, MipNeRF~\cite{barron2021mipnerf}.}
	\label{fig:uav01}
\end{figure*}

{
In this section, we show the results of ground scene reconstruction from UAV video in \figref{fig:uav01}. UAV views contain comparatively large viewpoint and camera pose change, which is a challengine task for neural rendering. NeRF~\cite{mildenhall2020nerf} shows blur rendering results due to incorrect density estimation of scene. MipNeRF~\cite{barron2021mipnerf} fails to estimate correct density in second row of results, partially because a large viewpoint changes leads to sample in the space that has insufficient training samples and outputs random density values. Our method takes advantage of (1) octtree that bounds whole scene and skip empty space, and (2) distributed sub-network architecture that trains and renders locally to avoid inbalanced sampling inside each octtree nodes and adaptively reallocate computational resource for each node.
}

%\newpage
%\input{algo}

\end{document}